\documentclass[times,final]{elsarticle}
\usepackage{jcomp}
\usepackage{framed,multirow}
\usepackage[utf8]{inputenc}
\usepackage[T1]{fontenc}
\usepackage{siunitx}
\usepackage{graphicx}
\usepackage{subcaption}
\usepackage{hyperref,float}
\usepackage{booktabs}
\usepackage{makecell}
\usepackage{tabu}
\usepackage{float}
\usepackage{placeins}
\usepackage{multirow} 
\usepackage{amssymb}
\usepackage{latexsym}
\usepackage{amsmath}
\usepackage{geometry}
\usepackage{ulem}
\usepackage{algpseudocode}
\usepackage{algorithmicx}
\usepackage{algorithm}
\usepackage{caption}
\usepackage{amsfonts}
\usepackage{float}
\usepackage{url}
\usepackage{xcolor}

\newtheorem{remark}{Remark}[section]
\definecolor{newcolor}{rgb}{.8,.349,.1}

\journal{Journal of Computational Physics}

\usepackage[nolist]{acronym}
\begin{acronym}
\acro{PDEs}[PDEs]{partial differential equations}
\acro{BC}{boundary condition}
\acro{KANs}{Kolmogorov-Arnold networks}
\acro{KAN}{Kolmogorov-Arnold network}
\acro{SciML}{Scientific machine learning}
\acro{BCs}{\ac{BC}s}
\acro{MLPs}{multilayer perceptrons}
\acro{ML}{machine learning}
\acro{RBFs}{radial basis functions}
\acro{PINNs}{physics-informed neural networks}
\acro{DeepONet}{deep operator networks}
\acro{EDNNs}{evolutionary deep neural networks}
\acro{EvoKAN}{evolutionary \ac{KAN}}
\acro{GP}{Gaussian process}
\acro{SAV}{scalar auxiliary variable}
\acro{EDNN}{evolutionary deep neural networks}
\acro{EvoKAN}{evolutionary \ac{KAN}}
\acro{PINN}{physics-informed neural networks}
\end{acronym}

\begin{document}

% ========================================= %
% ================ authors ================ %
% ========================================= %
\verso{Bongseok Kim \textit{et al.}}
\begin{frontmatter}
\title{BEKAN: Boundary condition--guaranteed evolutionary Kolmogorov--Arnold networks with radial basis functions for solving PDE problems}%\footnote{This work is partially supported by  NSF DMS-1720442 and AFOSR FA9550-20-1-0309.}}%
\author[1]{Bongseok \snm{Kim}\fnref{fn1}}
\author[2]{Jiahao \snm{Zhang}\fnref{fn1}}
%% e-mail
\fntext[fn1]{These two authors contributed equally to this work.}
\author[1,2]{Guang \snm{Lin}\corref{cor1}}
\cortext[cor1]{Corresponding authors.}
\ead{guanglin@purdue.edu}
\address[1]{School of Mechanical Engineering, Purdue University, West Lafayette, IN 47907, USA}
\address[2]{Department of Mathematics, Purdue University, West Lafayette, IN 47907, USA}

% ========================================= %
% =============== abstract ================ %
% ========================================= %
\begin{abstract}

%% == motivation ==
Deep learning has gained attention for solving \ac{PDEs}, but the black-box nature of neural networks hinders precise enforcement of boundary conditions.
% While deep learning has garnered significant attention for its application in solving \ac{PDEs}, the inherent black-box nature of neural networks presents challenges in precisely enforcing boundary conditions.
% Deep learning approaches for solving partial differential equations (PDEs) have been gaining attention in scientific computing. 
% However, due to The black--box nature of the network architectures, precise implementation of boundary conditions remains a challenging problem. 
% Hard constraints may reduce the representational capacity of the network, while soft constraints do not guarantee exact satisfaction of the BCs.
%% proposed method
To address this, we propose a boundary condition-guaranteed evolutionary Kolmogorov-Arnold Network (KAN) with radial basis functions (BEKAN).
In BEKAN, we propose three distinct and combinable approaches for incorporating Dirichlet, periodic, and Neumann boundary conditions into the network.
% with embedded boundary information.
% To address this, we propose BEKAN, a novel evolutionary neural network architecture based on Kolmogorov-Arnold Networks (KANs) with Gaussian radial basis functions (RBFs).
% 1. Dirichlet
For Dirichlet problem, we use smooth and global Gaussian RBFs to construct univariate basis functions for approximating the solution and to encode boundary information at the activation level of the network.
% 3. periodic
To handle periodic problems, we employ a periodic layer constructed from a set of sinusoidal functions to enforce the boundary conditions exactly.
% 2. Neumann
For a Neumann problem, we devise a least-squares formulation to guide the parameter evolution toward satisfying the Neumann condition.
%% results
By virtue of the boundary-embedded RBFs, the periodic layer, and the evolutionary framework, we can perform accurate PDE simulations while rigorously enforcing boundary conditions.
For demonstration, we conducted extensive numerical experiments on Dirichlet, Neumann, periodic, and mixed boundary value problems.
The results indicate that BEKAN outperforms both multilayer perceptron (MLP) and B-splines KAN in terms of accuracy.
%% conclusion
% In conclusion, the proposed approach enhances the capability of KAN in solving PDE problems while satisfying boundary conditions, thereby facilitating advances in scientific computing and engineering applications.
In conclusion, the proposed approach enhances the capability of KANs in solving PDE problems while satisfying boundary conditions, thereby facilitating advancements in scientific computing and engineering applications.
% In this paper, we address the challenge of implementing boundary conditions (BCs) when solving partial differential equations (PDEs) using neural networks, a topic where traditional numerical methods offer mature solutions but neural approaches often falter. We propose BEKAN, a novel evolutionary neural network architecture based on Kolmogorov-Arnold Networks (KANs) that intrinsically embed boundary condition information into the network by leveraging the radial basis. Our comparative analysis shows that BEKAN surpasses traditional fully connected network architectures in both accuracy and efficiency when solving PDEs with Dirichlet, Neumann, periodic, and mixed boundary conditions. The BEKAN enhances the capability of neural networks in solving forward PDE problems with statisfying the BCs.
\end{abstract}

\begin{keyword}
Deep learning \\
Partial Differential Equations \\
Boundary Condition \\
Kolmogorov-Arnold Networks \\
Evolutionary Neural Networks \\
Radial Basis Functions
\end{keyword}

%\begin{keyword}
%% Keywords
%\KWD Keyword1\sep Keyword2\sep Keyword3
%\end{keyword}

\end{frontmatter}

%\linenumbers

% ====================
% document starts here
% =====================

% =====================
% section: introduction
% =====================
\section{Introduction}\label{sec:Introduction}

% \ac{SciML} has opened new directions in solving \ac{PDEs} by leveraging deep neural networks thanks to their universal approximation properties and high representational capacity~\cite{cuomo2022scientific}.
\ac{SciML} has opened new avenues for solving \ac{PDEs}, leveraging the universal approximation properties and high representational capacity of deep neural networks~\cite{cuomo2022scientific}.
Frameworks such as \ac{PINNs}~\cite{raissi2019physics, karniadakis2021physics,boulle2023mathematical}, \ac{DeepONet}~\cite{lu2022multifidelity, zhu2023fourier, lu2021learning}, and \ac{EDNNs}~\cite{du2021evolutional, zhang2024energy} have demonstrated promising performance in approximating various PDE solutions. 
However, the lack of interpretability in neural network~\cite{fan2021interpretability} continues to impede the rigorous implementation of boundary conditions~\cite{marquez2017imposing}.
A prevailing approach, soft constraint methods, relies on balancing PDE and boundary losses during training~\cite{raissi2019physics} and thus lack structural guarantees for satisfying boundary conditions~\cite{sukumar2022exact, berrone2023enforcing}. Consequently, the challenge of enforcing boundary constraints continues to motivate  extensive follow-up studies in \ac{SciML}.

To overcome the limitations of soft constraints, a variety of hard-constraint approaches have been proposed. 
% For instance, Liu et al.~\cite{liu2022unified} capitalized general solutions as part of the network ansatz to enforce the Dirichlet, Neumann, and Robin boundary conditions.
% Sukumar and Srivastava~\cite{sukumar2022exact} and Wang et al.~\cite{wang2023exact} employed distance functions to embed Dirichlet conditions directly into the network output.
For instance, Liu et al.\cite{liu2022unified} incorporated general solutions into the neural network formulation to satisfy Dirichlet, Neumann, and Robin boundary conditions. In a different approach, Sukumar and Srivastava\cite{sukumar2022exact}, along with Wang et al.~\cite{wang2023exact}, utilized distance functions to explicitly encode Dirichlet boundary constraints into the network output.
Dong and Ni~\cite{stevendong} introduced a periodic input transformation based on Fourier series to enforce periodic boundary conditions.
Straub et al.~\cite{straub2025hard} enforced Neumann constraints via Fourier feature embeddings and output transformations. 
% Horie \& Mitsume~\cite{horie2022physics} embedded Dirichlet, Neumann, and Robin conditions into GNNs by incorporating boundary rules into message passing and aggregation steps.
As a hybrid approach, PINN-FEM~\cite{sobh2025pinn} enforces Dirichlet conditions through variational loss and output transformation.
In parallel with neural networks, \ac{GP} can also incorporate boundary conditions, either through basis function design~\cite{solin2019know, lange2021linearly} or kernel construction based on Green’s functions~\cite{ding2019bdrygp}.

% In parallel with neural networks, \ac{GP} methods can also incorporate boundary conditions through basis function design. Solin and Kok~\cite{solin2019know} used harmonic features to enforce constraints in the feature space. Ding et al.~\cite{ding2019bdrygp} designed kernels via Green’s functions to satisfy Dirichlet and Neumann conditions. Lange-Hegermann~\cite{lange2021linearly} employed Groebner bases to define \ac{GP} priors in the null space of \ac{PDEs} and boundary operators, but mainly limited to linear systems.

While the aforementioned approaches have demonstrated considerable success, there remains room for improvement in the following aspects:
\begin{enumerate}[(i)]
    \item Previous works relied on \ac{MLPs} with fixed activations, requiring manual output shaping or applying distance metrics to encode boundary conditions.
    % \item Imposing Neumann boundary conditions remains challenging and requires effort due to problem-specific formulations~\cite{liu2022unified} or limited generalization across diverse domain shapes~\cite{straub2025hard}.
    \item Imposing Neumann conditions remains challenging and often requires problem-specific formulations~\cite{liu2022unified} or exhibits limited generalization across diverse domain shapes~\cite{straub2025hard}.
    \item Solving chaotic, nonlinear, and high--order \ac{PDEs} under hard constraints remains limited, as such constraints can reduce expressiveness~\cite{marquez2017imposing}.
\end{enumerate}

To address (i)–(iii), we propose BEKAN, a boundary condition-guaranteed evolutionary Kolmogorov–Arnold network composed of three key components: \ac{KANs}, an evolutionary network, and Gaussian \ac{RBFs}. 
Regarding (i), we leverage \ac{KANs}, which employ trainable spline-based activation functions~\cite{liu2024kan,liu2024kan2}, replacing the fixed nonlinearities used in MLPs~\cite{apicella2021survey,trentin2001networks}, and provide flexibility through locally adaptable basis functions~\cite{kolmogorov1957representation,kolmogorov1961representation,braun2009constructive}.
% This design allows boundary information to be directly embedded at the activation level.
This design enables direct embedding of boundary information at the activation level.
% To wit, we impose Dirichlet hard constraints on the \ac{KANs} basis functions and assemble them to represent the solution as a series.
We introduce a novel method to embed Dirichlet conditions directly into the KAN basis functions. This contrasts with traditional methods that modify the final network output~\cite{guyiqi}, and by building the constraint into the network functional structure, we achieve greater stability and expressiveness, particularly for problems with sharp gradients near boundaries, as demonstrated in Sec.~\ref{sec:Experiments}.

With regard to (ii), we adopt an evolutionary network~\cite{du2021evolutional, zhang2024energy, lin2025energy} to guide network parameters toward satisfying Neumann conditions. The evolutionary network initializes weights based on the initial condition and updates them over time using discretized forms of the PDE, enabling efficient temporal evolution.
In the context of the Neumann boundary condition, we formulate a least-squares problem that serves as an iterative step for parameter evolution, incorporating a Neumann term to direct the parameter adjustment toward boundary compliance.
The Neumann term in the least-squares problem, originating from the Neumann boundary condition, serves as an additional constraint to consistently satisfy the Neumann boundary condition, enabling stable solution prediction over the entire time range.

Finally, to tackle (iii), we employ smooth and globalized Gaussian RBFs to construct univariate representations, achieving accurate solutions while ensuring boundary enforcement.
Although B-spline-based \ac{KANs} offer local adaptability, they often suffer from training instability when inputs exceed the predefined spline domain, thereby requiring frequent rescaling~\cite{li2024kolmogorov}.
Moreover, B-splines vanish outside their support~\cite{liu2024kan}, which may limit expressiveness under hard constraints. In contrast, Gaussian RBFs smoothly approach zero when evaluated far from the center, maintaining their smoothness under higher-order derivatives, making them well-suited for challenging PDEs, as comprehensively demonstrated in Sec.~\ref{sec:Experiments}.

In contrast to previous approaches for solving boundary value problems, the proposed method provides several key benefits as outlined below:
\begin{enumerate}
\item The introduced method encodes boundary conditions at the activation level, enabling accurate and stable enforcement of Dirichlet conditions.
\item The introduced method leverages an evolutionary network framework that iteratively updates parameters via a least-squares formulation, incorporating Neumann boundary terms to naturally enforce boundary compliance over time.
\item The introduced method exploits Gaussian RBFs to construct univariate activation functions for \ac{KANs}, allowing effective handling of chaotic, nonlinear, and high--order PDE problems, such as Kuramoto–Sivashinsky equation, under hard constraints.
\item The introduced method effectively solves \ac{PDEs} with mixed boundary conditions, as demonstrated in Sec.~\ref{sec:experiment_mixed}.
% \item The proposed method provides a comprehensive framework capable of Dirichlet, Neumann, periodic, and mixed boundary conditions.
\end{enumerate}

The subsequent sections of this paper are structured as follows. Section~\ref{sec:Preliminary} introduces the BEKAN architecture and outlines its key components: \ac{KANs}, \ac{KANs} with Gaussian \ac{RBFs}, and the evolutionary network. Section~\ref{sec:Methodology} details the formulation of the BEKAN framework for Dirichlet, Neumann, and periodic boundary conditions. Section~\ref{sec:Experiments} presents numerical experiments on benchmark PDEs to assess the effectiveness of the proposed approach with respect to both accuracy and satisfaction of boundary constraints. Finally, Sec.~\ref{sec:Conclusion} summarizes the main contributions and outlines possible directions for future research.

% ===================
% section: Preliminary
% ===================
\section{Evolutionary Kolmogorov-Arnold networks with radial basis functions} \label{sec:Preliminary}

% In this section, we begin by introducing \ac{KAN}, Gaussian \ac{RBFs}, and evolutionary network. Then we present the formulation of the evolutionary \ac{KAN} with \ac{RBFs} by integrating the three components.
% ============================
% subsection: Kolmogorov-Arnold Networks
% ============================
\subsection{Kolmogorov–Arnold networks}
Conceptually, unlike an MLP, which learns a single fixed activation function per layer, a \ac{KAN} learns univariate activation functions on each edge connecting the neurons. The output of a neuron is the sum of these transformed signals, allowing for a more expressive and interpretable function representation. \ac{KAN} is formulated based on the Kolmogorov--Arnold theorem, which asserts that every continuous function of multiple variables over a bounded domain can be approximated by a finite sum of continuous univariate functions~\cite{liu2024kan,koppen2002training,lin1993realization,lai2021kolmogorov,leni2013kolmogorov,fakhoury2022exsplinet,he2023optimal}.
Following this theoretical foundation, for a smooth mapping \( f: [0,1]^n \to \mathbb{R} \), we adopt the following structured representation:
\begin{equation}
    f(\mathbf{x}) = f(x_1,\dots,x_n) = \sum_{q=1}^{2n+1} \Phi_q\left(\sum_{p=1}^n \phi_{q,p}(x_p)\right), \label{eq:KART}
\end{equation}
where \( \phi_{q,p}: [0,1] \to \mathbb{R} \) and \( \Phi_q: \mathbb{R} \to \mathbb{R} \) denote continuous univariate functions.
To implement Eq.~\eqref{eq:KART}, the functions $\phi_{q,p}$ and $\Phi_q$ are instantiated using third-order B-spline basis functions~\cite{liu2024kan}. The transformation defined by these functions can be collected into a matrix $\mathbf{\Phi} = \{\phi_{q,p}\}$, where indices range over \( p = 1, \dots, n_{\rm in} \) and \( q = 1, \dots, n_{\rm out} \). According to the Kolmogorov--Arnold construction, the inner layer performs functional compositions with \( n_{\rm in} = n \) and \( n_{\rm out} = 2n + 1 \), and the subsequent outer layer transforms the resulting \( 2n + 1 \) outputs into a single scalar value, thus setting \( n_{\rm out} = 1 \) for the final mapping.

% In the context of structural similarity with traditional \ac{MLPs}, the configuration in Eq.~\eqref{eq:KART} resembles a three-layer neural network, with the input and middle layers having widths of $n$ and $2n+1$, respectively. However, unlike standard architectures, the activation functions are trainable and are assigned to edges rather than nodes. For the later stacking of the deep layer.

% To stack the layers using Eq.~\eqref{eq:KART}, we aggregate the univariate functions successively.
% In a \ac{KAN} architecture, each layer receives information from the preceding one through a collection of univariate transformations. These transformed signals are then combined by summing them at each node, enabling the entire network to be interpreted as a hierarchical composition of scalar functions. Formally, the network architecture is specified by a sequence of integers $[n_0, n_1, \dots, n_L]$, where each $n_i$ denotes the number of neurons in the $i^{\text{th}}$ layer.
% Let $x_{l,i}$ denote the activation output from the $i^{\text{th}}$ neuron in the $l^{\text{th}}$ layer. For each adjacent pair of layers $l$ and $l+1$, a unique activation function $\phi_{l,j,i}$ is assigned to the connection from neuron $i$ in layer $l$ to neuron $j$ in layer $l+1$. These functions are indexed as follows:
% \begin{equation}
% \label{eq:phi_tensor}
%     \phi_{l,j,i}, \quad l = 0, \dots, L-1, \quad i = 1, \dots, n_l, \quad j = 1, \dots, n_{l+1}.
% \end{equation}
To stack the layers using Eq.~\eqref{eq:KART}, we sequentially compose the univariate functions.  
Within a \ac{KAN} architecture, each layer processes signals passed from the previous one through a set of univariate transformations.  
These outputs are then aggregated via summation at each node, allowing the overall network to be interpreted as a hierarchical composition of scalar mappings.  
The architecture is characterized through a list of integers $[n_0, n_1, \dots, n_L]$, with each $n_i$ indicating the number of neurons in layer $i$.
Let $x_{l,i}$ represent the activation from the $i^{\text{th}}$ neuron in the $l^{\text{th}}$ layer.  
For every pair of consecutive layers $l$ and $l+1$, we assign a distinct activation function $\phi_{l,j,i}$ to the link connecting neuron $i$ in layer $l$ to neuron $j$ in layer $l+1$.  
These activation functions are indexed as:
\begin{equation}
\label{eq:phi_tensor}
    \phi_{l,j,i}, \quad l = 0, \dots, L-1, \quad i = 1, \dots, n_l, \quad j = 1, \dots, n_{l+1}.
\end{equation}
Each function takes $x_{l,i}$ as input and returns a processed value given by $\tilde{x}_{l,j,i} = \phi_{l,j,i}(x_{l,i})$. The total input received by neuron $j$ in layer $l+1$ is obtained by summing over all activations from the preceding layer:
\begin{equation}
\label{eq:tensor_KAN}
    x_{l+1,j} = \sum_{i=1}^{n_l} \tilde{x}_{l,j,i} = \sum_{i=1}^{n_l} \phi_{l,j,i}(x_{l,i}), \quad j = 1, \dots, n_{l+1}.
\end{equation}
We can represent Eq.~\eqref{eq:tensor_KAN} in matrix form as follows:
\begin{equation}
\label{eq:matrix}
    \mathbf{x}_{l+1} = 
    \underbrace{\begin{pmatrix}
        \phi_{l,1,1}(\cdot) & \phi_{l,1,2}(\cdot) & \cdots & \phi_{l,1,n_l}(\cdot) \\
        \phi_{l,2,1}(\cdot) & \phi_{l,2,2}(\cdot) & \cdots & \phi_{l,2,n_l}(\cdot) \\
        \vdots & \vdots &  & \vdots \\
        \phi_{l,n_{l+1},1}(\cdot) & \phi_{l,n_{l+1},2}(\cdot) & \cdots & \phi_{l,n_{l+1},n_l}(\cdot) \\
    \end{pmatrix}}_{\mathbf{\Phi}_l}
    \mathbf{x}_l.
\end{equation}
Here, $\mathbf{\Phi}_l$ represents the collection of univariate functions applied at the $l^{\text{th}}$ layer of the KAN. A standard KAN architecture applies $L$ such layers in sequence, transforming an input vector $\mathbf{x}_0 \in \mathbb{R}^{n_0}$ through a series of function compositions to ultimately produce the network output:
\begin{equation}\label{eq:kan}
    \text{KAN}(\mathbf{x}) = (\mathbf{\Phi}_{L-1} \circ \mathbf{\Phi}_{L-2} \circ \cdots \circ \mathbf{\Phi}_1 \circ \mathbf{\Phi}_0)\mathbf{x}.
\end{equation}
In case of $n_L = 1$, we define $f(\mathbf{x}) \equiv \text{KAN}(\mathbf{x})$ and express the equation in a form analogous to Eq.~(\ref{eq:KART}):
\begin{equation}
    f(\mathbf{x}) = \sum_{i_{L-1}=1}^{n_{L-1}} \phi_{L-1,i_L,i_{L-1}} \left(\sum_{i_{L-2}=1}^{n_{L-2}} \cdots \left(\sum_{i_2=1}^{n_2} \phi_{2,i_3,i_2} \left(\sum_{i_1=1}^{n_1} \phi_{1,i_2,i_1}\left(\sum_{i_0=1}^{n_0} \phi_{0,i_1,i_0}(x_{i_0})\right)\right)\right)\cdots\right).
\end{equation}

% =================================================== %
% =================================================== %
\subsection{Kolmogorov-Arnold networks with radial basis functions}

We now introduce a \ac{RBFs}-based \ac{KAN}, which not only provides a simplified implementation and improved computational efficiency of \ac{KAN}~\cite{li2405kolmogorov}, but also serves as a crucial component for embedding boundary conditions, as discussed later in Sec.~\ref{sec:Methodology}.
% \ac{RBFs}~\cite{orr1996introduction, buhmann2000radial} determine their output solely as a function of the radial distance from a center point.
\ac{RBFs}~\cite{orr1996introduction, buhmann2000radial} compute their values depending only on the radial distance between the input and a predefined center.
A fundamental approach in RBF modeling involves approximating a target function using a linear combination of radially symmetric functions, each localized around a specific point in the input space. In \ac{KAN} with \ac{RBFs}, we approximate each function $\phi_{l,j,i}$ in Eq.~\eqref{eq:phi_tensor} as $\hat{\phi}_{l,j,i}$ using a sum of \ac{RBFs}, expressed as:
\begin{equation}
\label{eq:RBFs}
    \phi_{l,j,i}(x) \approx \hat{\phi}_{l,j,i}(x) = \sum_{k=1}^{g} w^k_{l,j,i} \, \psi(\|x - c_k\|),
\end{equation}
where $w^k_{l,j,i}$ are the learnable weights, $c_k$ are the center locations, and the symbol $\psi$ represents the chosen radial basis function. 
% Equation~\eqref{eq:RBFs} applies to \ac{KAN} layer indices $l = 0, \dots, L-1$, neuron indices $i = 1, \dots, n_l$, and $j = 1, \dots, n_{l+1}$.
Equation~\eqref{eq:RBFs} is applied across all layers indexed by $l$ from $0$ to $L{-}1$, covering all connections from neuron $i$ in layer $l$ to neuron $j$ in the subsequent layer.

Among various types of \ac{RBFs}, we adopt the Gaussian form for $\psi$, defined by:
\begin{equation}
    \psi(r) = \exp\left(-\frac{r^2}{2h^2}\right).
\end{equation}
% where $r$ denotes the distance from the center, and $h$ is a scale parameter that determines the spread or influence of the function.
Here, $r$ indicates the radial distance to the center, while $h$ serves as a scaling factor controlling the function width and influence.
A sequence of cubic B-spline basis functions can be closely approximated using Gaussian RBFs through linear transformations~\cite{li2405kolmogorov}.

By replacing $\phi_{l,j,i}$ and $\Phi_l$ with their \ac{RBFs} approximations $\hat{\phi}_{l,j,i}$ and $\hat{\Phi}_l$, respectively, we obtain the corresponding \ac{KAN} architecture based on \ac{RBFs} as follows:
\begin{equation}\label{eq:RadialKAN}
    \text{RadialKAN}(\mathbf{x}) = \left(  \mathbf{\hat{\Phi}}_{L-1} \circ \mathbf{\hat{\Phi}}_{L-2} \circ \cdots \circ \mathbf{\hat{\Phi}}_1 \circ \mathbf{\hat{\Phi}}_0 \right)  \mathbf{x}.
\end{equation}
In case of $n_L = 1$, we define $f(\mathbf{x}) \equiv \text{RadialKAN}(\mathbf{x})$ and express the equation in a form analogous to Eq.~(\ref{eq:KART}):
\begin{equation}
    f(\mathbf{x}) = \sum_{i_{L-1}=1}^{n_{L-1}} \hat{\phi}_{L-1,i_L,i_{L-1}} \left(\sum_{i_{L-2}=1}^{n_{L-2}} \cdots \left(\sum_{i_2=1}^{n_2} \hat{\phi}_{2,i_3,i_2} \left(\sum_{i_1=1}^{n_1} \
    \hat{\phi}_{1,i_2,i_1}\left(\sum_{i_0=1}^{n_0} \hat{\phi}_{0,i_1,i_0}(x_{i_0})\right)\right)\right)\cdots\right).
\end{equation}
% Additionally, to stabilize training and ensure that inputs remain within the effective range of the \ac{RBFs}, we apply layer normalization~\cite{ba2016layer} to each layer of the network.
To enhance training stability and keep the inputs within the responsive domain of the \ac{RBFs}, we employ layer normalization~\cite{ba2016layer} at every layer of the network.

% =================================================== %
% =================================================== %

\subsection{Evolutionary Kolmogorov-Arnold networks with radial basis functions}
\label{EvoKAN_RBFs}

\begin{figure}[htp!]
    \centering
    \begin{subfigure}[b]{0.75\linewidth}
        \centering
        \includegraphics[width=\linewidth]{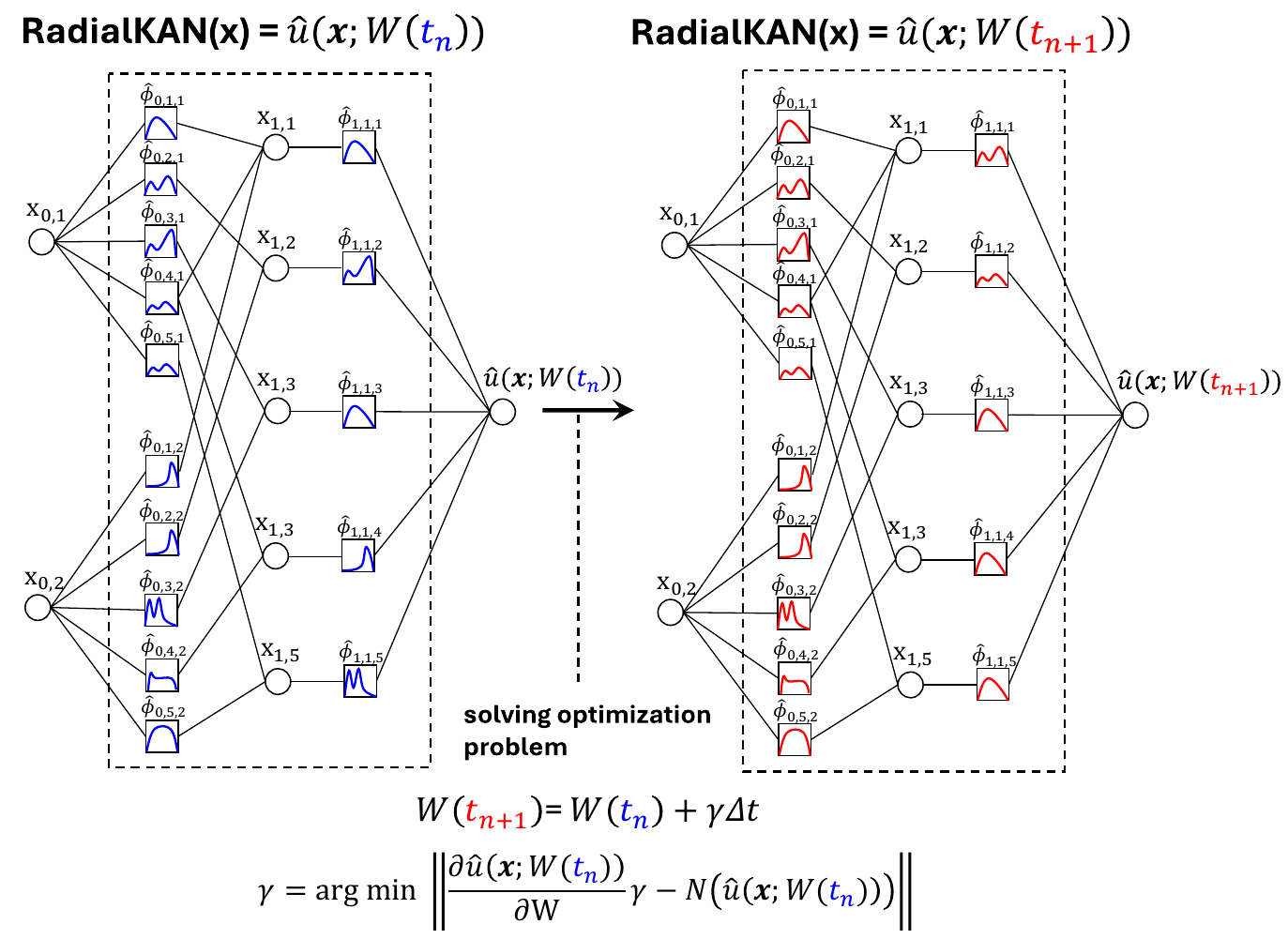}
    \end{subfigure}
    \caption{Evolutionary Kolmogorov-Arnold networks with Gaussian \ac{RBFs}.}
    \label{fig:EvoRadialKAN}
\end{figure}

% In this section, we extend the RadialKAN to an evolutionary network~\cite{du2021evolutional, zhang2024energy, lin2025energy}. We begin with a general nonlinear PDE equipped with an initial condition:
This section generalizes the RadialKAN framework to incorporate an evolutionary network architecture~\cite{du2021evolutional, zhang2024energy, lin2025energy}. As a starting point, we consider a general form of a nonlinear PDE accompanied by an initial condition:
\begin{equation}
\label{eqn:general-pde}
 \begin{aligned}
    &\frac{\partial u}{\partial t} + \mathcal{N}(u) = 0, \\
    &u(\mathbf{x}, 0) = f(\mathbf{x}),
 \end{aligned}
\end{equation}
% Here, $u(x, t) = (u_1, u_2, \dots, u_m)$ denotes a vector-valued function, $x = (x_1, x_2, \dots, x_d)$ specifies the spatial position, and $\mathcal{N}$ is a nonlinear differential operator applied to $u$.
In Eq.~\eqref{eqn:general-pde}, the function $u(\mathbf{x}, t) = (u_1, u_2, \dots, u_m)$ represents a multicomponent field, where $\mathbf{x} = (x_1, x_2, \dots, x_d)$ denotes the spatial coordinates, and $\mathcal{N}$ indicates a nonlinear differential operator acting on $u$.

We now represent the solution $u$ using the RadialKAN approximation $\hat{u}$, parameterized by a network with $L+1$ layers:
\begin{equation}
\hat{u}(\mathbf{x}, {W}(t)) = (\hat{u}_1, \hat{u}_2, \dots, \hat{u}_m) = \text{RadialKAN}(\mathbf{x}),
\end{equation}
where ${W}(t)$ is a time-dependent vector that collects all trainable parameters of the network.  
% Based on the chain rule, we construct the time derivative of $\hat{u}$ as follows:
Applying the chain rule yields the following expression for the time derivative of $\hat{u}$:
\begin{equation}
\frac{\partial \hat{u}}{\partial t} = \frac{\partial \hat{u}}{\partial W} \frac{\partial {W}}{\partial t},
\end{equation}
where the derivative $\frac{\partial {W}}{\partial t}$ governs the direction of parameter evolution.  
In the evolutionary network, we require the derivative $\frac{\partial {W}}{\partial t}$ at each time step.  
For this purpose, we solve the following optimization problem, where we minimize $\mathcal{J}$ derived from the residual of Eq.~\eqref{eqn:general-pde}:
\begin{equation}\label{Wmin}
\frac{\partial {W}}{\partial t} = \arg\min_{\gamma} \mathcal{J}(\gamma), \quad \mathcal{J}(\gamma) = \frac{1}{2} \int_{X} \left\Vert \frac{\partial \hat{u}}{\partial {W}} \gamma + \mathcal{N}(\hat{u}) \right\Vert_2^2 \, \mathrm{d}\mathbf{x}.
\end{equation}
By the first-order optimality condition, we seek the optimal solution of Eq.~\eqref{Wmin} by solving the following system:
\begin{equation}\label{OptCond}
\nabla_{\gamma} \mathcal{J}(\gamma_{\text{opt}}) = 
\int_{X} 
\left( \frac{\partial \hat{u}}{\partial {W}}\right)^T
\left(
\frac{\partial \hat{u}}{\partial {W}}  \gamma_{\text{opt}} + \mathcal{N}(\hat{u}) \right)\mathrm{d} \mathbf{x} = 0.
\end{equation}
% To obtain an approximate solution $\gamma_{\text{opt}}$ to Eq.~\eqref{OptCond}, we reformulate the Eq.~\eqref{OptCond} as a least--squares problem as follow:
To approximate the solution $\gamma_{\text{opt}}$ to Eq.~\eqref{OptCond}, we recast Eq.~\eqref{OptCond} into a least-squares formulation as follows:
\begin{equation}\label{OptCondApprox}
\mathbf{J}^{T} \mathbf{J} \hat{\gamma}_{\text{opt}} + \mathbf{J}^{T} \mathbf{N}=0,
\end{equation}
% Here, $\mathbf{J}$ denotes the Jacobian of the network output with respect to its parameters, while $\mathbf{N}$ represents the residual of the governing PDE evaluated at a set of collocation points. 
Here, $\mathbf{J}$ indicates the sensitivity matrix of the network prediction with respect to trainable parameters, whereas $\mathbf{N}$ denotes the residual values obtained by evaluating the governing equation at selected collocation nodes. The entries of these matrices are defined as follows:
% The elements of these matrices are given by:
\begin{equation}
\left( \mathbf{J} \right)_{ij} = \frac{\partial \hat{u}^i}{\partial {W}_j}, \quad \left( \mathbf{N} \right)_i = \mathcal{N}(\hat{u}^i),
\end{equation}
% where $i = 1, 2, \dots, N_{\hat{u}}$ indexes the collocation points and $j = 1, 2, \dots, N_{{W}}$ corresponds to the network parameters. 
where the index $i = 1, 2, \dots, N_{\hat{u}}$ refers to the evaluation locations used for enforcing the PDE, and $j = 1, 2, \dots, N_W$ labels the trainable parameters of the network.
The entries of both $\mathbf{J}$ and $\mathbf{N}$ are computed using automatic differentiation. After computing $\gamma_{\text{opt}}$, we update the network parameters using any selected numerical methods, such as forward Euler method:
\begin{equation}
    {W}(t_{n+1}) = {W}(t_n) + \gamma_{\text{opt}} \, \Delta t.
\end{equation}

We summarized the evolutionary \ac{KANs} with Gaussian \ac{RBFs} in Fig.~\ref{fig:EvoRadialKAN}. 
% The network evolves in time by updating its parameters along a direction $\gamma$ computed from the governing PDE, allowing it to capture the temporal dynamics of the solution. 
The network parameters are updated over time in the direction $\gamma$ derived from the governing PDE, enabling the model to reflect the time-dependent behavior of the solution.
Each evolved network state corresponds to a solution snapshot at a given time, and continued updates yield the full solution trajectory.

\section{Boundary condition-guaranteed evolutionary KAN with radial basis functions} \label{sec:Methodology}
% In this section, we present our proposed frameworks for handling Dirichlet, periodic, and Neumann boundary conditions within the evolutionary \ac{KAN} architecture using Gaussian \ac{RBFs}. 
This section introduces our methodology for incorporating Dirichlet, periodic, and Neumann boundary conditions into the evolutionary \ac{KAN} architecture with Gaussian \ac{RBFs}.
For each type of boundary condition, we introduce boundary condition-guaranteed networks to ensure accurate enforcement. Note that these three distinct approaches for different boundary conditions can be flexibly integrated to handle mixed boundary conditions, as demonstrated and validated through later numerical experiments in Sec.~\ref{sec:Experiments}.

% ============================
% subsection: Dirichilet Boundary Condition
% ============================
\subsection{Dirichlet Boundary Condition}
\label{subsec:Dirichlet}

A commonly used approach to strongly impose Dirichlet conditions within neural network formulations of PDE problems is the method proposed in \cite{guyiqi}.
This method incorporates auxiliary functions $h(\mathbf{x})$ and $l(\mathbf{x},t)$ into the network output $\hat{u}(\mathbf{x};W(t))$, producing $u(\mathbf{x};W(t))$ that satisfies the boundary conditions by construction, without relying on the particular configuration of $W(t)$:
\begin{equation}
\label{eq:Dirichlet_basic}
u(\mathbf{x};W(t)) \;=\; h(\mathbf{x}) \hat{u}\left(\mathbf{x};W(t) \right) \;+\; l(\mathbf{x},t).
\end{equation}
To illustrate Eq.~\eqref{eq:Dirichlet_basic}, we may consider a one-dimensional case, as the extension to higher dimensions is straightforward. 
Suppose the boundary values are specified as $u(k_1) = a$ and $u(k_2) = b$. Then, we construct the lifting function $l(x,t)$ to interpolate between these values as:
\begin{equation}
l(x,t) \;=\; \frac{(b - a)\,(x - k_1)}{k_2 - k_1} \;+\; a.
\end{equation}
Next, we require $h(x)$ to vanish at $x=k_1$ and $x=k_2$. A convenient polynomial choice with roots at $k_1$ and $k_2$ is
\begin{equation}
\label{eq:h_1 function}
h(x) \;=\; (x - k_1)^{p_1}\,(x - k_2)^{p_2}, 
\end{equation}
where $0 < p_1, p_1 \le 1$. In typical implementations, one sets $p_1 = p_2 = 1$ for simplicity.
However, the output shaping in Eq.~\eqref{eq:Dirichlet_basic} can reduce expressiveness, as the fixed function $h(x)$ constrains the solution space and causes vanishing or exploding gradients near boundaries, ultimately degrading the stability and performance of the neural network~\cite{marquez2017imposing}.

% ==== proposed approach

\begin{figure}[htp!]
    \centering
    \begin{subfigure}[b]{0.95\linewidth}
        \centering
        \includegraphics[width=\linewidth]{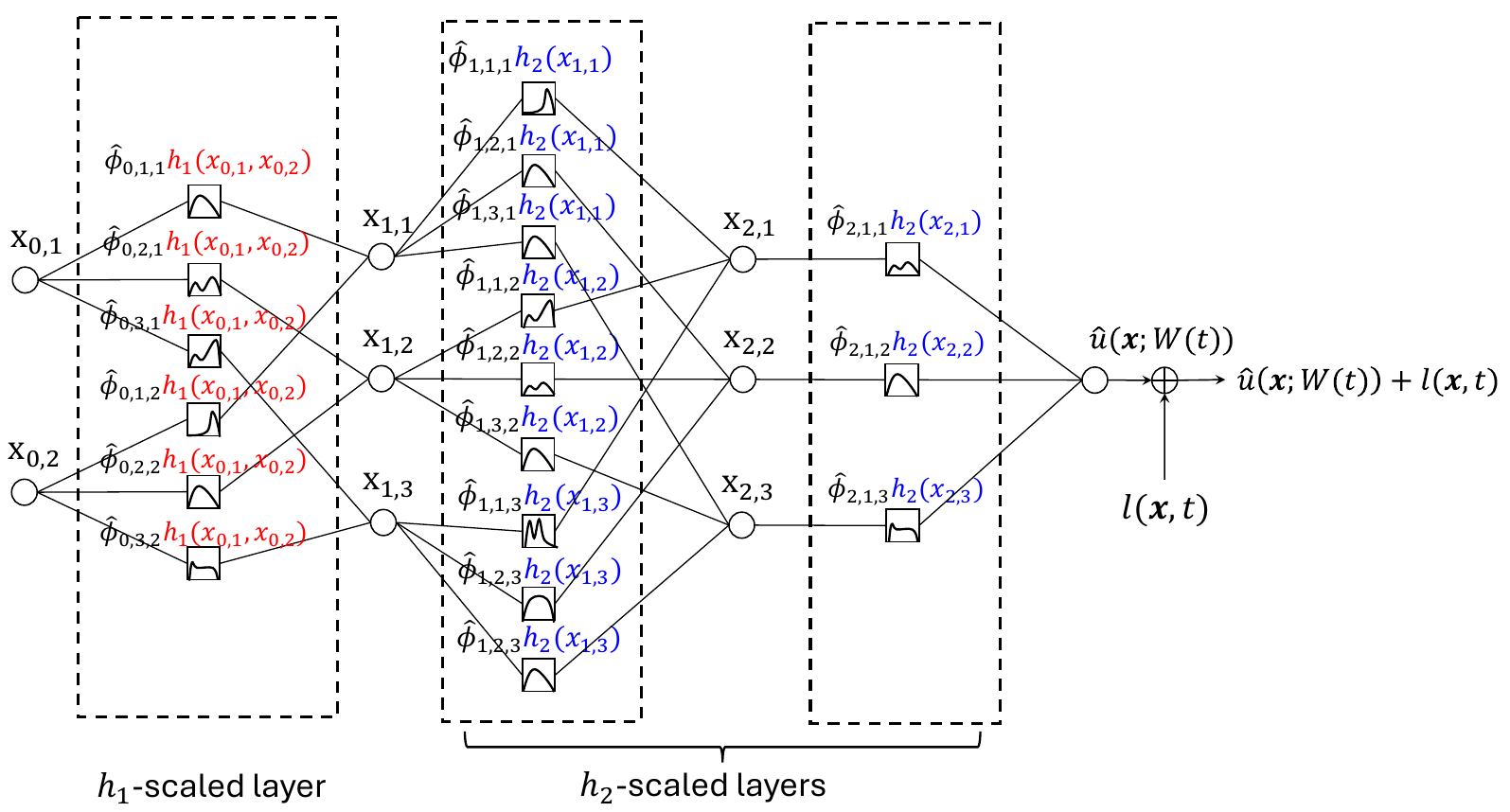}
        % \caption{}
    \end{subfigure}
    \caption{Boundary condition-guaranteed evolutionary KAN with radial basis functions: Dirichlet boundary condition.  
The illustrated $[2, 3, 3, 1]$ architecture depicted the use of $h_1$- and $h_2$-scaled activation layers, where $h_1$ ensures vanishing at the domain boundaries and $h_2$ maps zero inputs to zero outputs to enforce homogeneous conditions. For non-homogeneous boundary condition, a lifting function \( l(x,t) \) is employed.
 }
    \label{fig:dirichlet_bc}
\end{figure}

To effectively address this, we introduce a novel approach for enforcing the Dirichlet boundary condition. Unlike the method in Eq.~\eqref{eq:Dirichlet_basic}, which modifies the network output explicitly, our technique embeds the boundary condition directly into the basis functions of the network.
Instead of multiplying $h(\mathbf{x},t)$ with the final layer output, we incorporate the boundary information at the basis level. 
% The resulting network output is given by
The resulting network output takes the form:
\begin{equation}
\label{eq:Dirichlet_RadialKAN}
u(\mathbf{x}; W(t)) = \hat{u}(\mathbf{x}; W(t)) + l(\mathbf{x},t),
\end{equation}
where $l(\mathbf{x},t)$ is the time-dependent lifting function, and $\hat{u}(\mathbf{x}; W(t))$ represents the prediction produced by the RadialKAN architecture. To enforce the boundary condition within the network structure, we introduce two types of scaling functions, $h_1(\mathbf{x})$ and $h_2(x)$, which are applied directly to the basis functions of RadialKAN.
In the RadialKAN example with a $[2, 3, 3, 1]$ architecture shown in Fig.~\ref{fig:dirichlet_bc}, we illustrate the use of $h_1$- and $h_2$-scaled layers, where $h_1$ and $h_2$ are applied to the activation functions of their respective layers. The function $h_1$ ensures that the activation functions vanish at the domain boundaries, while $h_2$ maps zero inputs to zero outputs, thereby guaranteeing that the final network output also satisfies the zero boundary condition.
For the non-homogeneous case, the time-dependent lifting function $l(x,t)$ adjusts the solution to exactly satisfy the boundary conditions.

For general expression of the boundary condition-guaranteed RadialKAN, we formulate the Gaussian radial basis function $\hat{\phi}_{0,j,i}$ in the first hidden layer with $d$-dimensional problem as follows:
\begin{equation}
% \label{eq:phi_tensor}
    \tilde{\phi}_{0,j,i}(x_{0,i}) = h_1(\mathbf{x})\,\hat{\phi}_{0,j,i}(x_{0,i}), \quad i = 1, \dots, n_0, \quad j = 1, \dots, n_{1},
\end{equation}
where
\begin{equation}
h_1({\mathbf{x}}) = (x_{0,1} - k_1)^{p_1}(x_{0,2} - k_2)^{p_2} \cdots (x_{0.d} - k_d)^{p_d}  \quad \text{and} \quad {x} = (x_{0,1}, x_{0,2}, \dots, x_{0,d}).
\end{equation}
Next, for all subsequent hidden layers, each basis function $\hat{\phi}_{l,j,i}$ is modified as:
\begin{equation}
\tilde{\phi}_{l,j,i}(x_{l,i}) = h_2(x)\,\hat{\phi}_{l,j,i}(x), \quad l = 1, \dots, L-1, \quad i = 1, \dots, n_l, \quad j = 1, \dots, n_{l+1}.
\end{equation}
where
\begin{equation}
h_2(x) = x_{l,i}^{p_i},
\end{equation}
and the exponent satisfies $0 < p_i \leq 1$.
The layer scaled by \( h_1 \) enforces the basis functions of the network to be zero at the boundaries of the domain. The following layers, scaled by \( h_2 \), preserve this zero-value property since a zero input to these layers results in a zero output. This two step mechanism guarantees that the final network output \( \hat{u}(\mathbf{x};W(t)) \) remains zero on the boundary. For nonhomogeneous boundary conditions, this output is then modified by the lifting function \( l(\mathbf{x},t) \).

By replacing $\mathbf{\hat{\Phi}}_l$ with $\mathbf{\tilde{\Phi}}_l$, where $h_1(\mathbf{x})$ and $h_2(x)$ are respectively applied at each layer of the RadialKAN, we obtain the boundary condition-guaranteed RadialKAN as follows:
\begin{equation}\label{eq:RadialKAN}
    \text{RadialKAN}(\mathbf{x}) = \left(  \mathbf{\tilde{\Phi}}_{L-1} \circ \mathbf{\tilde{\Phi}}_{L-2} \circ \cdots \circ \mathbf{\tilde{\Phi}}_1 \circ \mathbf{\tilde{\Phi}}_0 \right)  \mathbf{x},
\end{equation}
which corresponds to $\hat{u}$ in Eq.~\eqref{eq:Dirichlet_RadialKAN}.
In case of $n_L = 1$, we define $f(\mathbf{x}) \equiv \text{RadialKAN}(\mathbf{x})$ and express the equation in a form analogous to Eq.~(\ref{eq:KART}):
\begin{equation}
    f(\mathbf{x}) = \sum_{i_{L-1}=1}^{n_{L-1}} h_2(x_{L-1,i_{L-1}})\hat{\phi}_{L-1,i_L,i_{L-1}} \left(\sum_{i_{L-2}=1}^{n_{L-2}} \cdots \left(\sum_{i_2=1}^{n_2} h_2(x_{2,i_2})\hat{\phi}_{2,i_3,i_2} \left(\sum_{i_1=1}^{n_1} \
    h_2(x_{1,i_1})\hat{\phi}_{1,i_2,i_1}\left(\sum_{i_0=1}^{n_0} h_1(\mathbf{x})\hat{\phi}_{0,i_1,i_0}(x_{i_0})\right)\right)\right)\cdots\right).
\end{equation}

% This construction allows us to strongly enforce the Dirichlet boundary conditions, i.e., $u(a) = a_0$ and $u(b) = b_0$, regardless of the values of the network parameters $\theta$. Furthermore, this approach mitigates the singularity and smoothness issues caused by directly multiplying $h(x)$ with the network output. The benefits of this method will be further illustrated and discussed in Sec.~\ref{sec:Experiments}.
% However, this conventional method may encounter regularity issues. In our numerical experiments, both EDNN and EvoKAN adopt this approach, but they fail to produce accurate solutions for certain complex problems.

% ============================
% subsection: Periodic Boundary Condition
% ============================
\subsection{Periodic Boundary Condition}
\label{subsec:periodic}

\begin{figure}[htp!]
    \centering
    \begin{subfigure}[b]{0.42\linewidth}
        \centering
        \includegraphics[width=\linewidth]{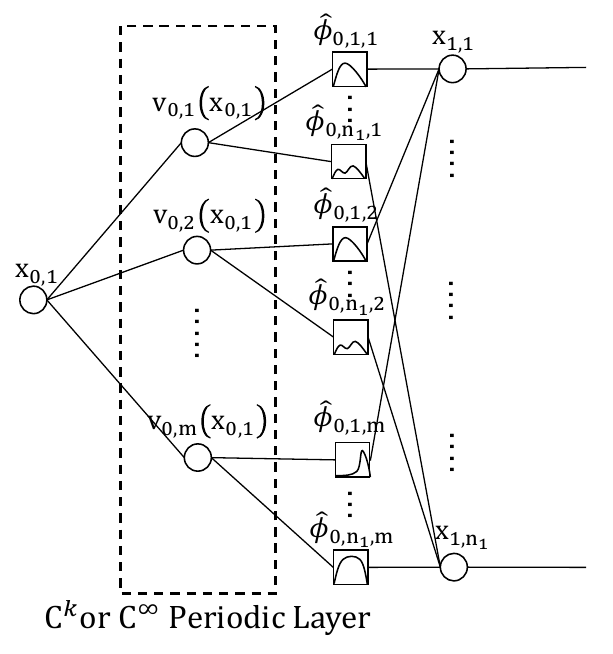}
        \caption{One-dimensional case}
    \end{subfigure}
    % \hfill
    \begin{subfigure}[b]{0.42\linewidth}
        \centering
        \includegraphics[width=\linewidth]{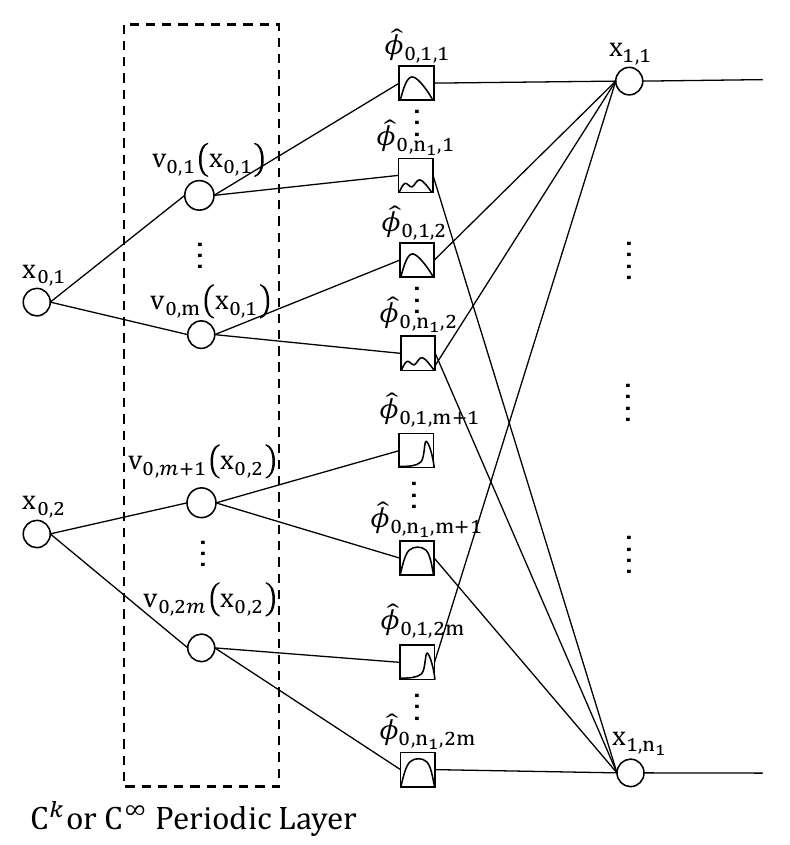}
        \caption{Two-dimensional case}
    \end{subfigure}
    \caption{Boundary condition-guaranteed evolutionary KAN with radial basis functions: Periodic boundary condition.}
    \label{fig:periodic_bc_radialkan}
\end{figure}

% To impose periodic boundary conditions exactly, we incorporate a periodic layer as proposed in~\cite{stevendong}.
% Consider a function $f(x)$ that is periodic with period $L$ on the real line:
% \begin{equation}\label{eq:f_global_period}
%     f(x+L)=f(x), \qquad \forall\,x\in\mathbb R .
% \end{equation}
% Restricting $f$ to a finite interval $[a, b]$ with length $L = b - a$ yields the desired boundary identities:
% \begin{equation}\label{eq:Cinf_conditions}
%     f^{(\ell)}(a)=f^{(\ell)}(b),\qquad \ell=0,1,2,\dots ,
% \end{equation}
% which we call the $C^{\infty}$ periodic conditions.  
% In practice one often needs periodicity only up to a finite derivative order
% $k\ge0$:
% \begin{equation}\label{eq:Ck_conditions}
%     f^{(\ell)}(a)=f^{(\ell)}(b),\qquad 0\le\ell\le k .
% \end{equation}
% We refer to Eq.~\eqref{eq:Ck_conditions} as the $C^{k}$ periodic conditions.
% Our objective is to build networks whose outputs automatically satisfy
% either Eq.~\eqref{eq:Cinf_conditions} or
% Eq.~\eqref{eq:Ck_conditions} for a finite $k$.

To rigorously impose periodic boundary conditions, we adopt a periodic layer formulation inspired by the methodology introduced in~\cite{stevendong}.  
Let \( f(x) \) be a function that exhibits periodic behavior over the entire real axis with a fixed period \( L \), satisfying
\begin{equation}\label{eq:f_global_period}
    f(x+L)=f(x), \qquad \forall\,x\in\mathbb R .
\end{equation}
By confining the domain to a bounded interval \( [a, b] \) such that the length \( L \) equals \( b - a \), the periodic property leads to the following identity at the boundaries:
\begin{equation}\label{eq:Cinf_conditions}
    f^{(\ell)}(a)=f^{(\ell)}(b),\qquad \ell=0,1,2,\dots ,
\end{equation}
Infinite-order periodicity is ensured by Eq.~\eqref{eq:Cinf_conditions}, which requires that the function and all of its derivatives match at the endpoints.  
In practical applications, matching a function and its derivatives up to the highest differential order of the governing PDE typically provides sufficient periodicity without requiring infinite differentiability.
This leads to a relaxed form of the periodicity condition, given by
\begin{equation}\label{eq:Ck_conditions}
f^{(\ell)}(a) = f^{(\ell)}(b),\qquad 0 \le \ell \le k,
\end{equation}
which defines periodicity of order $k$, corresponding to the highest differential order of the PDE.
Our objective is to construct RadialKAN models that naturally satisfy finite-order counterpart in Eq.~\eqref{eq:Ck_conditions}.

% Throughout this section we denote by $m_{\mathrm{PDE}}$ the highest spatial
% derivative order appearing in the governing \ac{PDEs}.
% Let $v(x)$ be any $C^{m_{\mathrm{PDE}}}$ periodic function
% satisfying \eqref{eq:Ck_conditions} with $k=m_{\mathrm{PDE}}$,
% and let $f(\cdot)$ be an arbitrary function.
% Defining the composition
% \begin{equation}
%    u(x)\;=\;f\bigl(v(x)\bigr) 
% \end{equation}
% one verifies, by chain-rule, that $u$ inherits the same
% $C^{m_{\mathrm{PDE}}}$ periodicity:
% \begin{equation}
%     u^{(\ell)}(a)=u^{(\ell)}(b),\qquad 0\le\ell\le m_{\mathrm{PDE}}.
% \end{equation}
% To impose periodic boundary conditions exactly, we incorporate a periodic layer as proposed in~\cite{stevendong}. According to Lemma~2.1 in~\cite{stevendong}, if $v(x)$ is a smooth periodic function with period $L$, i.e.,
% \begin{equation}
% \label{eq:periodic_function}
% v(x + L) = v(x), \quad \forall x \in (-\infty, \infty),
% \end{equation}
% and $f(x)$ is an arbitrary smooth function, then the composition $u(x) = f(v(x))$ also satisfies
% \begin{equation}
% \label{eq:periodicity}
% u(x + L) = u(x), \quad \text{and} \quad u^{(d)}(a) = u^{(d)}(b), \quad \forall d \in \mathbb{N}_0,
% \end{equation}
% where $b - a = L$. This result ensures that $u(x)$ inherits the periodicity of $v(x)$, making it suitable for constructing networks that satisfy periodic boundary conditions.
Building upon Eqs.~\eqref{eq:f_global_period} and~\eqref{eq:Ck_conditions}, we now construct periodic layers within the RadialKAN architecture, as illustrated in Fig.~\ref{fig:periodic_bc_radialkan}. In Fig.~\ref{fig:periodic_bc_radialkan}a, we present the integration of a periodic layer into a one-dimensional RadialKAN. The input $x_{0,1}$ is first mapped into a periodic representation using $m$ neurons denoted by $v_{0,1}, v_{0,2}, \dots, v_{0,m}$.
% %% ===modi.=== %%
% For the perioidc transformation
% %% ===modi.=== %%
For the periodic transformation, a natural and straightforward choice is the use of sinusoidal functions, resulting in the following mapping in the first layer:
\begin{equation}
x \mapsto \left( \sin(\omega x_{0,1}), \cos(\omega x_{0,1}), \sin(2\omega x_{0,1}), \cos(2\omega x_{0,1}), \dots \right),
\end{equation}
where $\omega = \frac{2\pi}{L}$. The number of sine and cosine components can be adjusted to match the frequency content and complexity of the target function. Alternatively, one may construct $v(x)$ using Hermite polynomials to form a $C^k$-smooth periodic layer, thereby enabling the relaxed enforcement of $C^k$ periodic boundary conditions.

We now describe the construction to the two-dimensional case, as shown in Fig.~\ref{fig:periodic_bc_radialkan}b. Let the spatial coordinates be ${\mathbf{x}} = (x_{0,1}, x_{0,2}) \in \mathbb{R}^2$, and suppose that both $x_{0,1}$ and $x_{0,2}$ are periodic with periods $L_1$ and $L_2$, respectively. Following the strategy for the one-dimensional case, we construct a periodic layer by applying periodic mappings to each spatial dimension independently.
Specifically, we define:
\begin{equation}
\begin{aligned}
\mathbf{x} \mapsto \big(&
\sin(\omega_1 x_{0,1}), \cos(\omega_1 x_{0,1}), \dots, \sin(K_1 \omega_1 x_{0,1}), \cos(K_1 \omega_1 x_{0,1}), \\
& \sin(\omega_2 x_{0,2}), \cos(\omega_2 x_{0,2}), \dots, \sin(K_2 \omega_2 x_{0,2}), \cos(K_2 \omega_2 x_{0,2})
\big),
\end{aligned}
\end{equation}
where $\omega_1 = \frac{2\pi}{L_1}$, $\omega_2 = \frac{2\pi}{L_2}$, and $K_1,K_2\in\mathbb{Z}_{\ge 1}$ denote the numbers of Fourier harmonics
in the $x_{0,1}$- and $x_{0,2}$-directions, respectively. 
To wit, for each $k\in\{1,\dots,K_1\}$ we embed
\(\sin(k\omega_1 x_{0,1}),\ \cos(k\omega_1 x_{0,1})\),
and analogously for the second coordinate with $K_2$.
This transformation yields a higher-dimensional embedding that captures the periodic structure in both spatial variables.

The constructed feature vector is then processed by the subsequent radial basis function layers within the RadialKAN architecture.
By Lemma~2.1 in \cite{stevendong} and the smoothness of sine and cosine functions, this construction guarantees that the output function $u(\mathbf{x})$ satisfies
\begin{equation}
u(\mathbf{x} + L_i e_i) = u(\mathbf{x}), \quad \text{and} \quad \partial^{\alpha} u({\mathbf{x}}) = \partial^{\alpha} u({\mathbf{x}} + L_i {e}_i), \quad \forall \alpha \in \mathbb{N}_0^2,
\end{equation}
for $i = 1, 2$, where $e_i$ is the standard basis vector in $\mathbb{R}^2$.

\subsection{Neumann Boundary Condition}
\label{subsec:Neumann}
\begin{figure}[htp!]
    \centering
    \begin{subfigure}[b]{0.95\linewidth}
        \centering
        \includegraphics[width=\linewidth]{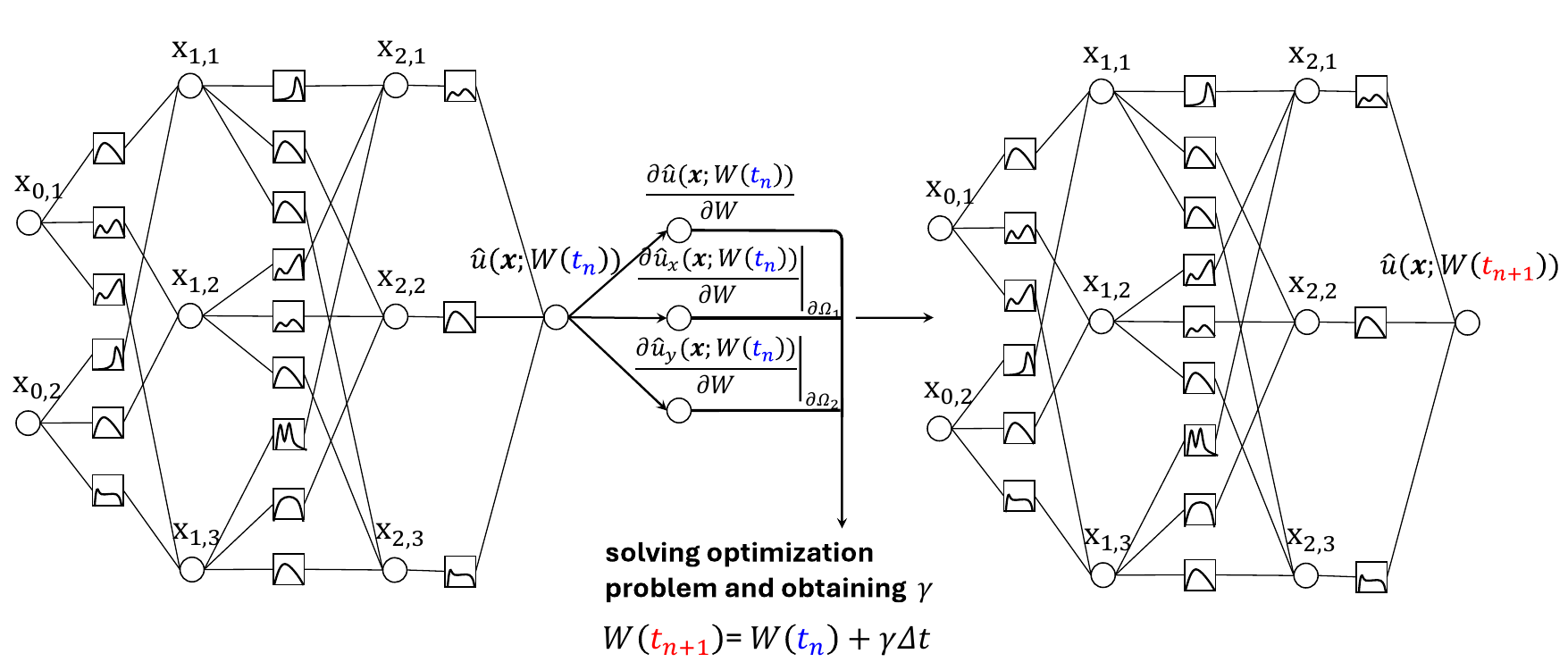}
    \end{subfigure}
    \caption{Boundary condition-guaranteed evolutionary KAN with radial basis functions: Neumann boundary condition.}
    \label{fig:Neumann_bc_radialkan}
\end{figure}

We now address the Neumann boundary condition in solving \ac{PDEs} by developing the evolutionary network described in Sec.~\ref{EvoKAN_RBFs}.  
To ensure compliance with Neumann boundary conditions, our approach introduces boundary information directly into the evolutionary update process.
%by reformulating a least-squares problem.  
%We illustrate the 2-D, time-dependent %boundary flux vector:
%\begin{equation}\label{Eq:gVec}
%g(\mathbf{x},t)
%=
%\begin{pmatrix}
%a(x,t)\\[2pt]
%b(y,t)
%\end{pmatrix},
%\qquad
%\mathbf{x}=(x,y).
%\end{equation}
Consider the Neumann boundary condition expressed as:
\begin{equation}
\label{Eq:NeumannBC}
\nabla u \cdot \vec{n} = g(\mathbf{x},t).
\end{equation}
where $\vec{n}$ is the unit outward normal vector at the boundary point $\mathbf{x}$. Taking the time derivative of both sides in Eq.~\eqref{Eq:NeumannBC} yields
\begin{equation}
\label{Eq:Diff_NeumannBC}
\frac{\partial}{\partial t}(\nabla u \cdot \vec{n} - g(\mathbf{x},t)) = 0.
\end{equation}
On axis-aligned boundary segments, $g(\mathbf{x}, t)$ can be given componentwise. 
For example, on boundaries with normals aligned with $\mathbf{e}_x$ or $\mathbf{e}_y$, it can be written as
\begin{equation}
\label{Eq:NeumannBC_axis}
\frac{\partial u}{\partial x} = a(\mathbf{x}, t),
\qquad
\frac{\partial u}{\partial y} = b(\mathbf{x}, t)
\end{equation}
where $a(\mathbf{x}, t)$ and $b(\mathbf{x}, t)$ are scalar-valued fluxes in the $x$ and $y$ directions, respectively.
Our objective is to embed the Neumann boundary condition expressed in Eq.~\eqref{Eq:Diff_NeumannBC} into the optimization framework defined by Eq.~\eqref{Wmin}. 
% To achieve this, we evaluate, at each discrete time step, not only the Jacobian of the network output with respect to the parameters, $\frac{\partial \hat{u}(x; \mathbf{W}(t_n))}{\partial \mathbf{W}}$, but also the directional derivatives of the output along the boundary components. 
To achieve this, we compute at each time instance both the derivatives of the network prediction with respect to the trainable weights, $\frac{\partial \hat{u}(\mathbf{x}; \mathbf{W}(t_n))}{\partial \mathbf{W}}$, and the directional derivatives of the output along the boundary segments.
Specifically, we compute $\left.\frac{\partial \hat{u}_x(\mathbf{x}; \mathbf{W}(t_n))}{\partial \mathbf{W}}\right|_{\partial \Omega_1}$ and $\left.\frac{\partial \hat{u}_y(\mathbf{x}; \mathbf{W}(t_n))}{\partial \mathbf{W}}\right|_{\partial \Omega_2}$. These terms are illustrated in Fig.~\ref{fig:Neumann_bc_radialkan}, and collectively form an augmented system that allows the network to account for Neumann-type constraints during parameter evolution.
 These three Jacobian matrices are then flattened and concatenated to form a multi-objective optimization problem as follows:
\begin{equation}\label{Wmin_2}
\frac{\partial {W}}{\partial t} = \arg\min_{\gamma} \mathcal{J}(\gamma), \quad \mathcal{J}(\gamma) = \frac{1}{2} \int_{X} \left\Vert \begin{pmatrix}
\frac{\partial \hat{u}(\mathbf{x}; W(t_n))}{\partial {W}} \\[6pt]
\left.\frac{\partial \hat{u}_x(\mathbf{x}; {W}(t_n))}{\partial {W}}\right|_{\partial \Omega_1} \\[6pt]
\left.\frac{\partial \hat{u}_y(\mathbf{x}; {W}(t_n))}{\partial {W}}\right|_{\partial \Omega_2}
\end{pmatrix}  \gamma + \begin{pmatrix}
\frac{r^{n+1}}{\sqrt{E(\hat{u}(\mathbf{x}; {W}(t_n)))}} \mathcal{N}(\hat{u}(\mathbf{x}; {W}(t_n))) \\[6pt]
-a_t(\mathbf{x},t) \\[6pt]
-b_t(\mathbf{x},t)
\end{pmatrix} \right\Vert_2^2 \, \mathrm{d}\mathbf{x},
\end{equation} 
where $a_t(\mathbf{x},t)=\partial a(\mathbf{x},t)/\partial t$ denotes the time derivative of the prescribed Neumann data, and the subscripts indicate the boundary segment on which this value is evaluated: $a_t := a_t(\mathbf{x},t)\big|_{\partial\Omega_1}$ on the faces whose outward normal is parallel to $e_x$, and $b_t := b_t(\mathbf{x},t)\big|_{\partial\Omega_2}$ on the faces whose outward normal is parallel to $e_y$.
Applying the first-order optimality condition, the optimal solution to Eq.~\eqref{Wmin_2} is obtained by solving the following linear system:
\begin{equation}\label{OptCond_3}
\nabla_{\gamma} \mathcal{J}(\gamma_{\text{opt}}) = 
\int_{X} 
\begin{pmatrix}
\frac{\partial \hat{u}(\mathbf{x}; W(t_n))}{\partial {W}} \\[6pt]
\left.\frac{\partial \hat{u}_x(\mathbf{x}; {W}(t_n))}{\partial {W}}\right|_{\partial \Omega_1} \\[6pt]
\left.\frac{\partial \hat{u}_y(\mathbf{x}; {W}(t_n))}{\partial {W}}\right|_{\partial \Omega_2}
\end{pmatrix}^T
\left(
\begin{pmatrix}
\frac{\partial \hat{u}(\mathbf{x}; W(t_n))}{\partial {W}} \\[6pt]
\left.\frac{\partial \hat{u}_x(\mathbf{x}; {W}(t_n))}{\partial {W}}\right|_{\partial \Omega_1} \\[6pt]
\left.\frac{\partial \hat{u}_y(\mathbf{x}; {W}(t_n))}{\partial {W}}\right|_{\partial \Omega_2}
\end{pmatrix}  \gamma_{\text{opt}} + \begin{pmatrix}
\frac{r^{n+1}}{\sqrt{E(\hat{u}(\mathbf{x}; {W}(t_n))}} \mathcal{N}(\hat{u}(\mathbf{x}; {W}(t_n))) \\[6pt]
-a_t(\mathbf{x},t) \\[6pt]
-b_t(\mathbf{x},t)
\end{pmatrix} \right)\mathrm{d}\mathbf{x} = 0.
\end{equation}

To compute an approximate solution $\gamma_{\text{opt}}$ to Eq.~\eqref{OptCond_3}, we recast it as the following least-squares problem:
\begin{equation}\label{OptCondApprox_Neumann}
\mathbf{J}^{T} \mathbf{J} \hat{\gamma}_{\text{opt}} + \mathbf{J}^{T} \mathbf{N}=0.
\end{equation}
% Here, $\mathbf{J}$ denotes the sensitivity matrix comprising partial derivatives of the neural approximation with respect to the learnable weights, and $\mathbf{N}$ is the vector of residuals computed by applying the differential operator to the network output at selected collocation locations.
Here, \( \mathbf{J} \) is the augmented Jacobian matrix formed by vertically concatenating the sensitivity matrices from the PDE residual points, the boundary points for the \( x \)-derivative, and the boundary points for the \( y \)-derivative. Similarly, \( \mathbf{N} \) is the concatenated vector of the corresponding PDE and time-differentiated boundary condition residuals. This formulation allows us to solve for the optimal parameter update \( \gamma_{\text{opt}} \) that simultaneously minimizes the errors in both the governing equation and the Neumann boundary conditions.
The components of these matrices are defined as
\begin{equation}
\left( \mathbf{J} \right)_{ij} = \frac{\partial \hat{u}^i}{\partial {W}_j}, \quad \left( \mathbf{N} \right)_i = \mathcal{N}(\hat{u}^i),
\end{equation}
where the indices $i = 1, 2, \dots, N_{\hat{u}}$ correspond to the locations where the residual is enforced, while $j = 1, 2, \dots, N_W$ label the entries of the trainable parameter set.
 All derivatives are computed via automatic differentiation.
Once $\gamma_{\text{opt}}$ is obtained, the parameters are advanced in time using the forward Euler update rule:
\begin{equation}
{W}(t_{n+1}) = {W}(t_n) + \gamma_{\text{opt}} \, \Delta t.
\end{equation}

\begin{remark}
% It is important to note that incorporating the time-differentiated Neumann condition in Eq. \ref{Eq:Diff_NeumannBC} into the optimization framework does not alter the solution of the original boundary value problem. To see this, observe that any solution $u(x, t)$ satisfying Eq. \ref{Eq:Diff_NeumannBC} must also satisfy Eq. \ref{Eq:NeumannBC} up to an integration constant in time. More precisely, integrating Eq. \ref{Eq:Diff_NeumannBC} with respect to $t$ yields
% $$ \nabla u \cdot \vec{n} = g(\mathbf{x},t) + k(\mathbf{x}), $$
% where $k(\mathbf{x})$ is a function independent of time. The presence of $k(\mathbf{x})$ introduces a discrepancy from the original Neumann condition. However, in the context of the full evolutionary PDE problem, the initial condition 
% $$ u(\mathbf{x}, 0) = u_0(\mathbf{x}) $$
% is also enforced. This initial condition, combined with the PDE dynamics, allows one to uniquesly determine $k(\mathbf{x})$ such that Eq. \ref{Eq:Diff_NeumannBC} reduces exactly to Eq. \ref{Eq:NeumannBC}. Therefore, it will be equivalent to solving the original boundary value problem. This justifies the correctness of our formulation, where the evolutionary network simultaneously enforces both the governing PDE and the differentiated Neumann condition.
Incorporating the time-differentiated Neumann condition in Eq.~\ref{Eq:Diff_NeumannBC} into the optimization framework does not alter the solution of the original boundary value problem. Any solution $u(\mathbf{x}, t)$ that satisfies Eq.~\ref{Eq:Diff_NeumannBC} necessarily satisfies Eq.~\ref{Eq:NeumannBC} up to an integration constant in time.
Integrating Eq.~\ref{Eq:Diff_NeumannBC} with respect to $t$ yields
\begin{equation}
\label{eq:integration_by_time}
\nabla u(\mathbf{x}, t) \cdot \vec{n} = g(\mathbf{x}, t) + \nabla u(\mathbf{x}, 0) \cdot \vec{n} - g(\mathbf{x}, 0),
\end{equation}
where the term $\nabla u(\mathbf{x}, 0) \cdot \vec{n} - g(\mathbf{x}, 0)$ is independent of time and introduces a discrepancy from the original Neumann boundary condition in Eq.~\eqref{Eq:NeumannBC}.
However, in the context of the full evolutionary PDE problem, we impose the initial condition
\[
u(\mathbf{x}, 0) = u_0(\mathbf{x}),
\]
where $u_0(\mathbf{x})$ is prescribed to satisfy $\nabla u_0(\mathbf{x}) \cdot \vec{n} = g(\mathbf{x}, 0)$.
This ensures that the discrepancy term in Eq.~\eqref{eq:integration_by_time} vanishes, i.e.,
\[
\nabla u(\mathbf{x}, 0) \cdot \vec{n} - g(\mathbf{x}, 0) = 0,
\]
so that Eq.~\ref{Eq:Diff_NeumannBC} reduces to Eq.~\ref{Eq:NeumannBC}. Therefore, enforcing the differentiated Neumann condition in the optimization framework is equivalent to solving the original boundary value problem.
This justifies the correctness of our formulation, in which the evolutionary network simultaneously enforces both the governing PDE and the time-differentiated Neumann condition.
\end{remark}

\section{Numerical Experiments} \label{sec:Experiments}

This section compares the performance of the introduced BEKAN method against three approaches: \ac{EDNN}, \ac{EvoKAN}, and the vanilla \ac{PINN}. 
In the numerical experiments, BEKAN, \ac{EDNN}, and \ac{EvoKAN} adopt the same enforcement strategies for periodic and Neumann boundary conditions discussed in Sec~\ref{subsec:periodic} and Sec~\ref{subsec:Neumann}, respectively. 
For the Dirichlet boundary condition, however, \ac{EDNN} and \ac{EvoKAN} utilize the output shaping method proposed in~\cite{guyiqi}, 
whereas BEKAN employs the approach described in Sec.~\ref{subsec:Dirichlet}. 
For all evolutionary models (BEKAN, EvoKAN, EDNN), we employ the energy-dissipative\ac{SAV} scheme to ensure stable time integration, as detailed in~\cite{shen2018scalar,shen2019new}.
The goal of this study is to evaluate the effectiveness of the proposed BEKAN relative to \ac{EDNN}, \ac{EvoKAN}, and the vanilla \ac{PINN}.

% ============================
% subsection: Allen-Cahn Equation
% ============================
\subsection{1D Allen-Cahn equation with Dirichlet Boundary Condition}

The Allen–Cahn equation is a representative reaction–diffusion model that is widely utilized in modeling phase separation and interface evolution phenomena in materials science. In one spatial dimension, it is expressed as
\begin{equation}
\label{eq:Allen-Cahn}
\frac{\partial u}{\partial t} = \frac{\partial^2 u}{\partial x^2} - g(u),
\end{equation}
under the initial and boundary conditions
\begin{equation}
\label{eq:Allen-Cahn_BC}
u(x, 0) = a \sin(\pi x), \quad u(-1) = u(1) = 0.
\end{equation}
Here, the nonlinear term \( g(u) = \frac{1}{\epsilon^2}u(u^2 - 1) \) arises as the derivative of a double-well potential. The parameter \( \epsilon \) controls the width of the interfacial region.
To generate a sharp interface and drive the steady-state solution toward a nearly binary profile, we set \( a = 0.08 \) and \( \epsilon = 0.002 \).
The Allen--Cahn dynamics can be characterized by the following Ginzburg--Landau energy functional:
\begin{equation}
E[u] = \int_{-1}^1 \left( \frac{1}{2} |u_x|^2 + G(u) \right) dx,
\end{equation}
where the potential energy density is given by \( G(u) = \frac{1}{4\epsilon^2}(u^2 - 1)^2 \) with the relation \( g(u) = G'(u) \).

Table~\ref{table:Allen-cahn_training_setting} outlines the training configuration.  
Both BEKAN and \ac{EvoKAN} share the same hidden layer architecture.
However, \ac{EvoKAN} employs B-splines, which introduce additional spline scalers, leading to a total of 330 trainable parameters, compared to 195 in BEKAN.
% A time step of $t = \SI{1e-6}{}$ is employed for the evolutionary training methods, while the vanilla \ac{PINN} is trained in a non-evolutionary manner to capture the solution over the entire time interval.
The evolutionary models are trained with a temporal resolution of $t = \SI{1e-6}{}$, whereas the vanilla \ac{PINN} is optimized in a static setting to learn the solution over the entire time interval.

\begin{figure}[htbp!]
    \centering
    \includegraphics[width=0.45\linewidth]{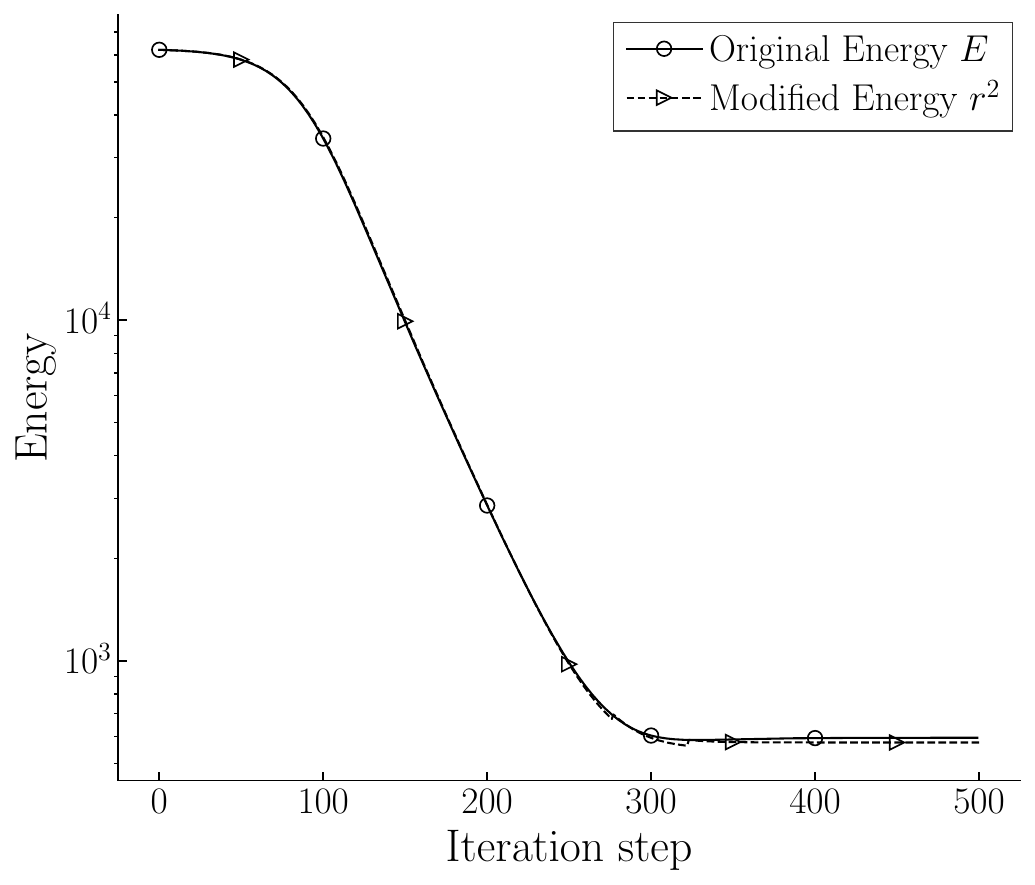}
    \caption{
Evolution of the original energy \( E \) and the modified energy \( r^2\) during the training process of BEKAN for the 1D Allen–Cahn equation (Eq.~\eqref{eq:Allen-Cahn}).  
Each iteration corresponds to one forward step in the time integration with a time increment \( \Delta t = 1.0 \times 10^{-6} \).  
The formulation is constructed such that the modified energy \( r^2 \) converges to the original energy \( E \), thereby ensuring stable energy evolution throughout the training.
}
    \label{fig:energy}
\end{figure}

In Fig.~\ref{fig:energy}, we plot the original energy $E$ and the modified energy $r^2$ during the training process of BEKAN. Each iteration corresponds to a single forward step in the numerical integration of the PDE, with a time increment of $\Delta t = 1.0 \times 10^{-6}$. 
The formulation is designed so that the adjusted energy term $r^2$ gradually converges to the original energy functional $E$, guaranteeing the stability of the energy evolution.

\begin{table} [htbp!]
\footnotesize
	\renewcommand{\arraystretch}{1.0}
	\begin{center} 
		\caption{Training configuration for the 1D Allen--Cahn equation (Eq.~\eqref{eq:Allen-Cahn}).
}
		\begin{tabular}{l c c c c}
			\hline
			{\, \, \, } & {BEKAN} & {EvoKAN} & {EDNN} & {Vanilla PINN} \\				
			\hline
			{Hidden layers} & {[3, 3, 3, 3]} & {[3, 3, 3, 3]} & {[15, 15, 15]} & {[15, 15, 15]}\\
            % {} & {} & {} & {} & {[90, 90, 90, 90]}\\
            % {} & {[7, 7, 7, 7]} & \makecell[c]{[7, 7, 7, 7]} & {[20, 20, 20, 20]} & {[20, 20, 20, 20]}\\
            {Activation functions} & {Gaussian RBFs/SiLU} & {B-splines/SiLU} & {tanh} & {tanh} \\
            % \hline
            \makecell[l]{Grid points number \\ of activation functions} & {5} & {5} & {-} & {-}\\
		\makecell[l]{Number of \\trainable parameters} & {195} & {330} & {526} & {526}\\
            {Optimizer} & {Adam} & {Adam} & {Adam} & {Adam/L-BFGS-B}\\
            {Timestep} & {1e-06} & {1e-06} & {1e-06} & {-}\\
			\hline
		\end{tabular}
        \label{table:Allen-cahn_training_setting}
	\end{center}
\end{table}

\begin{figure}[htbp]
    \centering
    \begin{subfigure}[b]{0.24\linewidth}
        \centering
        \includegraphics[width=\linewidth]{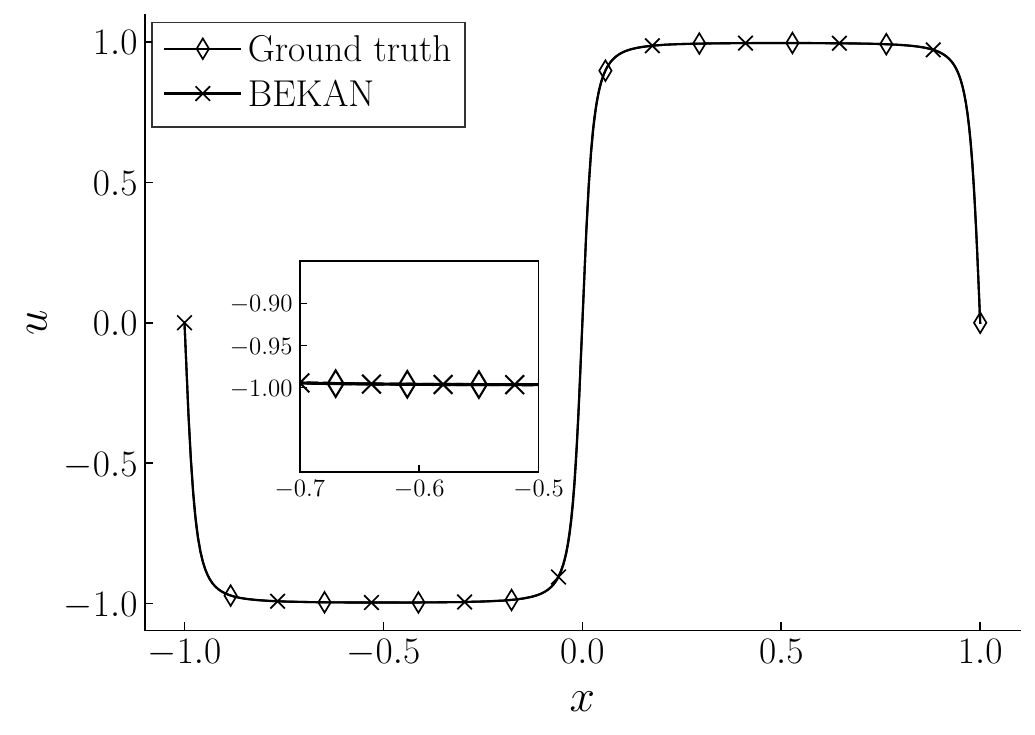}
        \caption{BEKAN solution}
    \end{subfigure}
    % \hfill
    \begin{subfigure}[b]{0.24\linewidth}
        \centering
        \includegraphics[width=\linewidth]{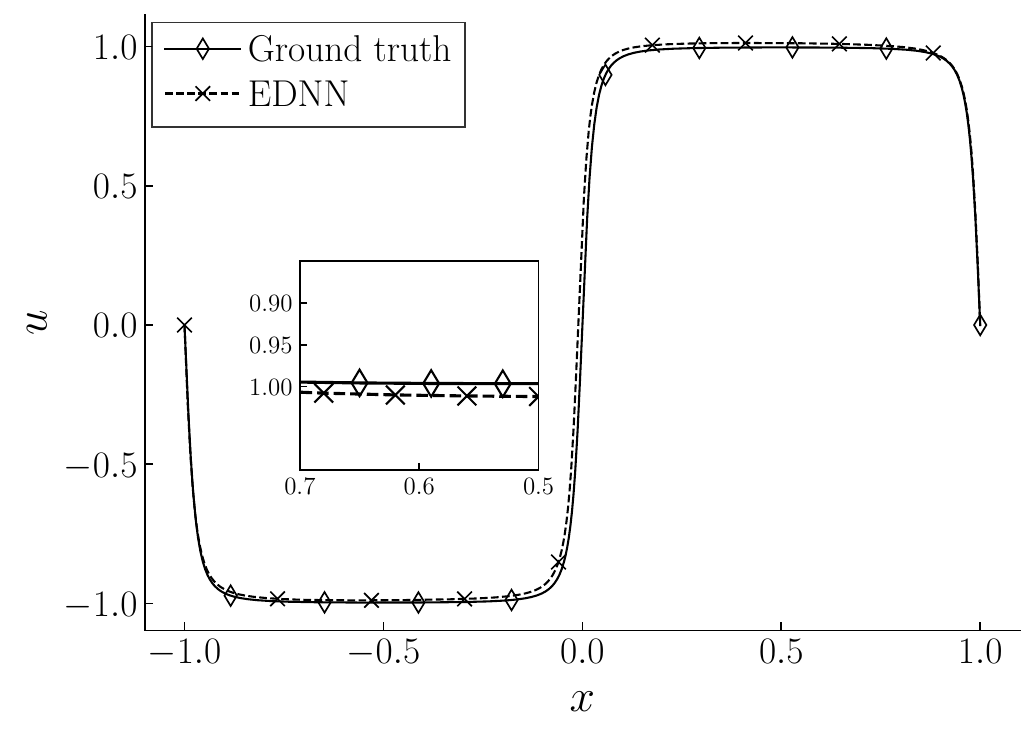}
        \caption{EDNN solution}
    \end{subfigure} 
    % \hfill
    \begin{subfigure}[b]{0.24\linewidth}
        \centering
        \includegraphics[width=\linewidth]{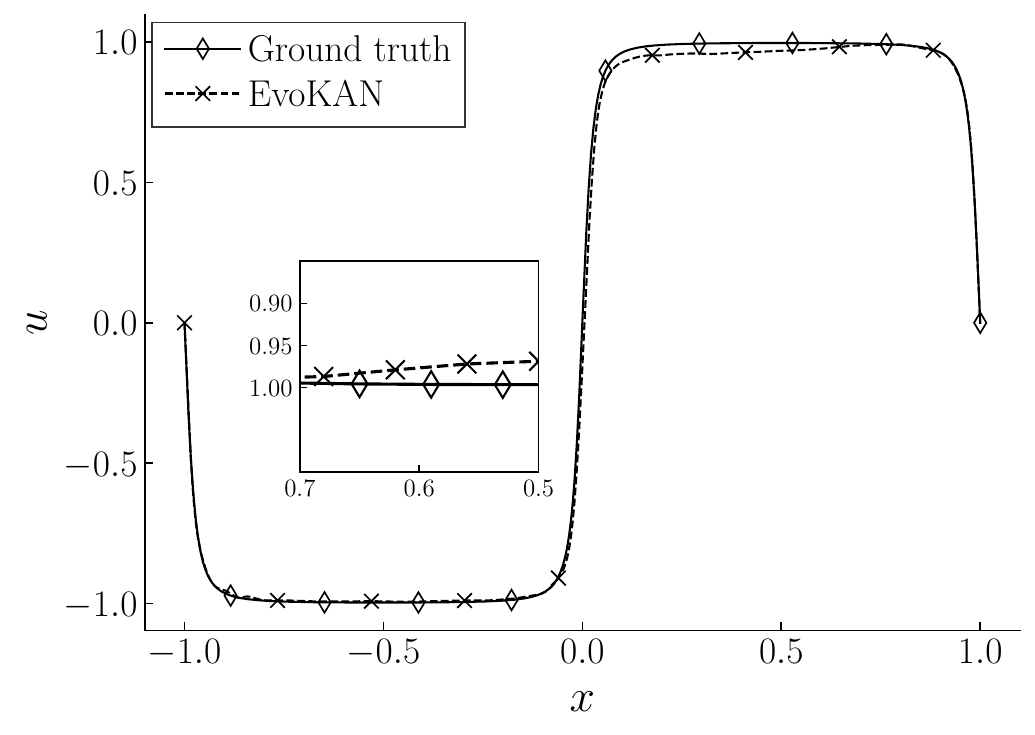}
        \caption{EvoKAN solution}
    \end{subfigure}
    \begin{subfigure}[b]{0.24\linewidth}
        \centering
        \includegraphics[width=\linewidth]{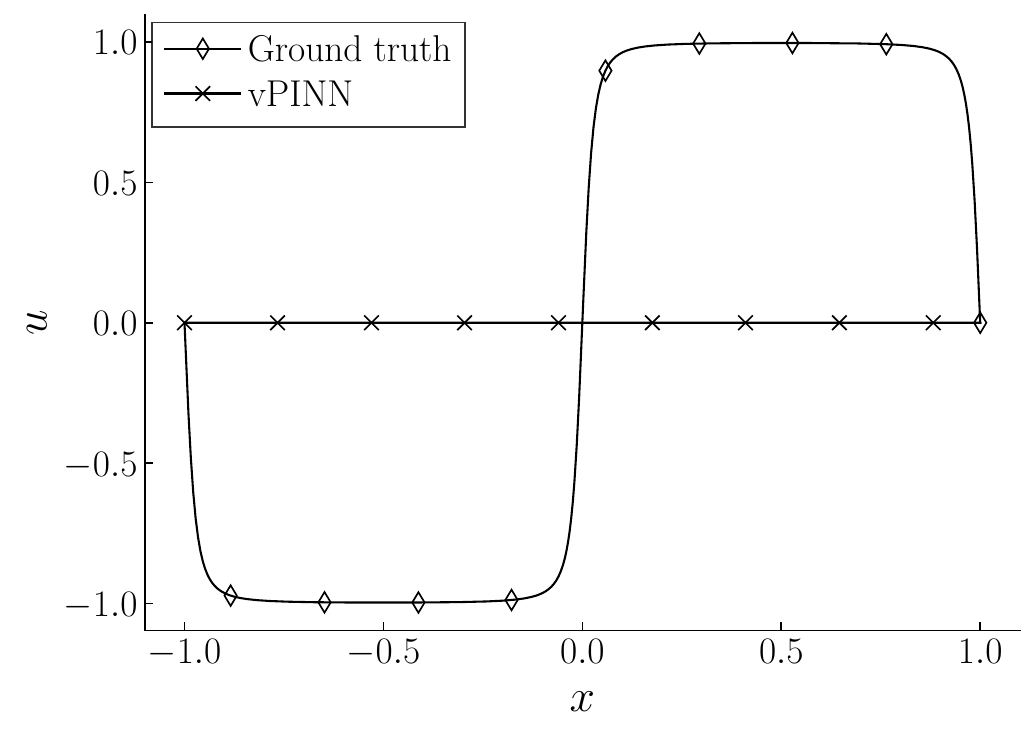}
        \caption{vPINN solution}
    \end{subfigure}
    \\
    
    \begin{subfigure}[b]{0.24\linewidth}
        \centering
        \includegraphics[width=\linewidth]{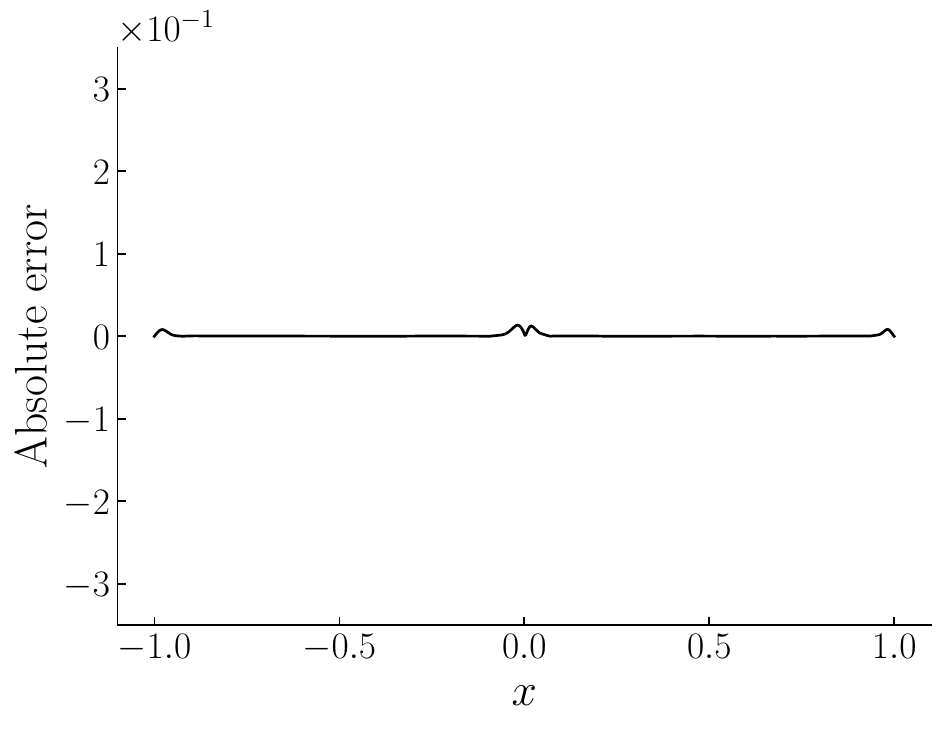}
        \caption{BEKAN error}
    \end{subfigure}
    % \hfill
    \begin{subfigure}[b]{0.24\linewidth}
        \centering
        \includegraphics[width=\linewidth]{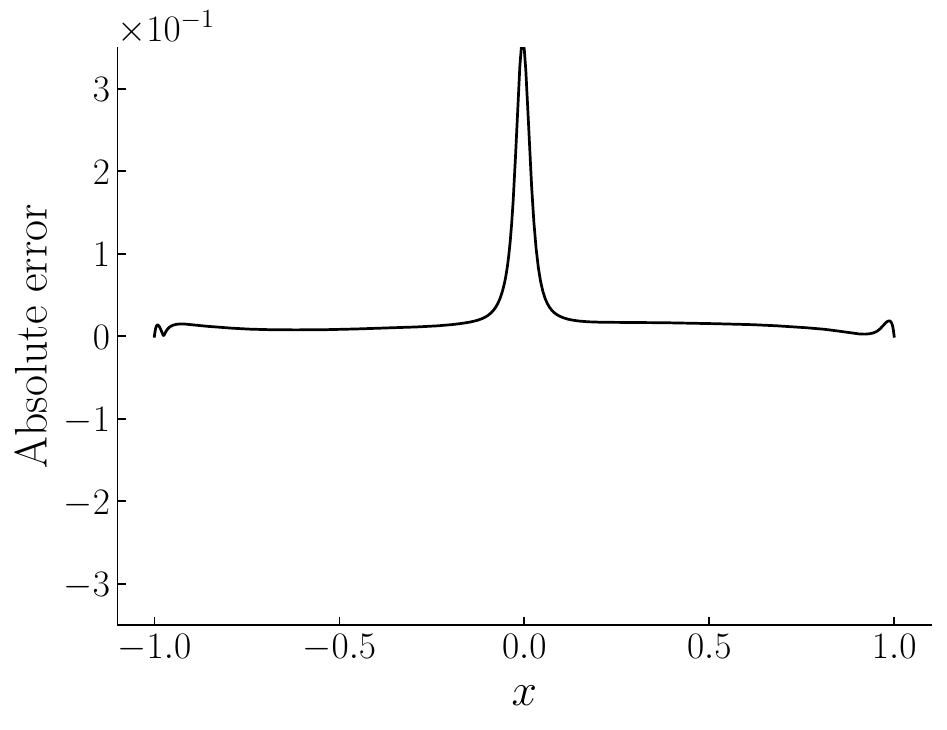}
        \caption{EDNN error}
    \end{subfigure} 
    % \hfill
    \begin{subfigure}[b]{0.24\linewidth}
        \centering
        \includegraphics[width=\linewidth]{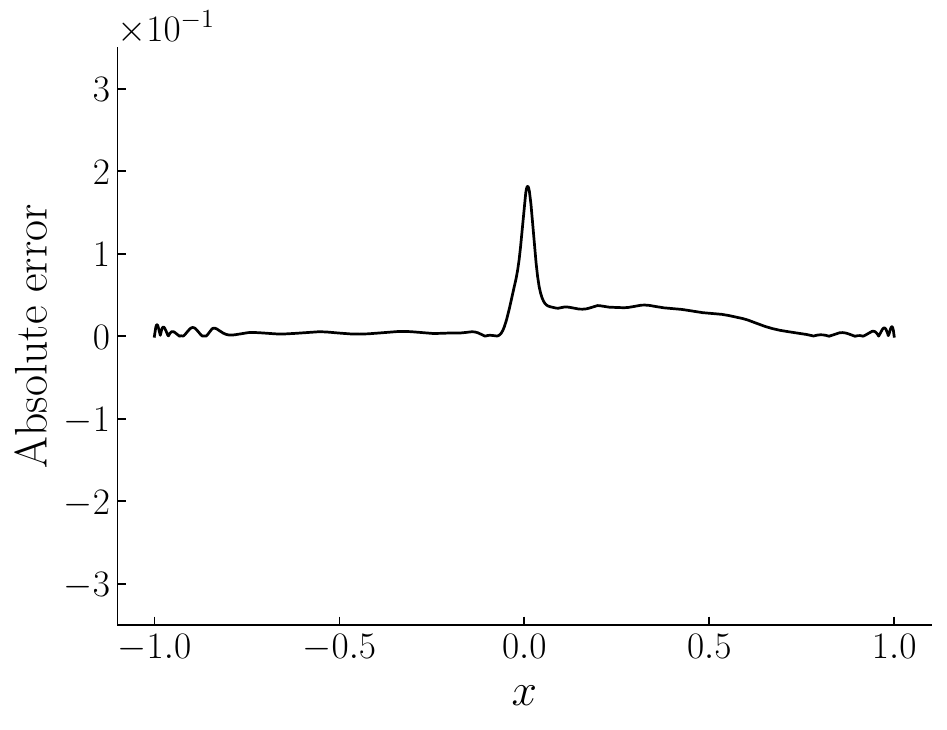}
        \caption{EvoKAN error}
    \end{subfigure}
    \begin{subfigure}[b]{0.24\linewidth}
        \centering
        \includegraphics[width=\linewidth]{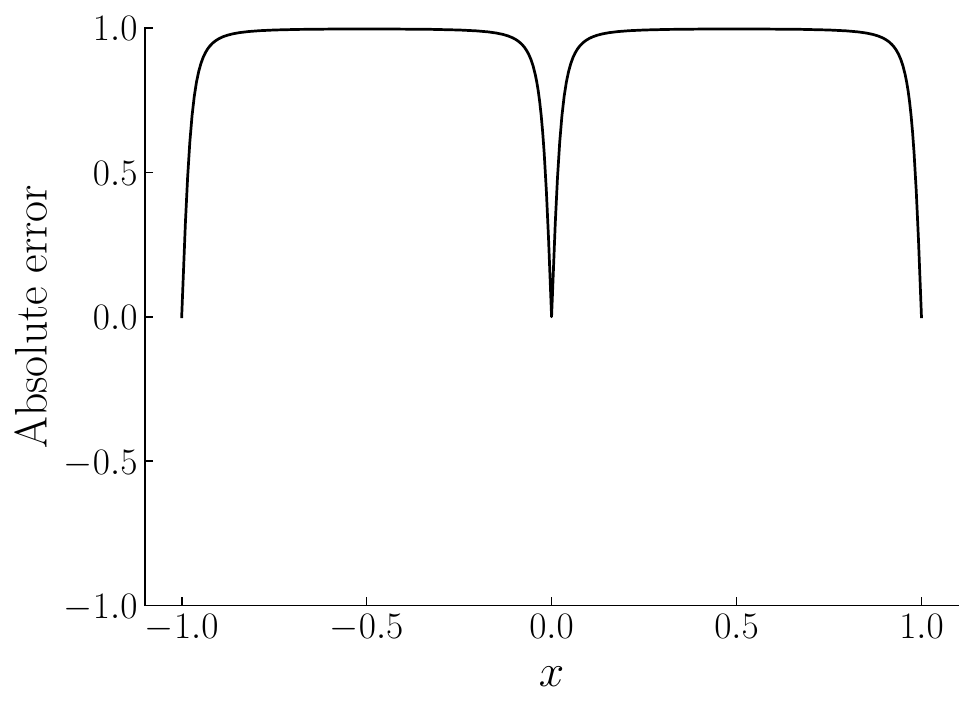}
        \caption{vPINN error}
    \end{subfigure}
    \caption{
Comparison of BEKAN, EvoKAN, EDNN, and vanilla PINN in solving the 1D Allen–Cahn equation (Eq.~\eqref{eq:Allen-Cahn}) at $t = \SI{2e-5}{}$\,s. 
Figures (a)–(d) show the predicted solutions, while (e)–(h) illustrate the corresponding absolute errors, measured against the reference solution.
BEKAN exhibits the smallest absolute difference, demonstrating the best performance among the four models.
}
    \label{fig:allen_cahn_bc_t_2}
\end{figure}

\begin{figure}[htbp]
    \centering
    \begin{subfigure}[b]{0.24\linewidth}
        \centering
        \includegraphics[width=\linewidth]{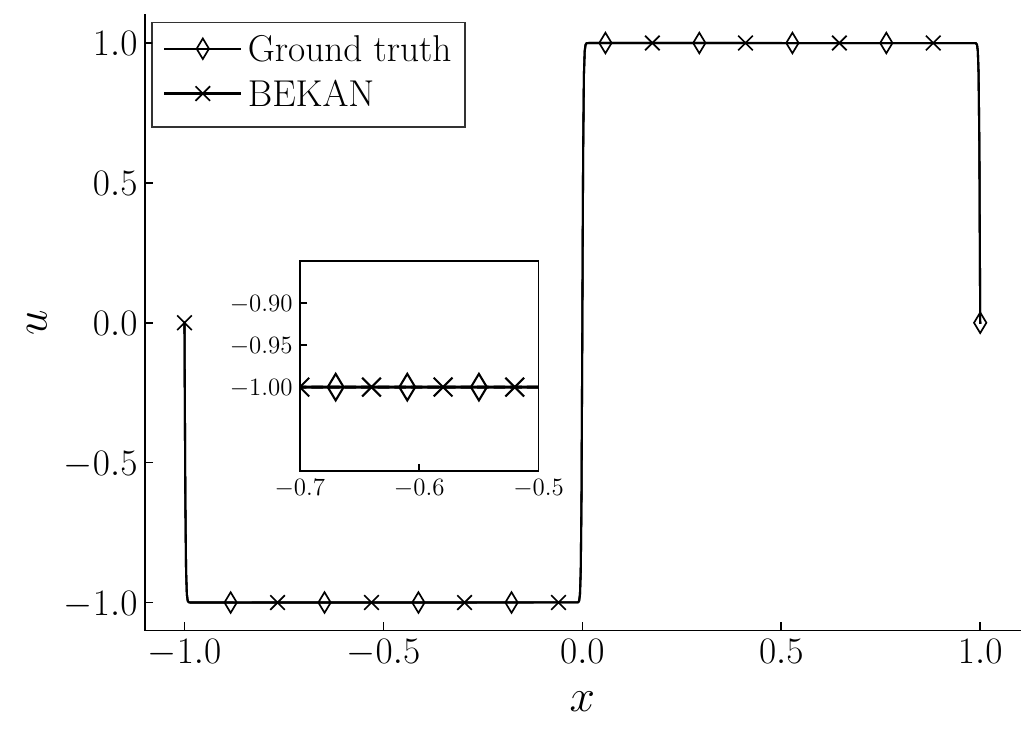}
        \caption{BEKAN solution}
    \end{subfigure}
    \hfill
    \begin{subfigure}[b]{0.24\linewidth}
        \centering
        \includegraphics[width=\linewidth]{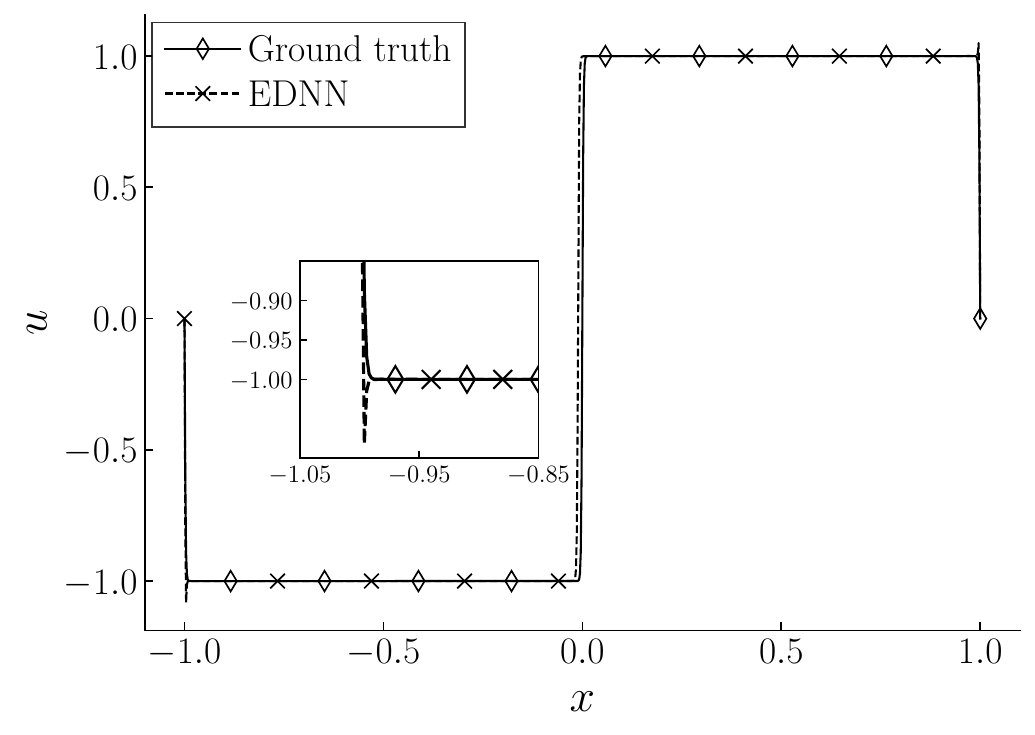}
        \caption{EDNN solution}
    \end{subfigure} 
    % \hfill
    \begin{subfigure}[b]{0.24\linewidth}
        \centering
        \includegraphics[width=\linewidth]{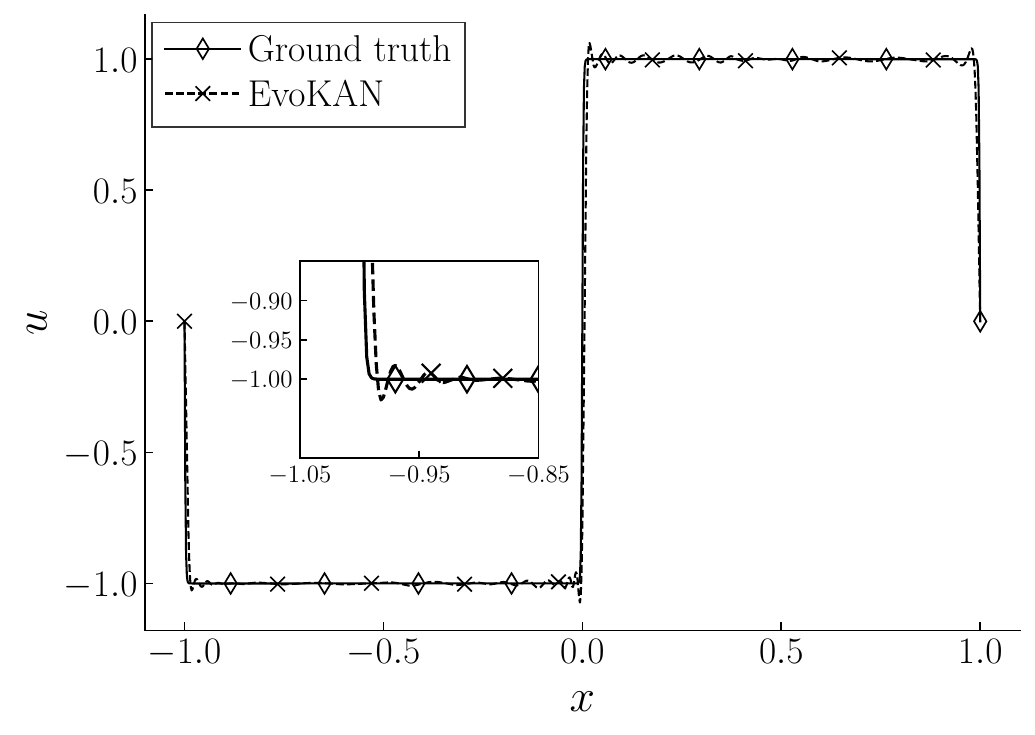}
        \caption{EvoKAN solution}
    \end{subfigure}
    \begin{subfigure}[b]{0.24\linewidth}
        \centering
        \includegraphics[width=\linewidth]{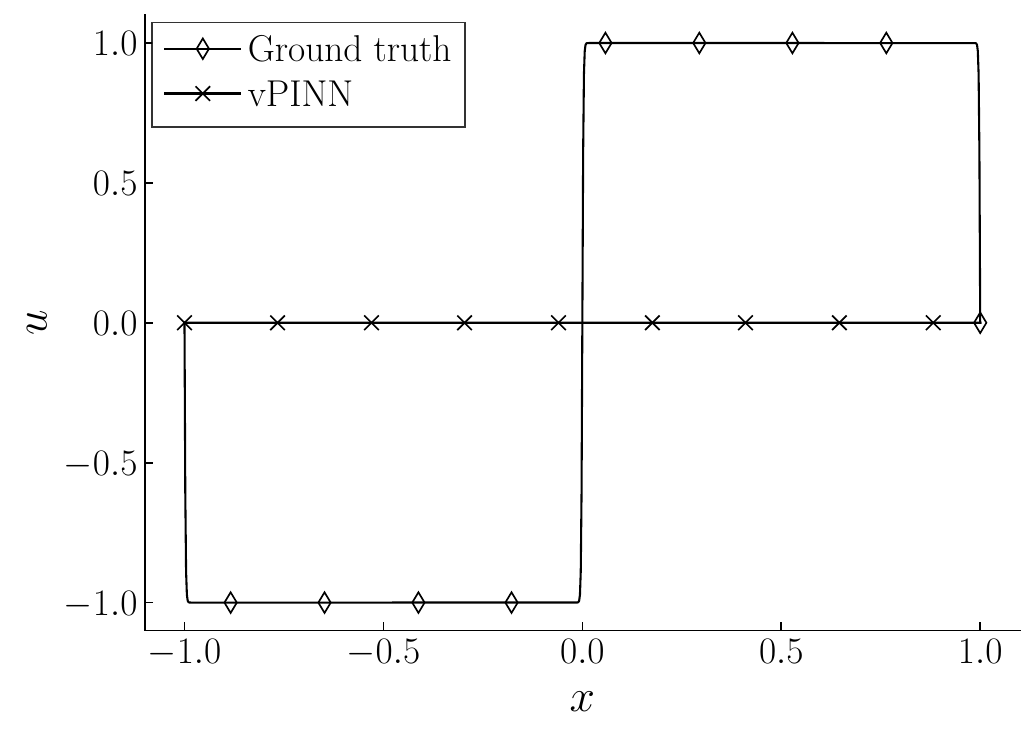}
        \caption{vPINN solution}
    \end{subfigure}
    \\
    
    \begin{subfigure}[b]{0.24\linewidth}
        \centering
        \includegraphics[width=\linewidth]{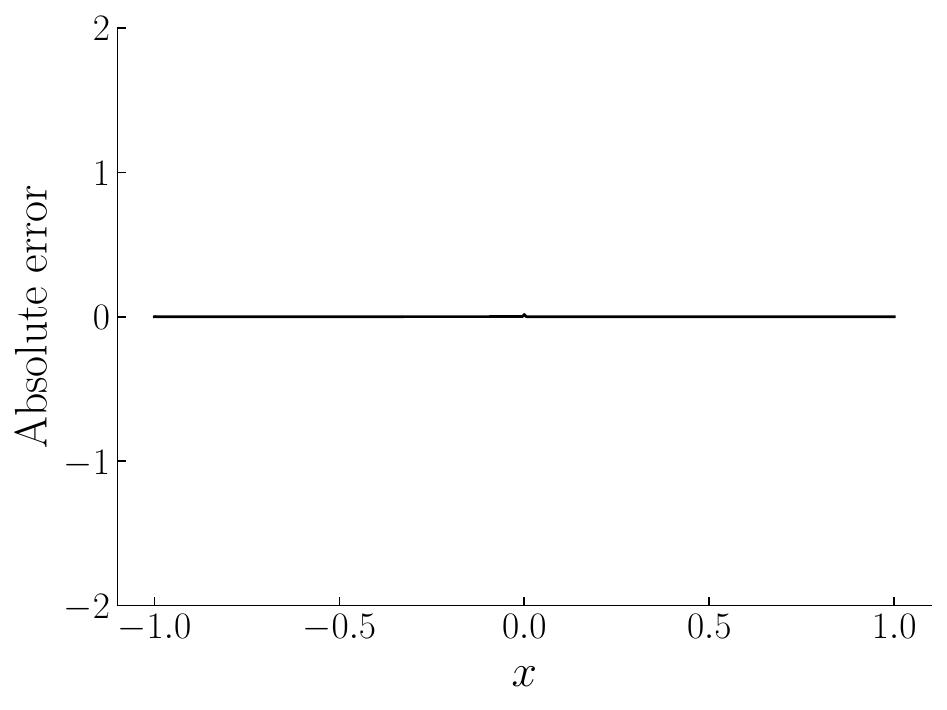}
        \caption{BEKAN error}
    \end{subfigure}
    \hfill
    \begin{subfigure}[b]{0.24\linewidth}
        \centering
        \includegraphics[width=\linewidth]{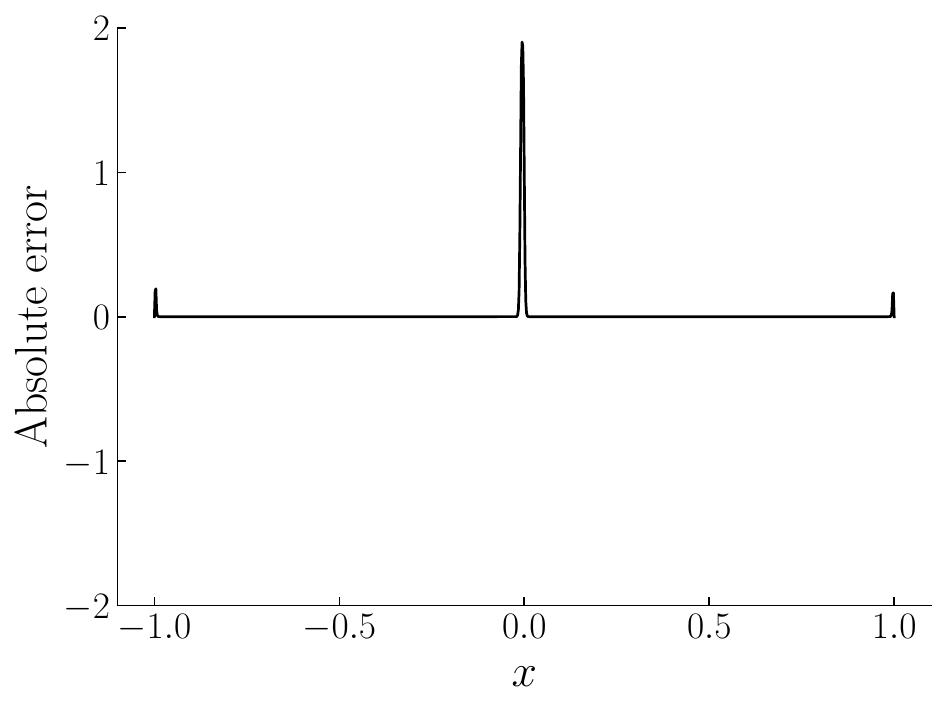}
        \caption{EDNN error}
    \end{subfigure} 
    % \hfill
    \begin{subfigure}[b]{0.24\linewidth}
        \centering
        \includegraphics[width=\linewidth]{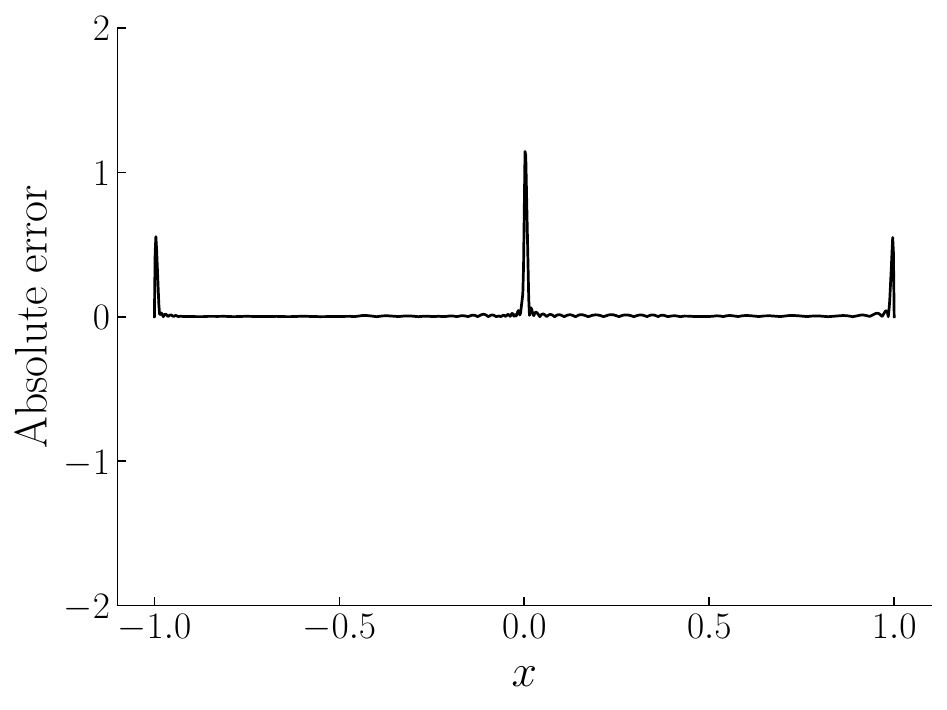}
        \caption{EvoKAN error}
    \end{subfigure}
    \begin{subfigure}[b]{0.24\linewidth}
        \centering
        \includegraphics[width=\linewidth]{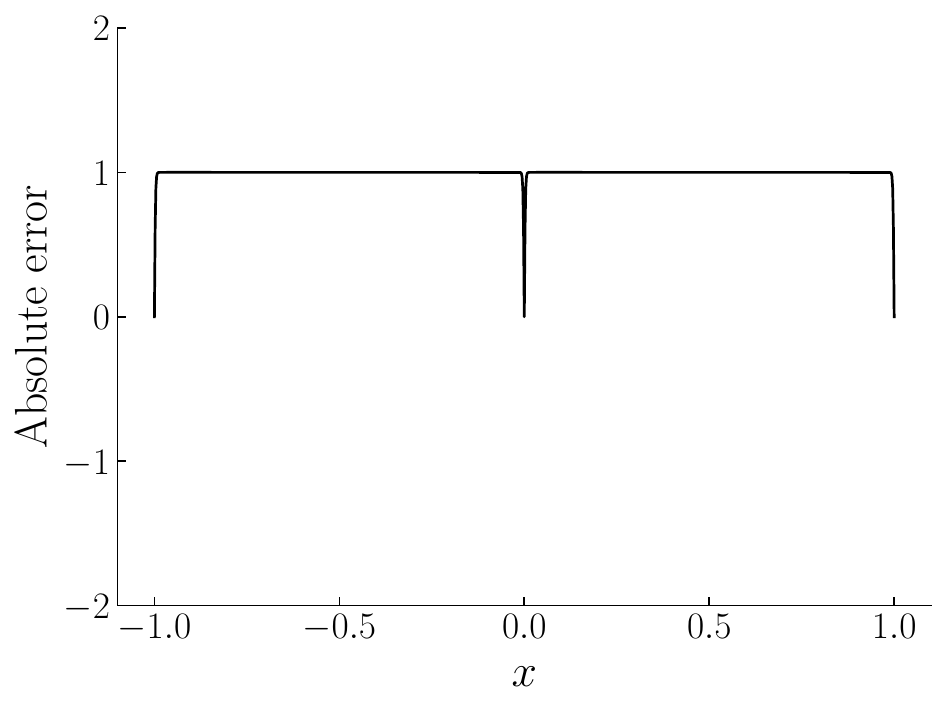}
        \caption{vPINN error}
    \end{subfigure}
    \caption{
Comparison of BEKAN, EvoKAN, EDNN, and vanilla PINN in solving the 1D Allen–Cahn equation (Eq.~\eqref{eq:Allen-Cahn}) at $t = \SI{5e-5}{}$\,s. 
Figures (a)–(d) present the predicted solutions, while (e)–(h) display the corresponding absolute errors evaluated by comparison to the reference solution.
Among the models, BEKAN achieves the highest accuracy, showing the lowest absolute error overall.
}
    \label{fig:allen_cahn_bc_t_5}
\end{figure}

To evaluate the accuracy, we plot the predicted solutions and compare them with the spectral method solution, which is taken as the reference solution.
The corresponding absolute error distributions are shown in Fig.~\ref{fig:allen_cahn_bc_t_2}. In this figure, both \ac{EDNN} and \ac{EvoKAN} show noticeable errors near the center of the domain, and the vanilla \ac{PINN} fails to capture the overall trend of the reference solution. In contrast, BEKAN exhibits close agreement with the ground truth.
As shown in Fig.~\ref{fig:allen_cahn_bc_t_5}, as the sharp interface becomes more pronounced, the errors increase for all three methods: \ac{EDNN}, \ac{EvoKAN}, and the vanilla \ac{PINN}.
In particular, \ac{EDNN} and \ac{EvoKAN} exhibit noticeable oscillations, as seen in the zoomed-in plots in Figs.~\ref{fig:allen_cahn_bc_t_5}b and c. In contrast, BEKAN accurately captures the sharp transition without such artifacts.
This experiment demonstrates that BEKAN offers improved stability and solution quality when solving nonlinear PDEs with sharp phase transitions. These results suggest that BEKAN may serve as a useful alternative to conventional neural network solvers for handling stiff and nonlinear problems.

\begin{figure}[htbp!]
    \centering
    \includegraphics[width=0.4\linewidth]{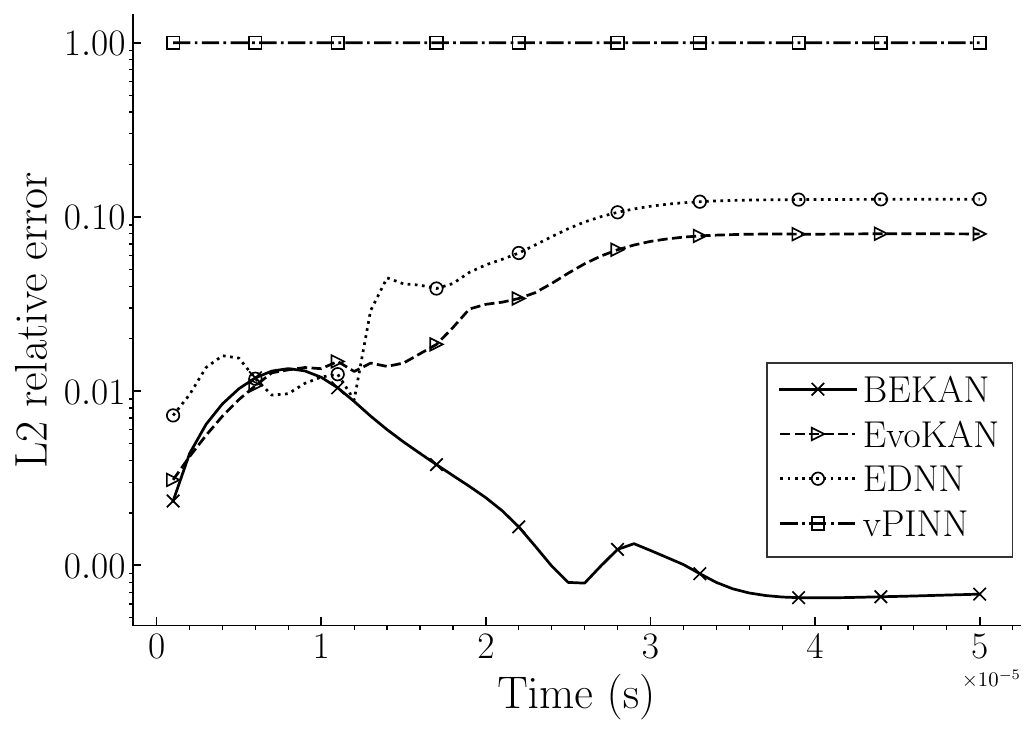}
    \caption{
Time evolution of the $L_2$ relative error for four models: BEKAN, EvoKAN, EDNN, and vanilla PINN, in solving the 1D Allen–Cahn equation (Eq.~\eqref{eq:Allen-Cahn}).  
The error is computed against a spectral solution used as the ground truth and evaluated at each time step. BEKAN outperforms EvoKAN, EDNN, and vanilla PINN in terms of $L_2$ relative error.
}
    \label{fig:AC_L2}
\end{figure}

To evaluate the error over the entire time interval, we plot the $L_2$ relative error in Fig.~\ref{fig:AC_L2}. Both \ac{EDNN} and \ac{EvoKAN} show a tendency for the error to increase over time, while BEKAN maintains a relatively low error throughout and even shows a decreasing trend as time progresses. The vanilla \ac{PINN}, on the other hand, fails to capture the solution accurately and exhibits the highest error among the methods. 

For quantitative assessment of boundary condition satisfaction, we summarizes the boundary values at time steps $t = \SI{1e-5}{}, \SI{3e-5}{},$ and $\SI{5e-5}{}$ in Table~\ref{table:AC_table}. \ac{EDNN} and \ac{EvoKAN} enforce boundary conditions through output shaping, while BEKAN imposes them by the proposed method in Sec.~\ref{subsec:Dirichlet}, resulting in exact satisfaction. In contrast, the vanilla \ac{PINN}, which employs a soft constraint approach, does not strictly satisfy the boundary conditions.

\begin{table}[hbtp!]
\footnotesize
\renewcommand{\arraystretch}{1.0}
\centering
\caption{1D Allen–Cahn equation (Eq.~\eqref{eq:Allen-Cahn}): Predicted solution values at the left and right boundaries of the domain ($x = \pm 1$) from four models (BEKAN, EvoKAN, EDNN, and vanilla PINN), evaluated at $t = \SI{1e-5}{}, \SI{3e-5}{}, \SI{5e-5}{}$. We examine their compliance with the homogeneous boundary condition in Eq.~\eqref{eq:Allen-Cahn_BC}. BEKAN employs the proposed method for enforcing the Dirichlet boundary condition, as described in Sec.~\ref{subsec:Dirichlet}, whereas EvoKAN and EDNN adopt hard constraints via output transformation~\cite{guyiqi}.}
\begin{tabular}{l c c c c c}
\hline
 & BEKAN & EvoKAN & EDNN & vanilla PINN & Exact value \\
\hline
$u(x=-1,\, t=\SI{1e-5}{} )$ & $0.00000\mathrm{e}{+00}$ & $0.00000\mathrm{e}{+00}$ & $0.00000\mathrm{e}{+00}$ & $-2.84327\mathrm{e}{-06}$ & $0.00000\mathrm{e}{+00}$ \\
$u(x=1,\, t=\SI{1e-5}{})$  & $0.00000\mathrm{e}{+00}$ & $0.00000\mathrm{e}{+00}$ & $0.00000\mathrm{e}{+00}$ & $-2.20158\mathrm{e}{-06}$ & $0.00000\mathrm{e}{+00}$\\
$u(x=-1,\, t=\SI{3e-5}{})$ & $0.00000\mathrm{e}{+00}$ & $0.00000\mathrm{e}{+00}$ & $0.00000\mathrm{e}{+00}$ & $2.86647\mathrm{e}{-06}$ & $0.00000\mathrm{e}{+00}$\\
$u(x=1,\, t= \SI{3e-5}{})$  & $0.00000\mathrm{e}{+00}$ & $0.00000\mathrm{e}{+00}$ & $0.00000\mathrm{e}{+00}$ & $-2.23379\mathrm{e}{-06}$ & $0.00000\mathrm{e}{+00}$\\
$u(x=-1,\, t=\SI{5e-5}{})$ & $0.00000\mathrm{e}{+00}$ & $0.00000\mathrm{e}{+00}$ & $0.00000\mathrm{e}{+00}$ & $-2.90209\mathrm{e}{-06}$ & $0.00000\mathrm{e}{+00}$\\
$u(x=1,\, t=\SI{5e-5}{})$  & $0.00000\mathrm{e}{+00}$ & $0.00000\mathrm{e}{+00}$ & $0.00000\mathrm{e}{+00}$ & $-2.26510\mathrm{e}{-06}$ & $0.00000\mathrm{e}{+00}$\\
\hline
\end{tabular}
\label{table:AC_table}
\end{table}

\subsection{2D Burgers' Equation with Dirichlet Boundary Condition}

The two-dimensional Burgers' equation is a canonical PDE in fluid dynamics, frequently employed to represent transport processes including compressible flow, turbulent behavior, and vehicular traffic. The equation for the velocity field $u = (u_1, u_2)$ in two spatial dimensions takes the form:
\begin{equation}
\label{eq:Burgers_equation}
\frac{\partial u}{\partial t} + u\frac{\partial u}{\partial x} + u\frac{\partial u}{\partial y} = \nu \left( \frac{\partial^2 u}{\partial x^2} + \frac{\partial^2 u}{\partial y^2} \right),
\end{equation}
where the viscosity is set to $\nu = 0.01$. The equation is supplemented by the following initial and boundary conditions:
% accompanied by the specified initial and boundary constraints:
\begin{align}
u_1(x, y, 0) &= \sin(\pi (x + 1)) \sin(\pi (y + 1)), \\
u_2(x, y, 0) &= \sin\left(\frac{1}{2}\pi (x + 1)\right) \sin\left(\frac{1}{2}\pi (y + 1)\right), \\
\label{Eq:Burgers_equation_BC}
u(-1, y, t) &= u(1, y, t) = u(x, -1, t) = u(x, 1, t) = 0.
\end{align}
Here, \( u_1(x, y, t) \) and \( u_2(x, y, t) \) indicate the flow velocities along the \( x \)- and \( y \)-axes, while \( \nu \) designates the kinematic viscosity.
To quantify the dissipation of energy due to viscosity, we define the following dissipation functional:
\begin{equation}
{E}[u] = \frac{1}{2} \int_{\Omega} \left( |\nabla u_1|^2 + |\nabla u_2|^2 \right) \, dx \, dy,
\end{equation}
where \( \nabla u_1 = (\frac{\partial u_1}{\partial x}, \frac{\partial u_1}{\partial y}) \) and \( \nabla u_2 = (\frac{\partial u_2}{\partial x}, \frac{\partial u_2}{\partial y}) \) are the derivatives of the velocity fields with respect to spatial coordinates.

The training setup is detailed in Table~\ref{table:burgers_training}.  
Both BEKAN and \ac{EvoKAN} adopt an identical hidden layer structure, with an increased number of nodes and a denser arrangement of activation grid points.  
In the case of \ac{EvoKAN}, the use of B-spline functions introduces extra spline scalers, bringing the total number of trainable parameters to 2,646, whereas BEKAN contains 2,037.  
For evolutionary training, a time increment of $t = \SI{5e-5}{}$ is applied, while the vanilla \ac{PINN} model is trained across the full time domain without iterative evolution.

% The Burgers' equation, particularly in two dimensions, is a fundamental partial differential equation from fluid dynamics, often used to model various physical phenomena such as gas dynamics and traffic flow. The 2D Burgers' equation is expressed as:

% $$ \frac{\partial u}{\partial t} + u\frac{\partial u}{\partial x} + v\frac{\partial u}{\partial y} = \nu (\frac{\partial^2 u}{\partial x^2} + \frac{\partial^2 u}{\partial y^2}) $$
% $$ u(x, y, 0) = \sin(\pi (x + 1))\sin(\pi (y + 1)) $$
% $$ v(x, y, 0) = \sin(\frac{1}{2}\pi (x + 1))\sin(\frac{1}{2}\pi(y + 1)) $$
% $$ u(-1, y, t) = u(1, y, t) = u(x, -1, t) = u(x, 1, t) = 0 $$

% The 2D setting presents a more challenging scenario due to the increased complexity of the dynamics and the higher dimensionality of the problem. The original KAN fails to provide a reasonable solution, reflecting its limitations in handling the intricate interplay of advection and diffusion processes in two dimensions. The result indicates excessive numerical instability and an inability to adhere to the Dirichlet boundary condition. In the contrast, RadialKAN demonstrates a commendable performance. It accurately estimates the solution, maintaining stability and fidelity to the physical constraints imposed by the boundary condition. The RadioKAN's advanced architecture and training regimen likely contribute to its ability to effectively manage the complex behaviors inherent in the 2D Burgers' equation, particularly under strict boundary conditions.

\sisetup{group-separator={,}, group-minimum-digits=4}
\begin{table} [hbt!]
\footnotesize
	\renewcommand{\arraystretch}{1.0}
	\begin{center} 
		\caption{Training configuration for the 2D Burgers' equation (Eq.~\eqref{eq:Burgers_equation}).}
		\begin{tabular}{l c c c c}
			\hline
			{\, \, \, } & \makecell[c]{BEKAN} & \makecell[c]{EvoKAN} & \makecell[c]{EDNN} & {Vanilla PINN} \\
			\hline
			{Hidden layers} & {[7, 7, 7]} & \makecell[c]{[7, 7, 7]} & {[35, 35, 35]} & {[35, 35, 35]}\\
            % {} & {} & {} & {} & {[90, 90, 90, 90]}\\
            % {} & {[7, 7, 7, 7]} & \makecell[c]{[7, 7, 7, 7]} & {[20, 20, 20, 20]} & {[20, 20, 20, 20]}\\
            {Activation functions} & \makecell[c]{Gaussian RBFs/SiLU} & \makecell[c]{B-splines/SiLU} & {tanh} & {tanh} \\
            % \hline
            \makecell[l]{Grid points number\\ of activation functions} & {16} & \makecell[c]{16} & {-} & {-}\\
			\makecell[l]{Number of \\ trainable parameters} & {\SI{2037}{}} & {\SI{2646}{}} & {\SI{2697}{}} & \makecell[c]{\SI{ 2697}{}}\\
            {Optimizer} & \makecell[c]{Adam} & \makecell[c]{Adam} & \makecell[c]{Adam} & \makecell[c]{Adam/L-BFGS-B}\\
            {Timestep} & \makecell[c]{5e-05} & \makecell[c]{5e-05} & \makecell[c]{5e-05} & \makecell[c]{-}\\
			\hline
		\end{tabular}
		\label{table:burgers_training}
	\end{center}
\end{table}

\begin{figure}[htbp]
    \centering
    \begin{subfigure}[b]{0.24\linewidth}
        \centering
        \includegraphics[width=\linewidth]{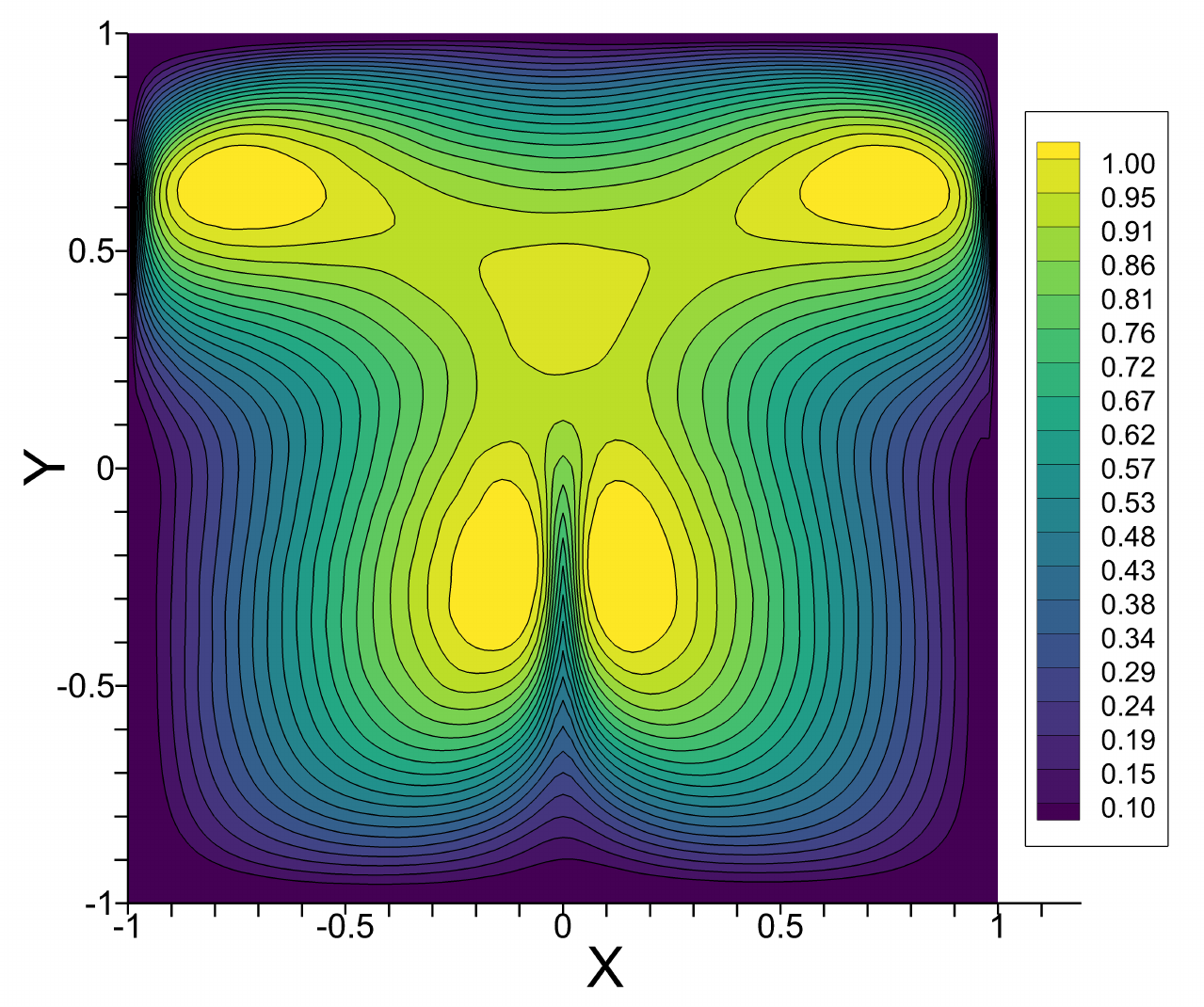}
        \caption{BEKAN solution}
    \end{subfigure}
    \hfill
    \begin{subfigure}[b]{0.24\linewidth}
        \centering
        \includegraphics[width=\linewidth]{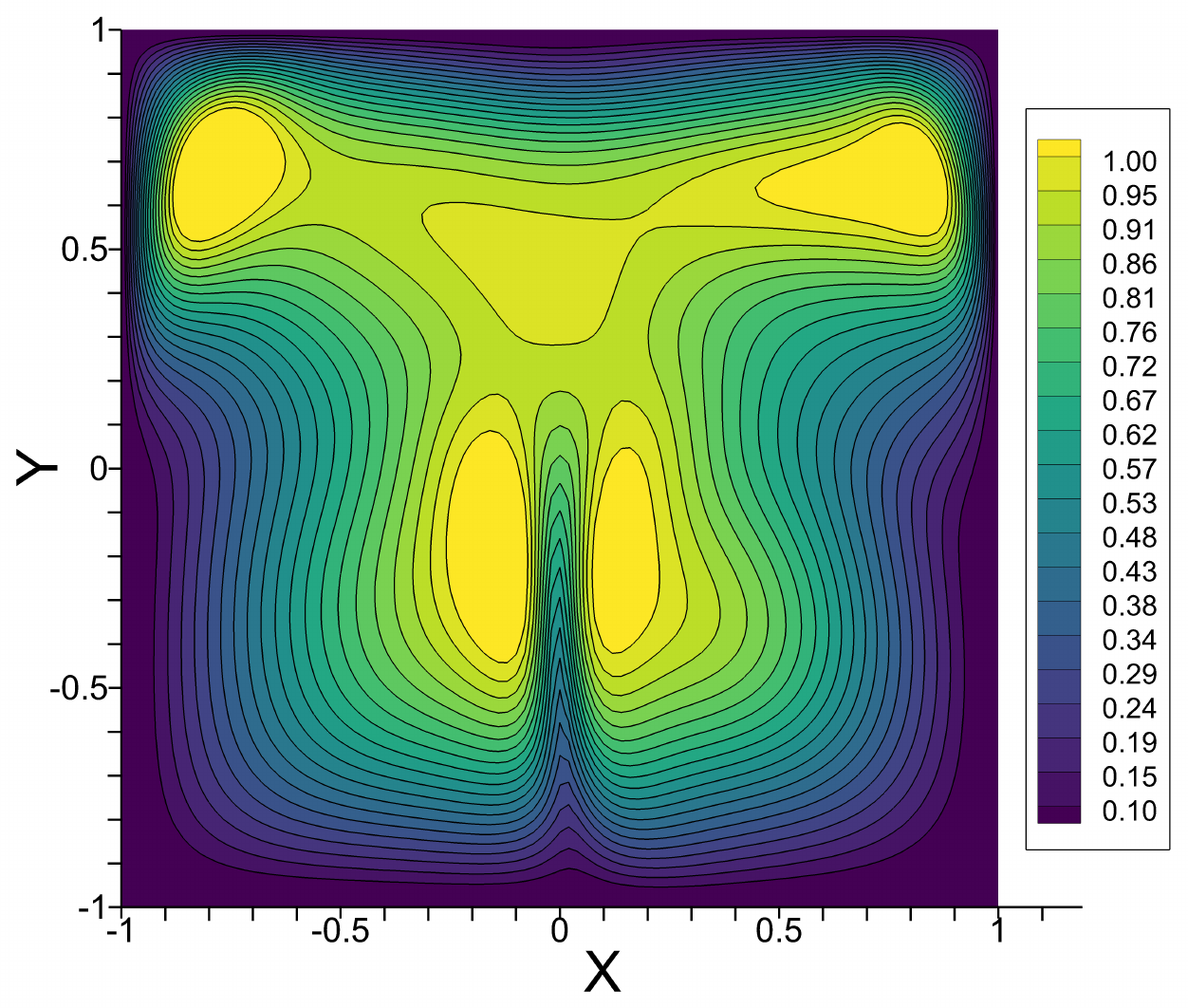}
        \caption{EDNN solution} 
    \end{subfigure} 
    \hfill
    \begin{subfigure}[b]{0.24\linewidth}
        \centering
        \includegraphics[width=\linewidth]{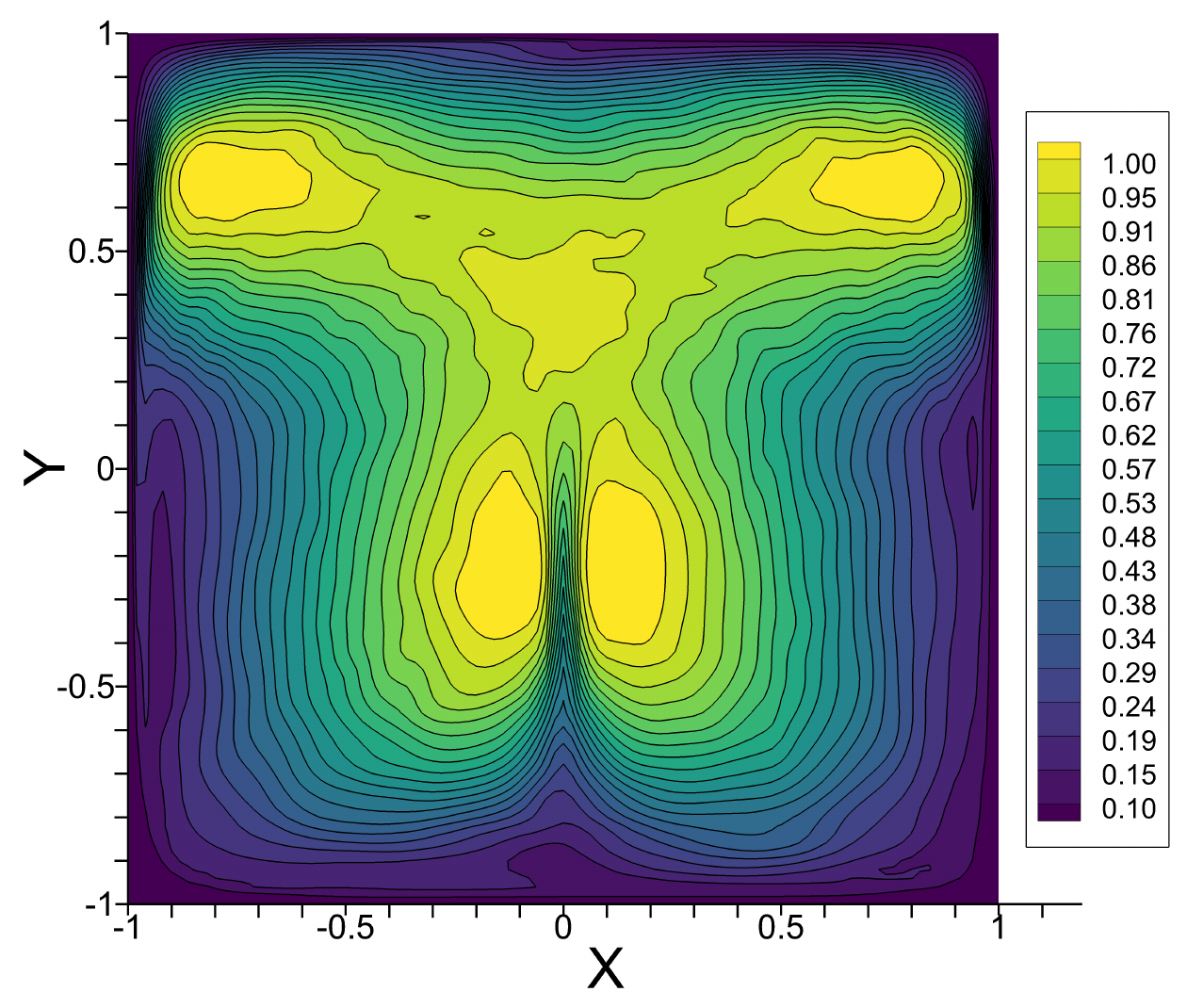}
        \caption{EvoKAN solution}
    \end{subfigure}
    \hfill
    \begin{subfigure}[b]{0.237\linewidth}
        \centering
        \includegraphics[width=\linewidth]{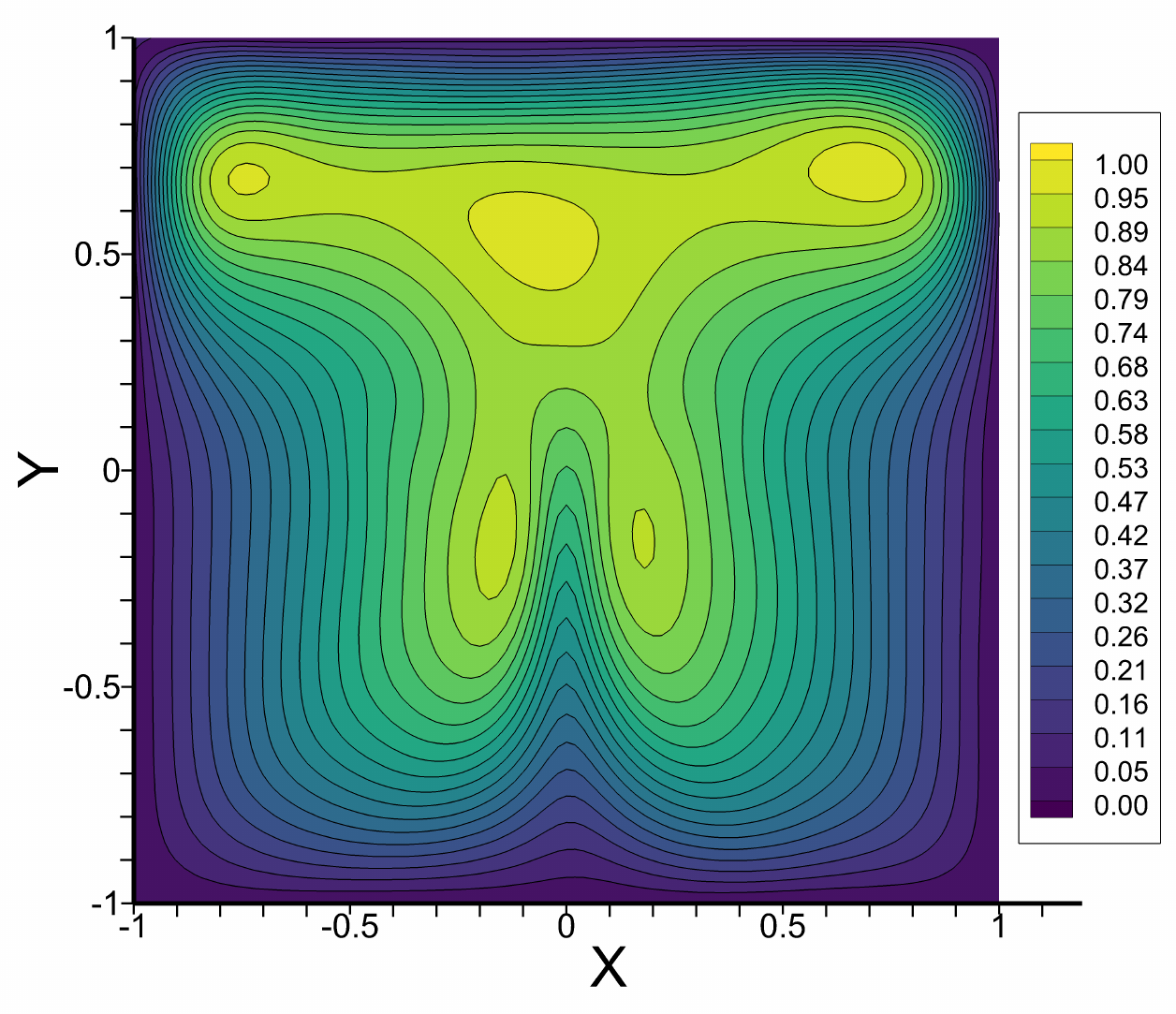}
        \caption{vPINN solution} 
    \end{subfigure} \\
    \begin{subfigure}[b]{0.24\linewidth}
        \centering
        \includegraphics[width=\linewidth]{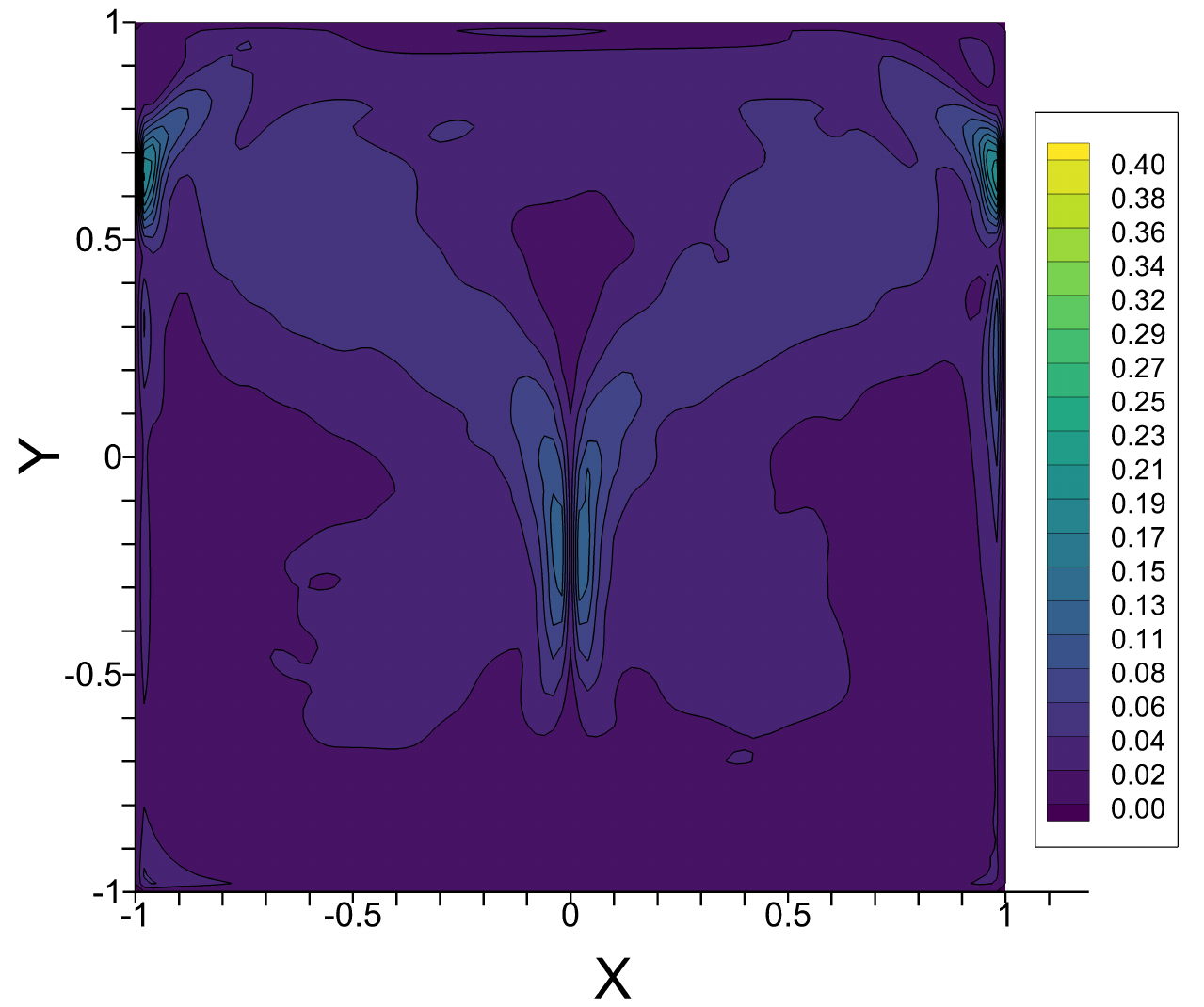}
        \caption{BEKAN absolute error}
    \end{subfigure}
    \hfill
    \begin{subfigure}[b]{0.24\linewidth}
        \centering
        \includegraphics[width=\linewidth]{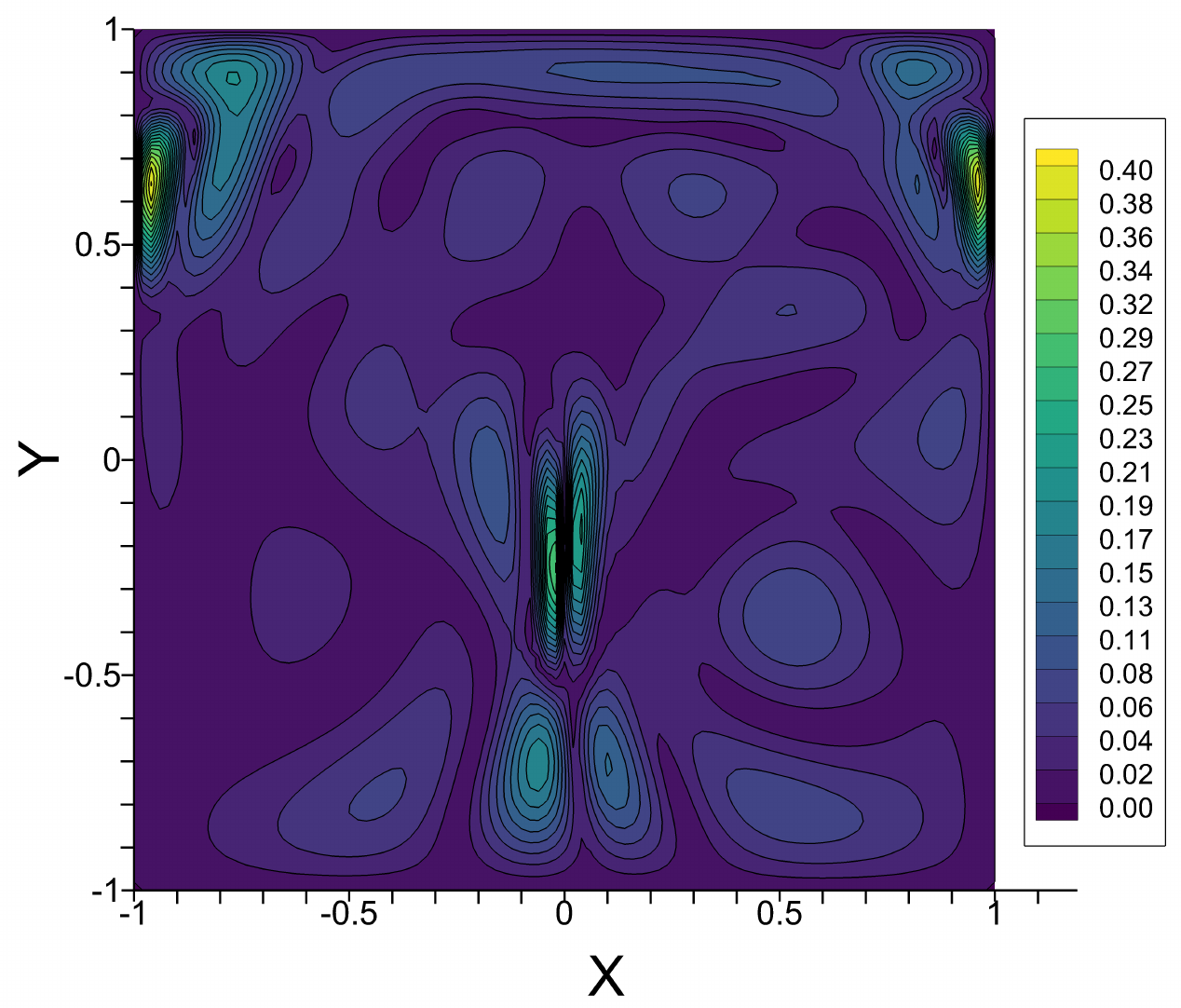}
        \caption{EDNN absolute error} 
    \end{subfigure}
    \begin{subfigure}[b]{0.24\linewidth}
        \centering
        \includegraphics[width=\linewidth]{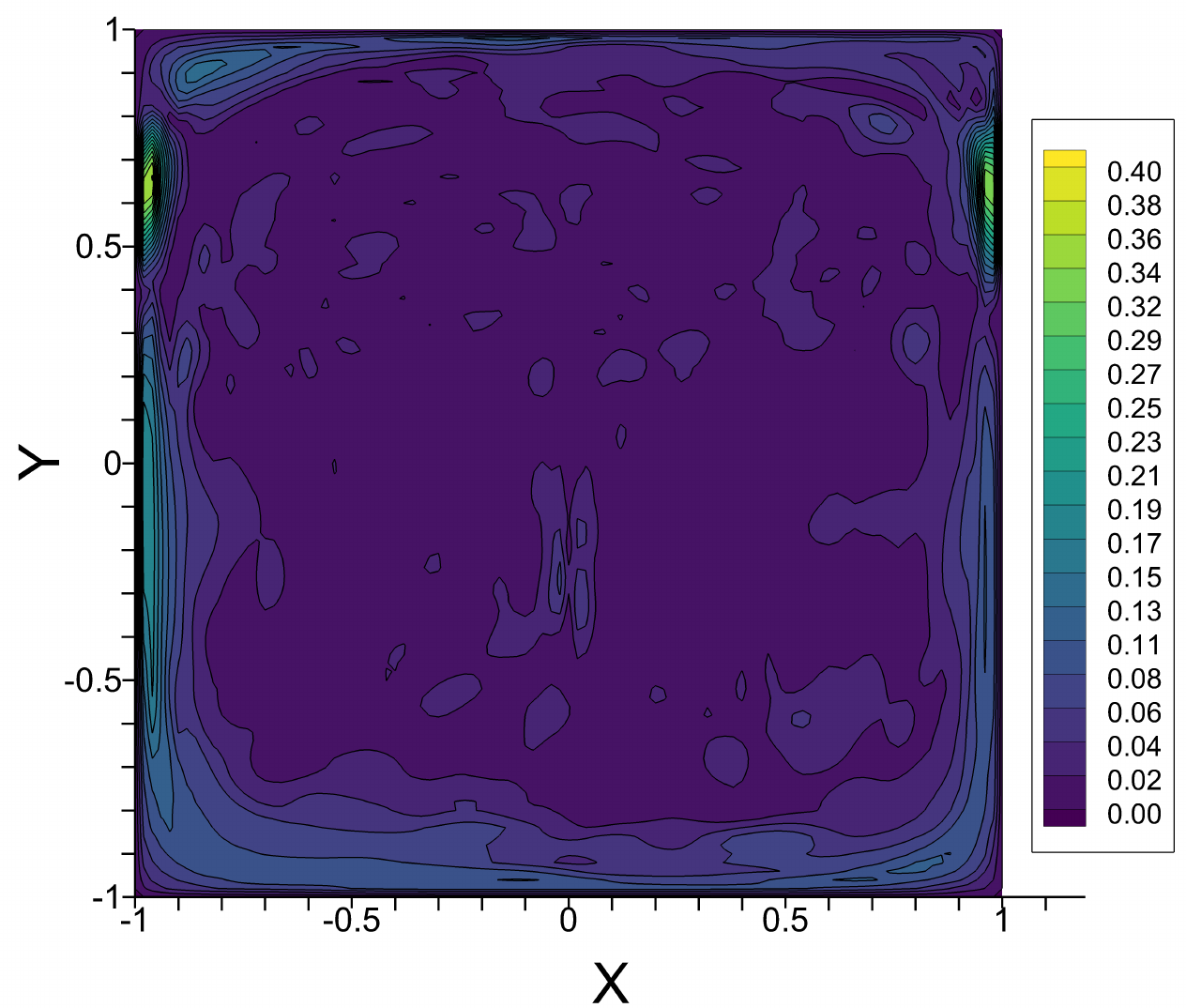}
        \caption{EvoKAN absolute error} 
    \end{subfigure}
    \begin{subfigure}[b]{0.24\linewidth}
        \centering
        \includegraphics[width=\linewidth]{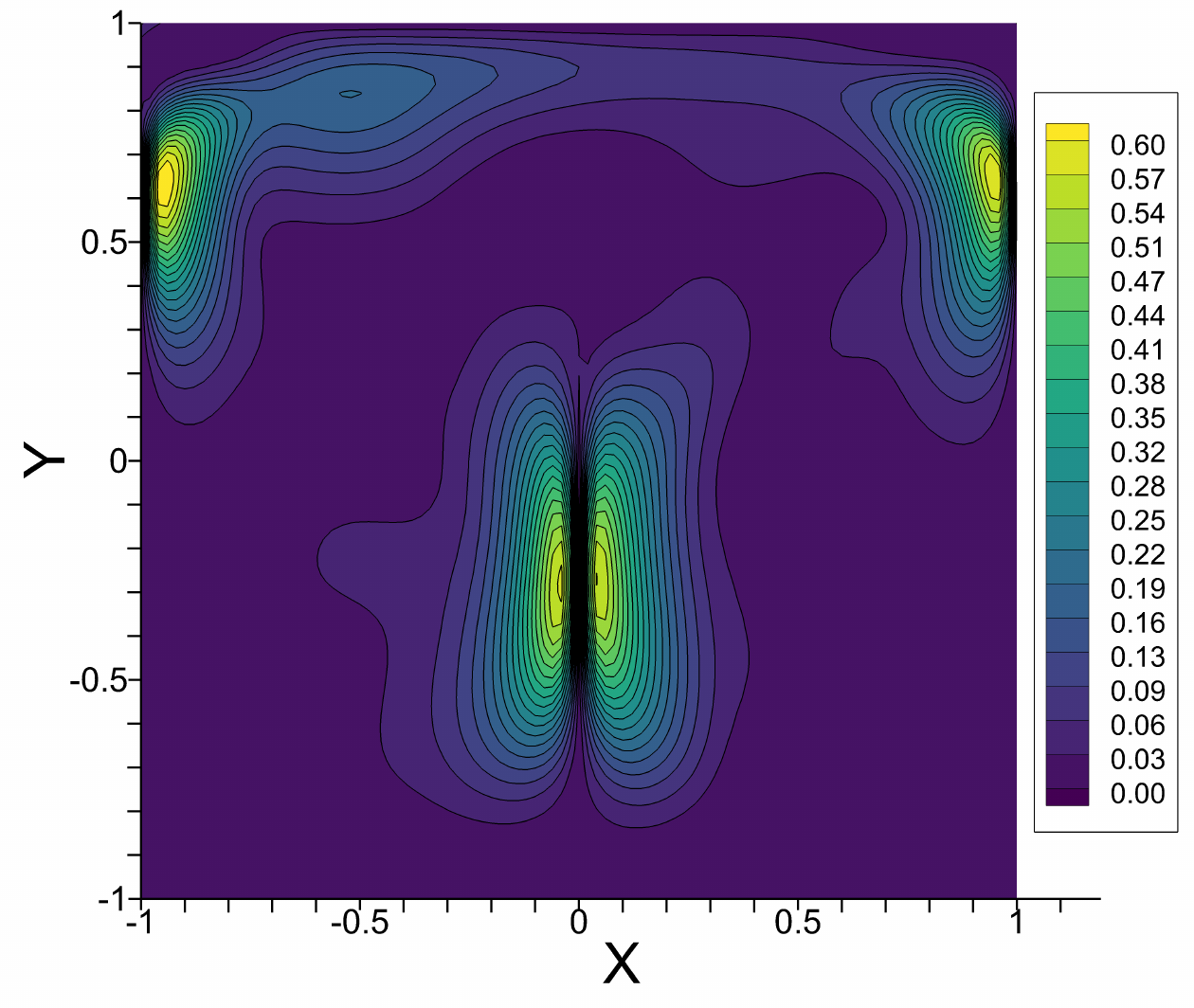}
        \caption{vPINN absolute error} 
    \end{subfigure}
    \caption{
The 2D Burgers' equation (Eq.~\eqref{eq:Burgers_equation}): Solution and absolute error distributions of BEKAN, EvoKAN, and EDNN at $t = \SI{5e-1}{}$. 
The absolute error is evaluated by comparing each prediction with the reference FDM solution. 
Among the models, BEKAN yields the most accurate and visually smooth, symmetric solution with the smallest absolute error.
}
    \label{fig:Burgers_field_distribution}
\end{figure}

For accuracy evaluation, we plot the solution of the 2D Burgers' equation at the final time step $t = \SI{5e-1}{}$. The corresponding absolute error distributions are shown in comparison with the finite difference method (FDM) solution used as the ground truth in Fig.~\ref{fig:Burgers_field_distribution}.
In Fig.~\ref{fig:Burgers_field_distribution}b and Fig.~\ref{fig:Burgers_field_distribution}c, corresponding to \ac{EDNN} and \ac{EvoKAN}, the contours are not symmetric, and oscillations appear in the \ac{EvoKAN} result. In Fig.~\ref{fig:Burgers_field_distribution}d, the vanilla \ac{PINN} does not apparently represent the steep gradient in the center of the domain, and its absolute error distribution in Fig.~\ref{fig:Burgers_field_distribution}h shows relatively large errors in the center region. In contrast, BEKAN in Fig.~\ref{fig:Burgers_field_distribution}a yields more symmetric contours, and its error distribution in Fig.~\ref{fig:Burgers_field_distribution}e exhibits smaller errors throughout the domain.

\begin{figure}[htbp]
    \centering
    \begin{subfigure}[b]{0.24\linewidth}
        \centering
        \includegraphics[width=\linewidth]{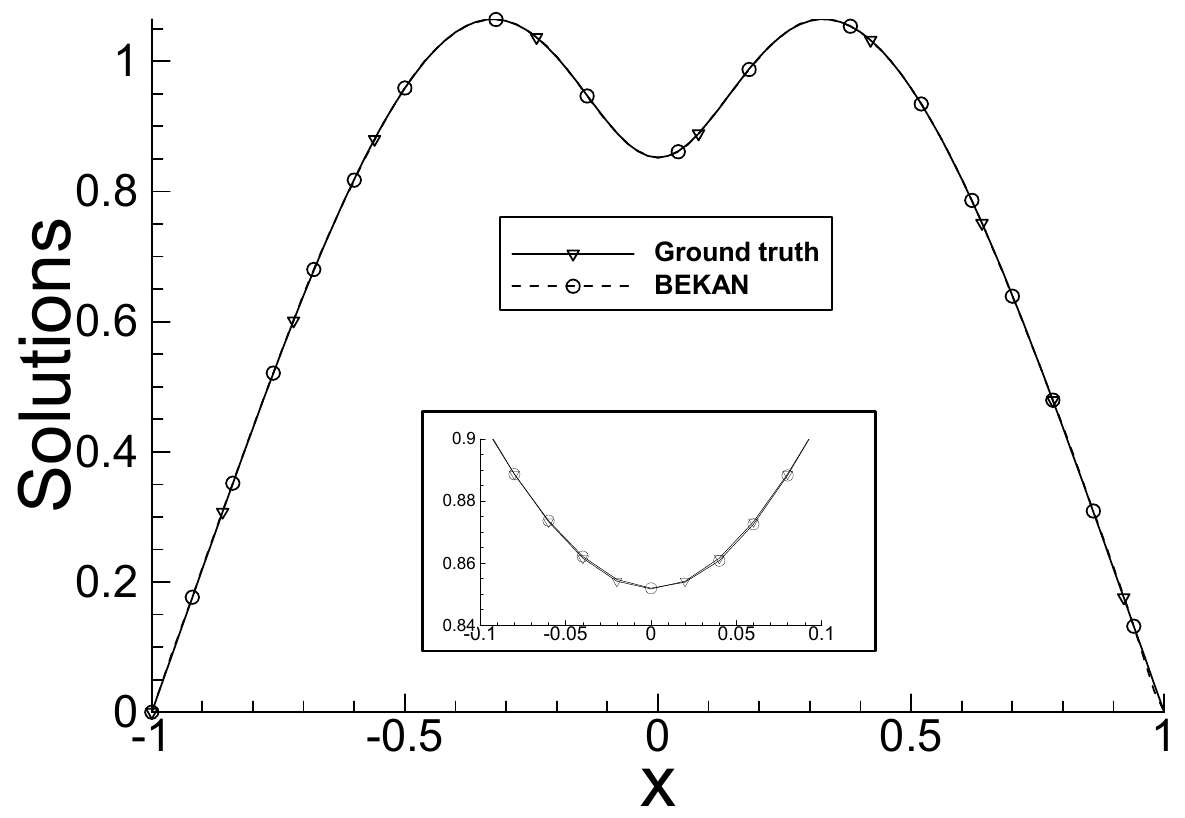}
        \caption{BEKAN solution}
    \end{subfigure}
    \hfill
    \begin{subfigure}[b]{0.24\linewidth}
        \centering
        \includegraphics[width=\linewidth]{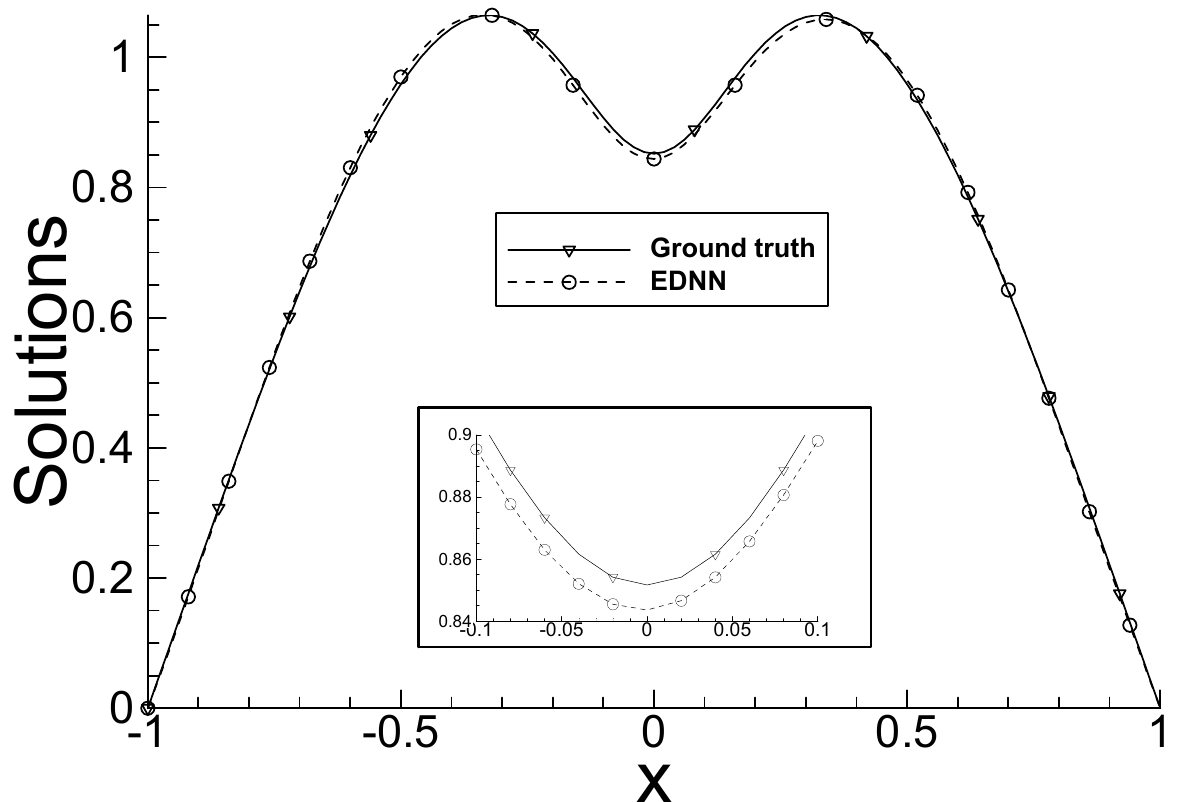}
        \caption{EDNN solution}
    \end{subfigure} 
    % \hfill
    \begin{subfigure}[b]{0.24\linewidth}
        \centering
        \includegraphics[width=\linewidth]{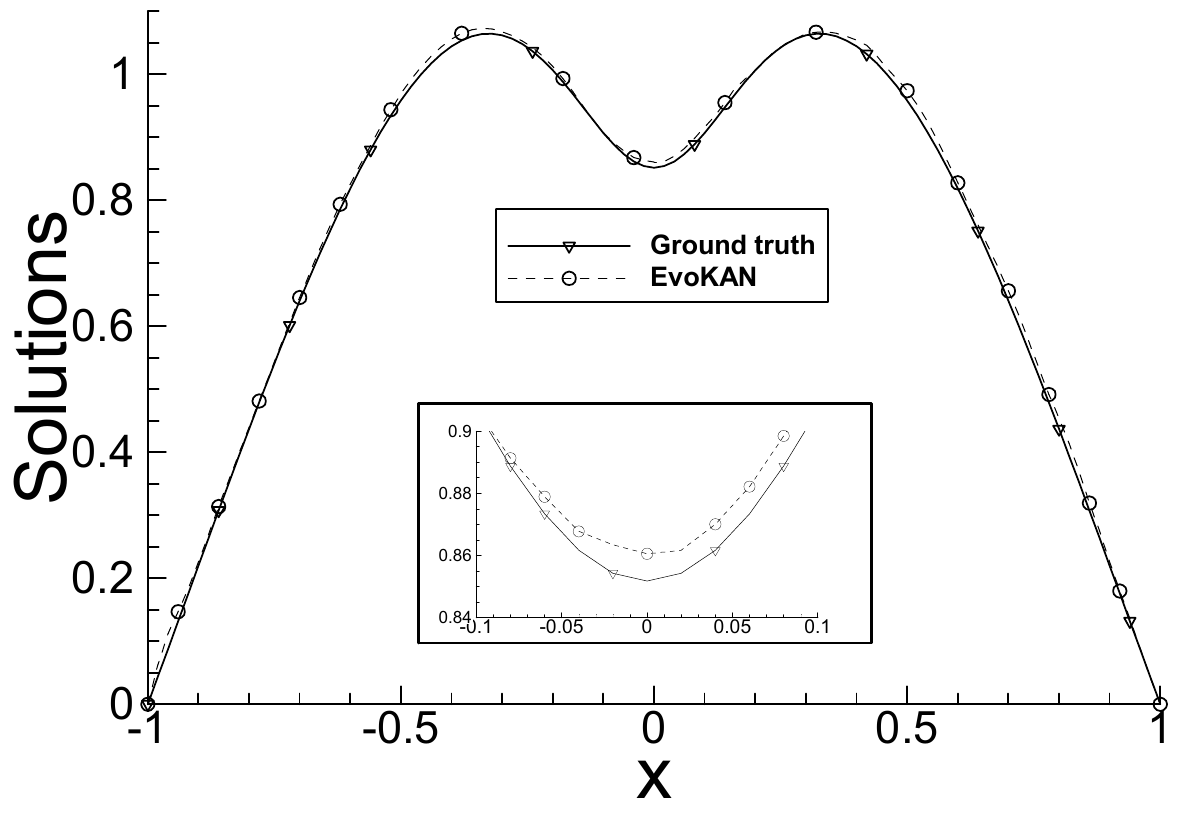}
        \caption{EvoKAN solution}
    \end{subfigure}
    \begin{subfigure}[b]{0.24\linewidth}
        \centering
        \includegraphics[width=\linewidth]{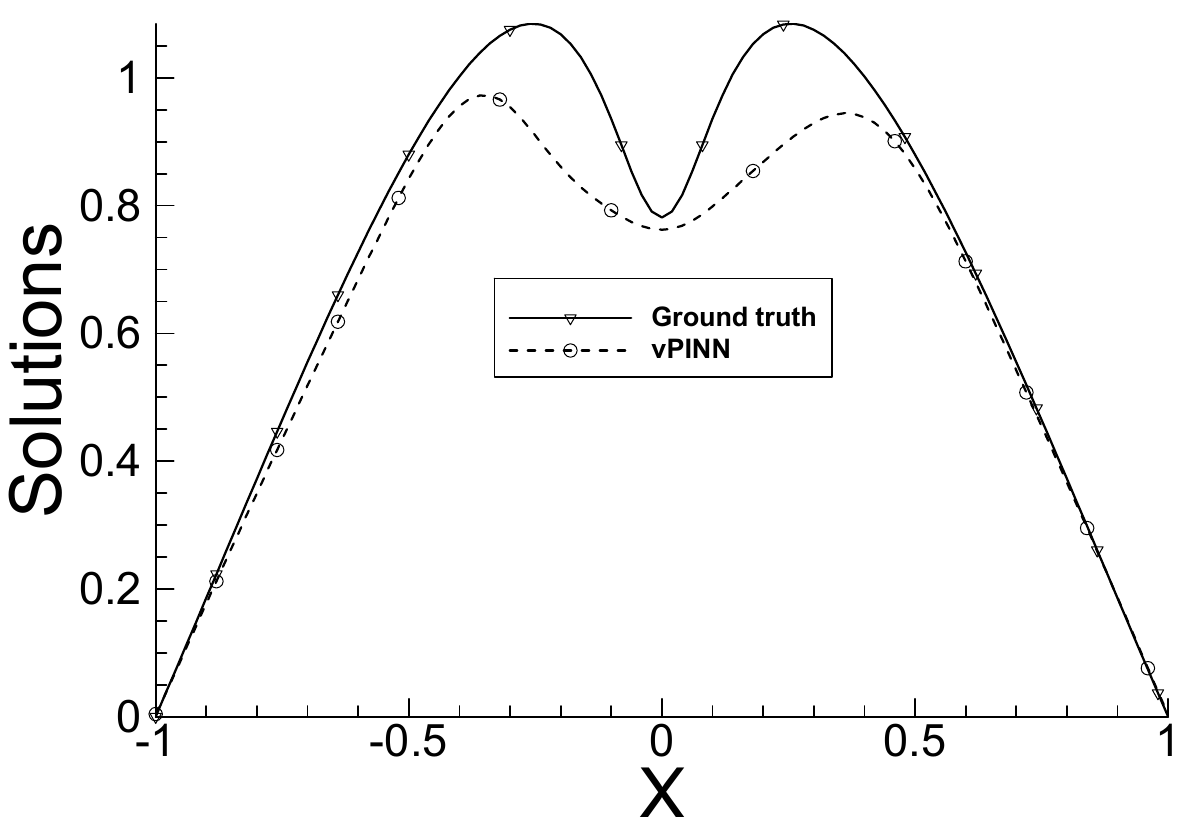}
        \caption{vPINN solution}
    \end{subfigure}
    \\
    \begin{subfigure}[b]{0.24\linewidth}
        \centering
        \includegraphics[width=\linewidth]{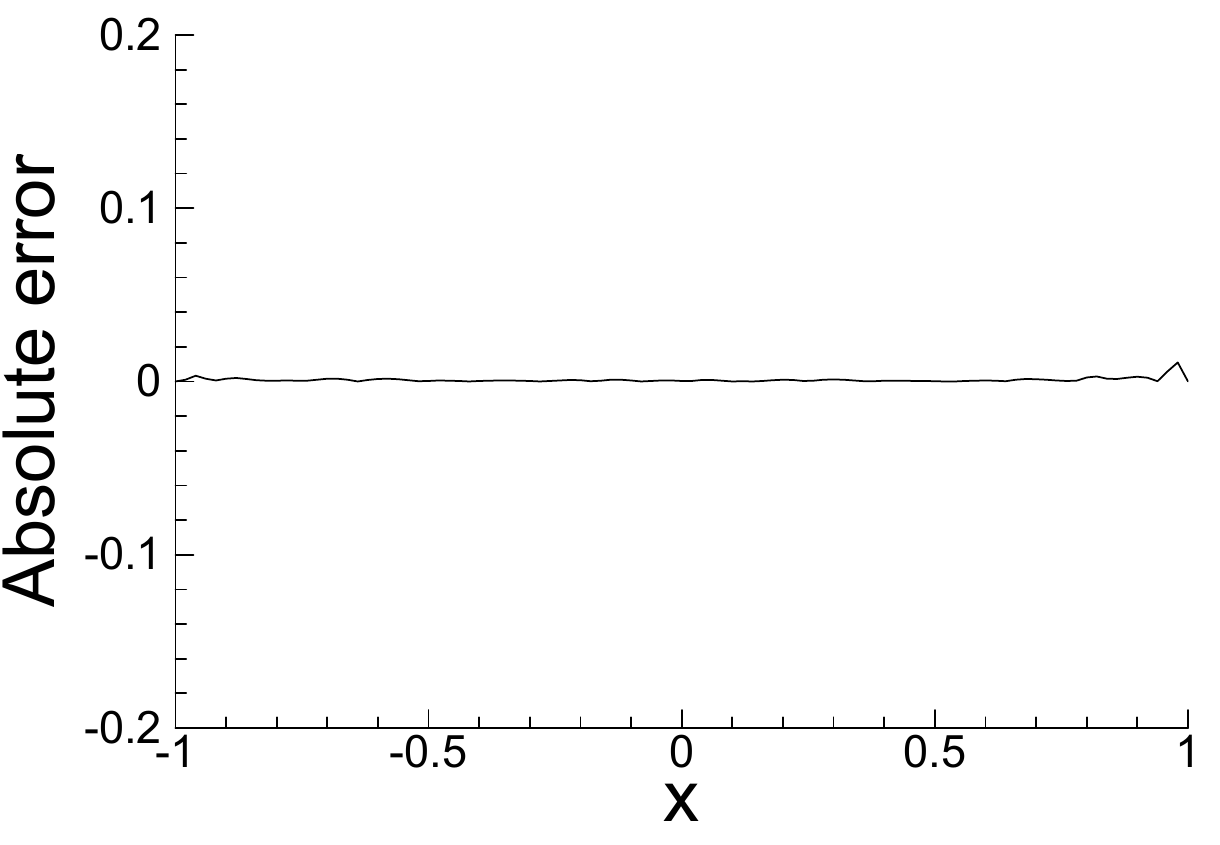}
        \caption{BEKAN absolute error}
    \end{subfigure}
    \hfill
    \begin{subfigure}[b]{0.24\linewidth}
        \centering
        \includegraphics[width=\linewidth]{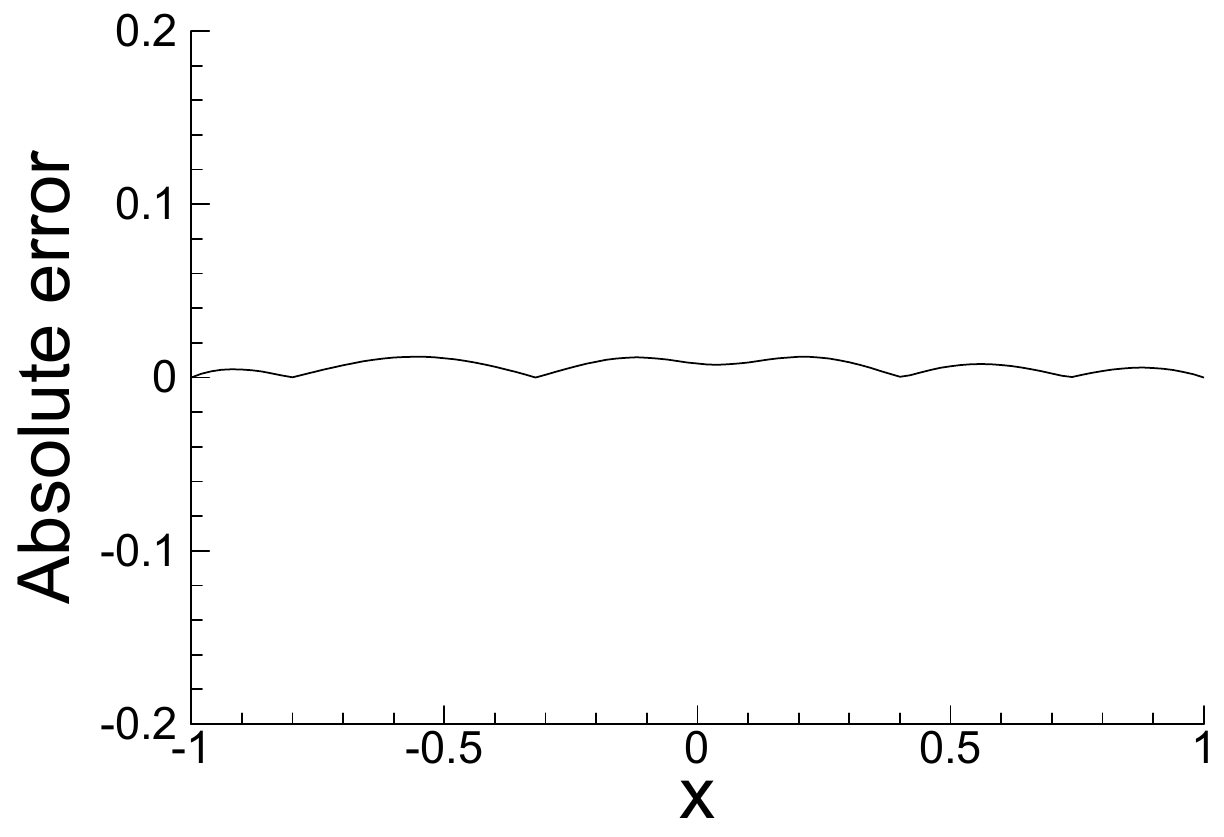}
        \caption{EDNN absolute error}
    \end{subfigure} 
    % \hfill
    \begin{subfigure}[b]{0.24\linewidth}
        \centering
        \includegraphics[width=\linewidth]{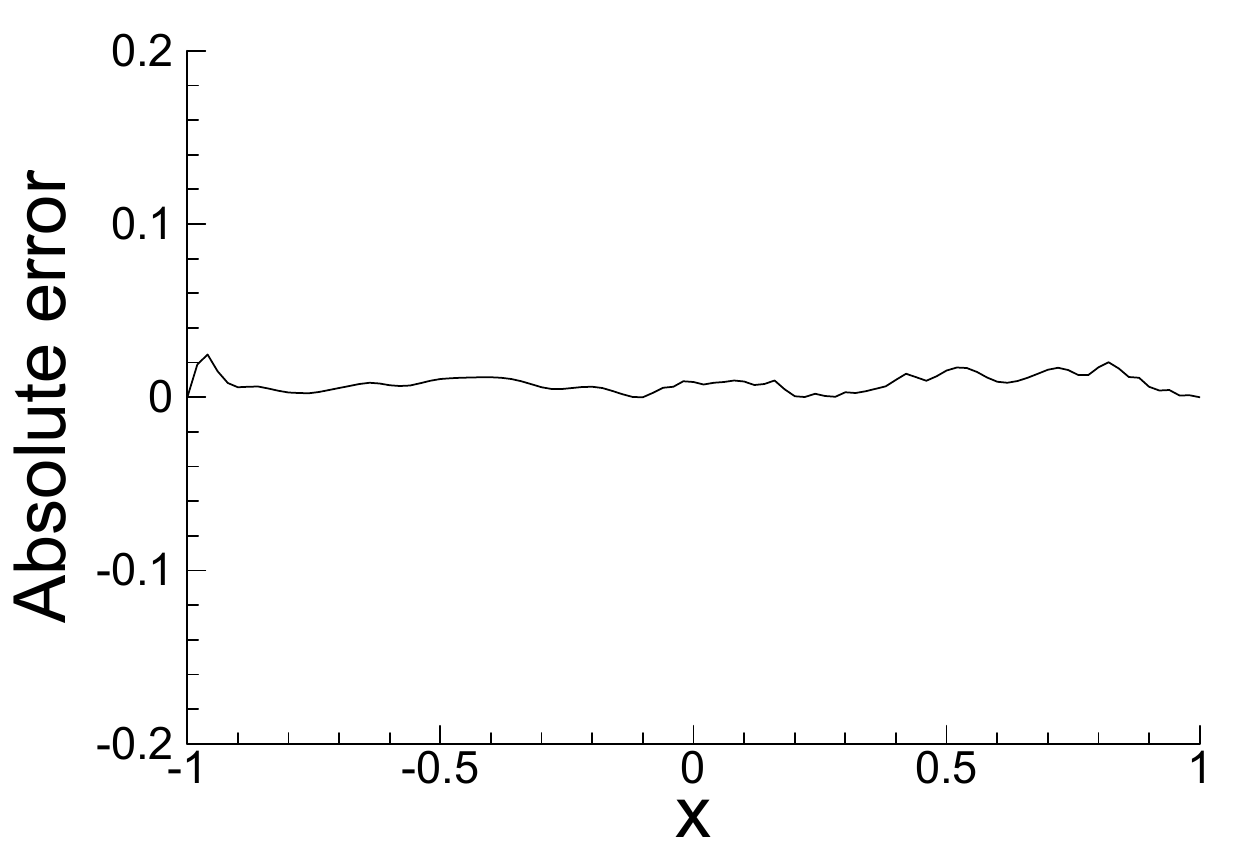}
        \caption{EvoKAN absolute error}
    \end{subfigure}
    \begin{subfigure}[b]{0.24\linewidth}
        \centering
        \includegraphics[width=\linewidth]{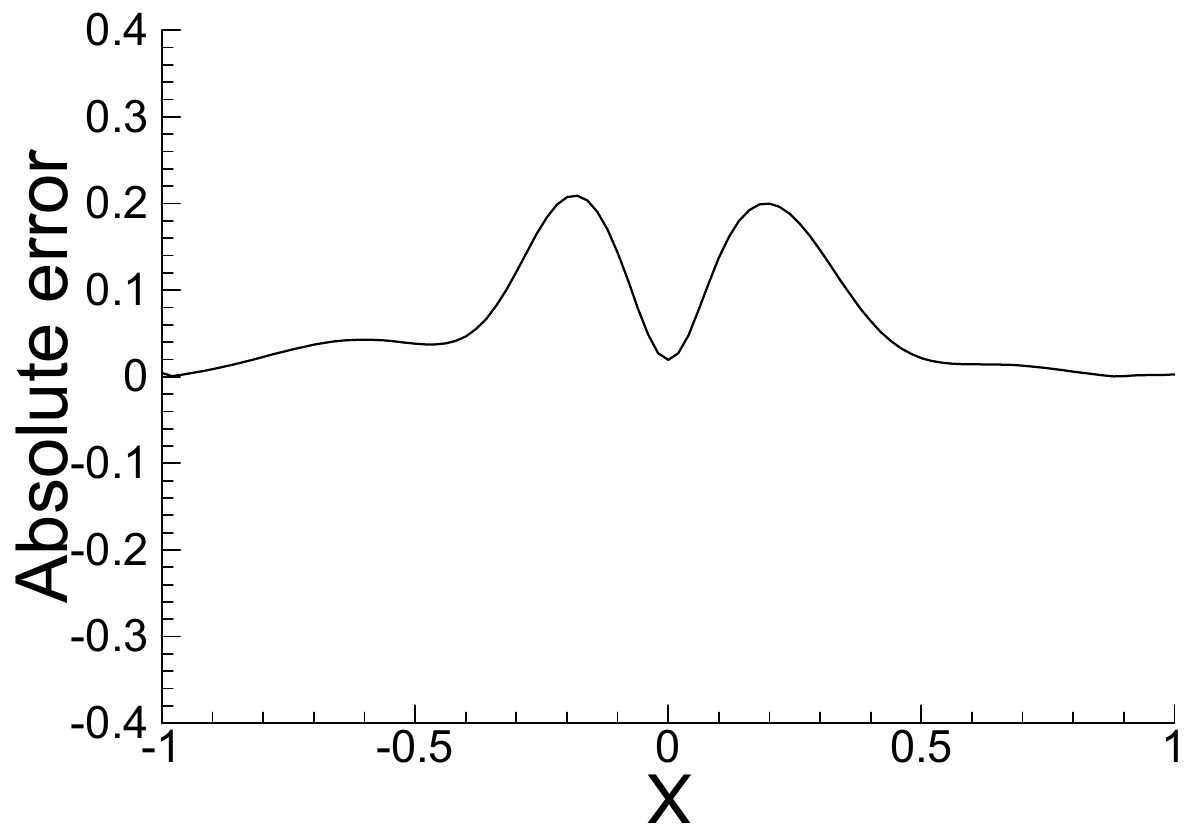}
        \caption{vPINN absolute error}
    \end{subfigure}
    \caption{
Comparison of BEKAN, EvoKAN, and EDNN in solving the 2D Burgers' equation (Eq.~\eqref{eq:Burgers_equation}) with boundary conditions $u(\pm1, y; t) = u(x, \pm1; t) = 0$ at $t = \SI{2e-1}{}$. 
Figures~(a)–(d) display the predicted solutions, while Figs.~(e)–(h) show the corresponding absolute errors. 
At this early stage, before the formation of sharp gradients, BEKAN, EDNN, and EvoKAN yield reasonable predictions. 
Notably, BEKAN solution most closely matches the FDM reference in the zoomed-in views, indicating the highest level of accuracy.
}
    \label{fig:Burgers_slice_0.2}
\end{figure}

\begin{figure}[htbp]
    \centering
    \begin{subfigure}[b]{0.24\linewidth}
        \centering
        \includegraphics[width=\linewidth]{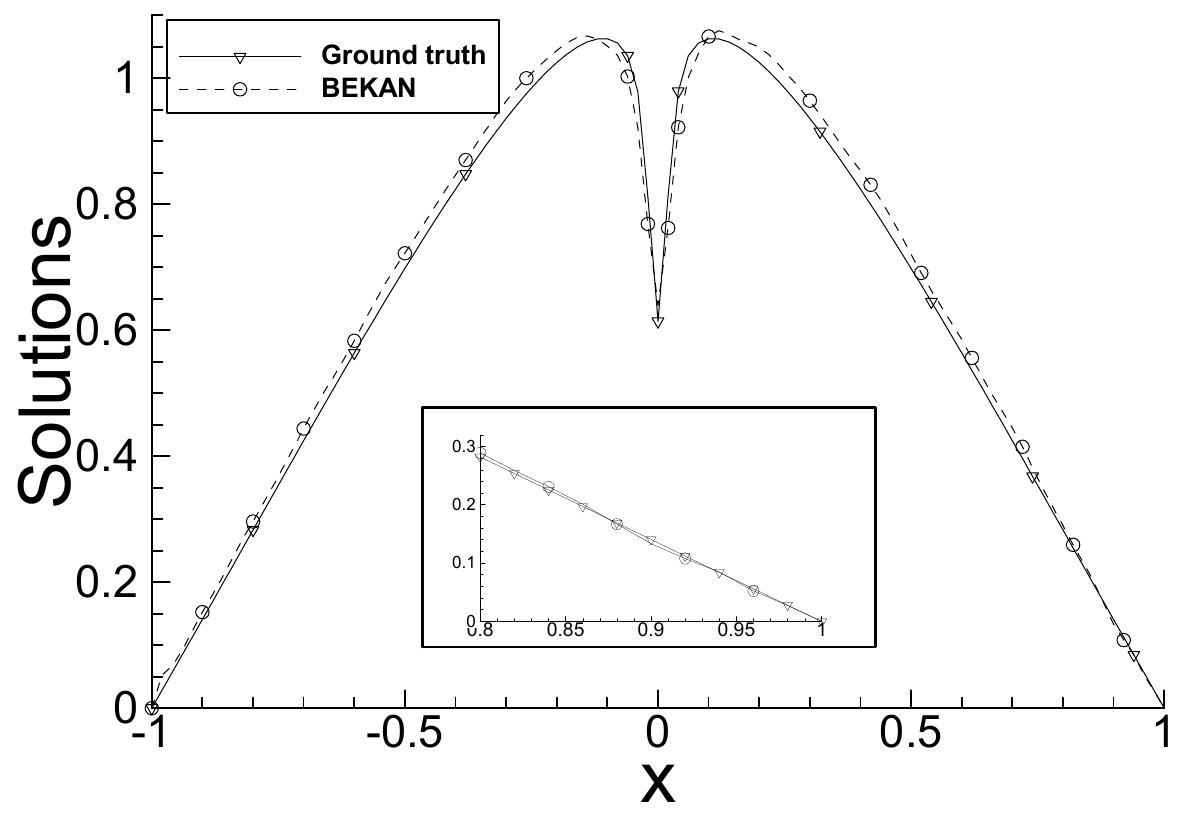}
        \caption{BEKAN solution}
    \end{subfigure}
    \hfill
    \begin{subfigure}[b]{0.24\linewidth}
        \centering
        \includegraphics[width=\linewidth]{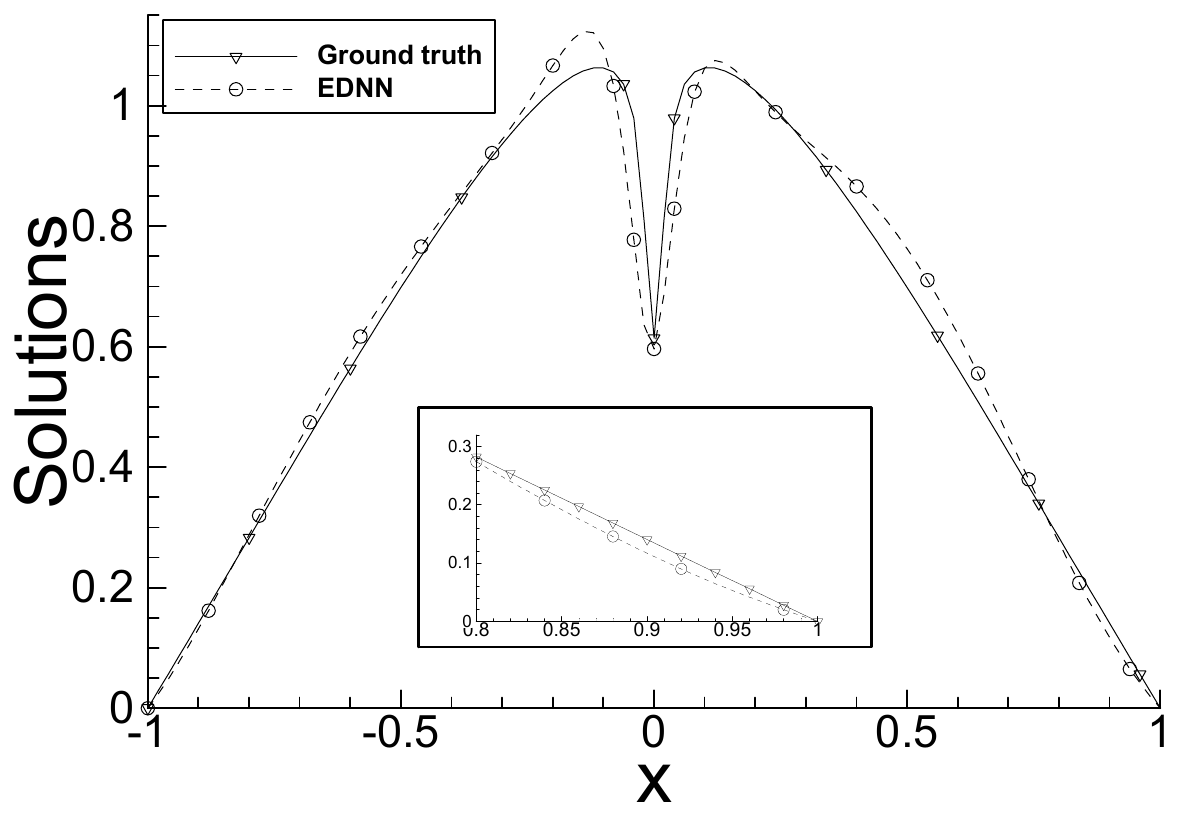}
        \caption{EDNN solution}
    \end{subfigure} 
    % \hfill
    \begin{subfigure}[b]{0.24\linewidth}
        \centering
        \includegraphics[width=\linewidth]{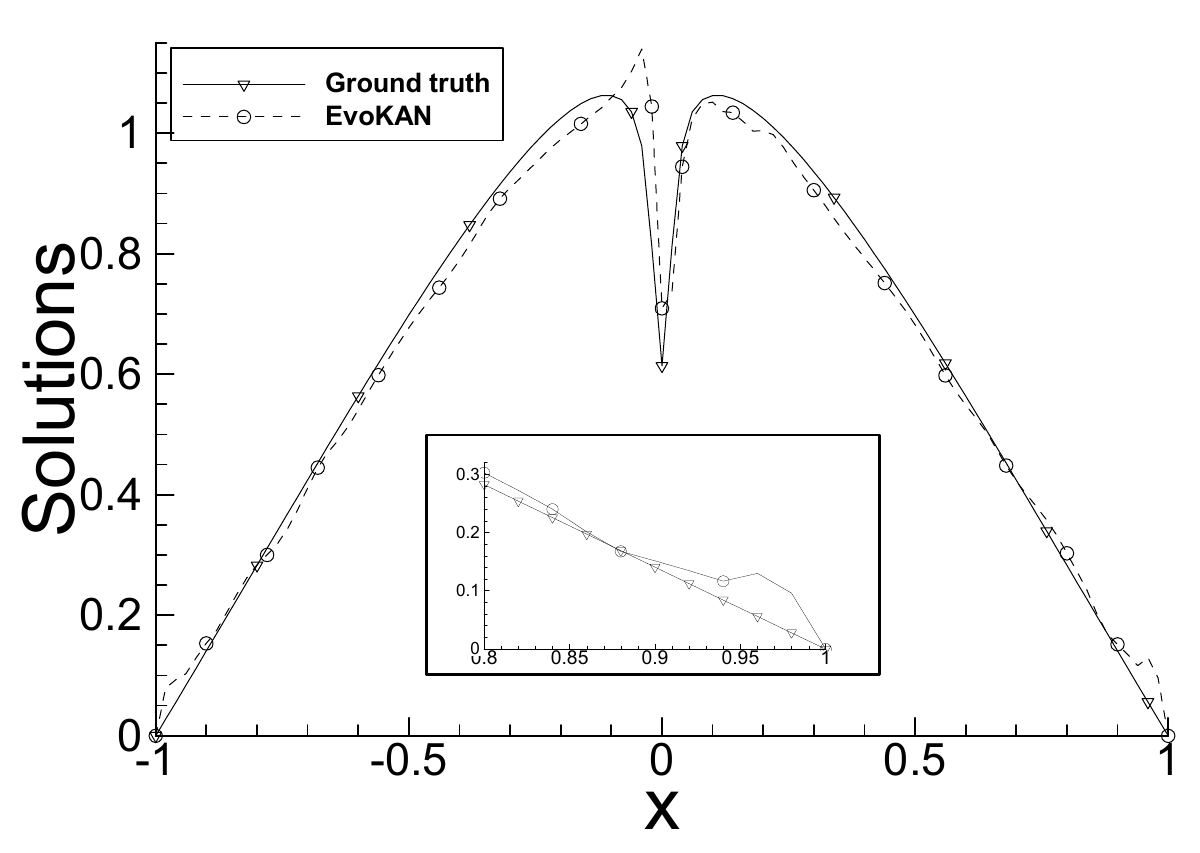}
        \caption{EvoKAN solution}
    \end{subfigure}
    \begin{subfigure}[b]{0.24\linewidth}
        \centering
        \includegraphics[width=\linewidth]{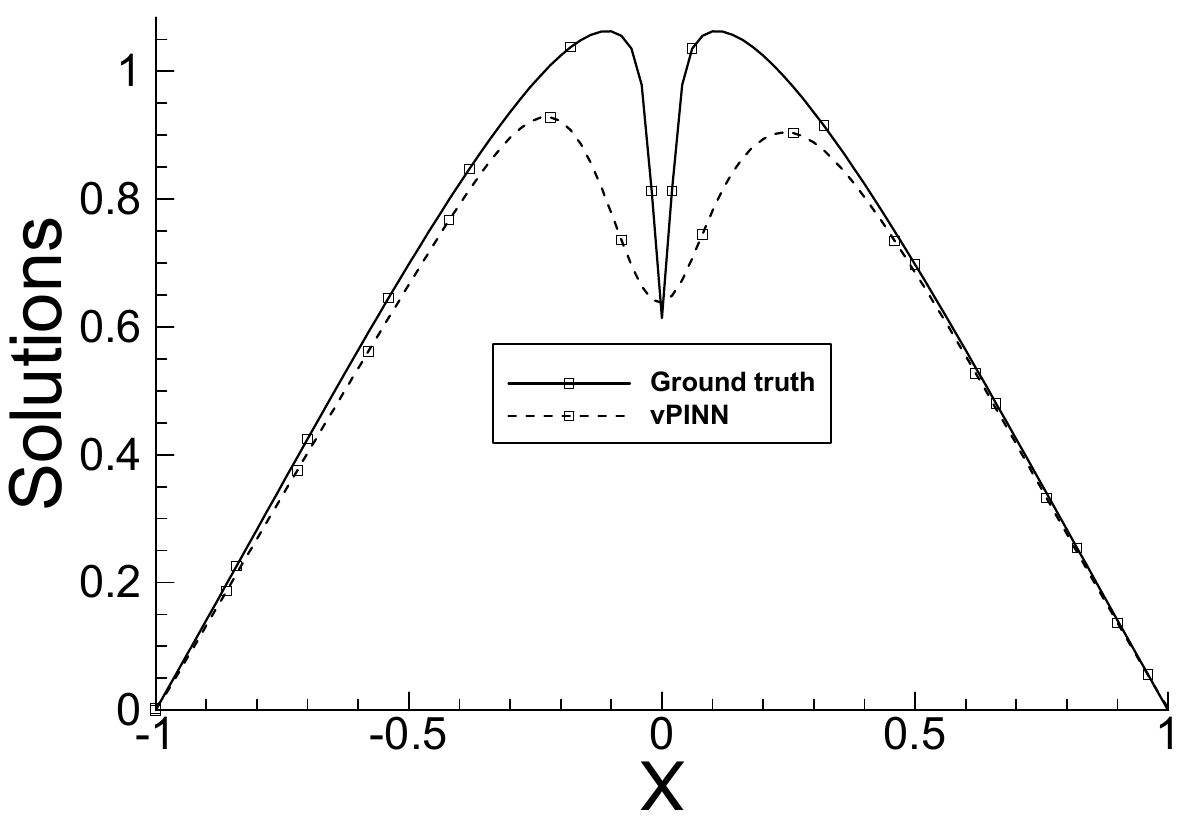}
        \caption{vPINN solution}
    \end{subfigure}
    \\
    \begin{subfigure}[b]{0.24\linewidth}
        \centering
        \includegraphics[width=\linewidth]{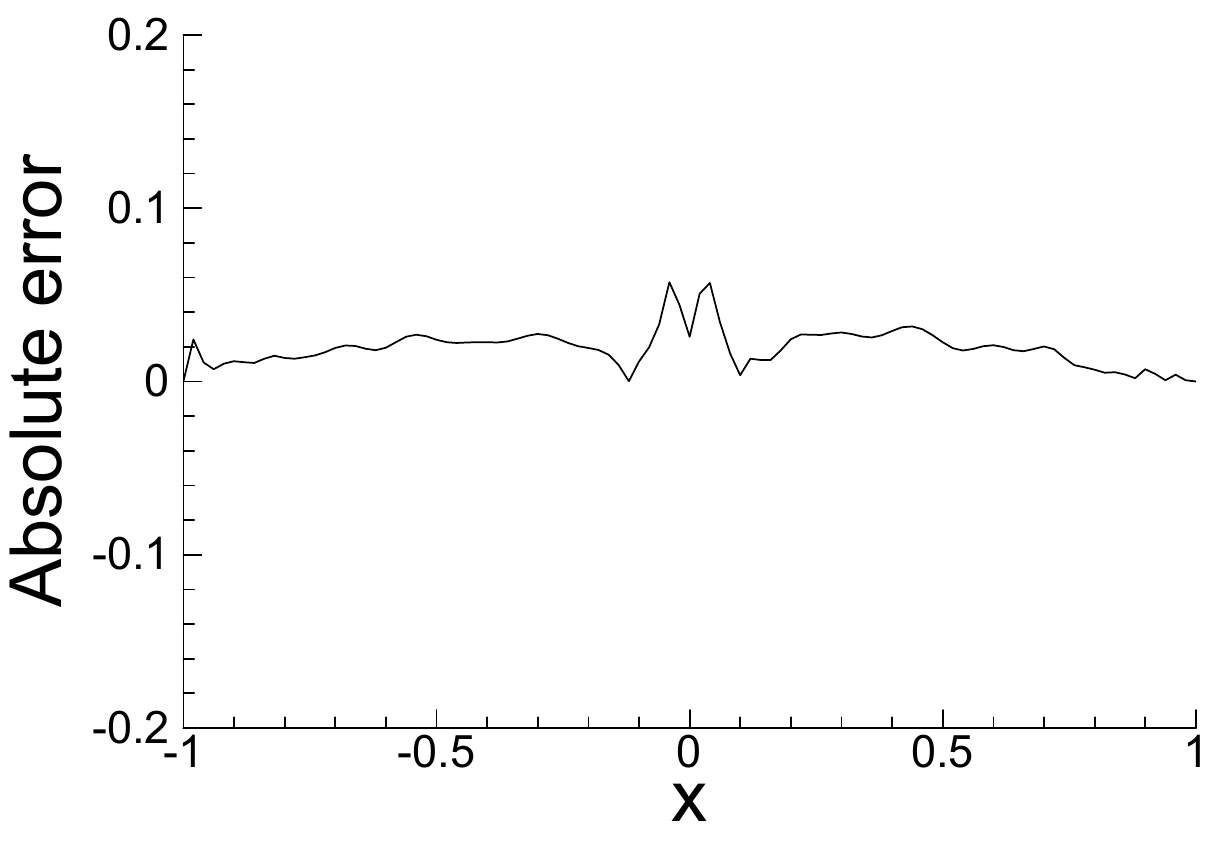}
        \caption{BEKAN absolute error}
    \end{subfigure}
    \hfill
    \begin{subfigure}[b]{0.24\linewidth}
        \centering
        \includegraphics[width=\linewidth]{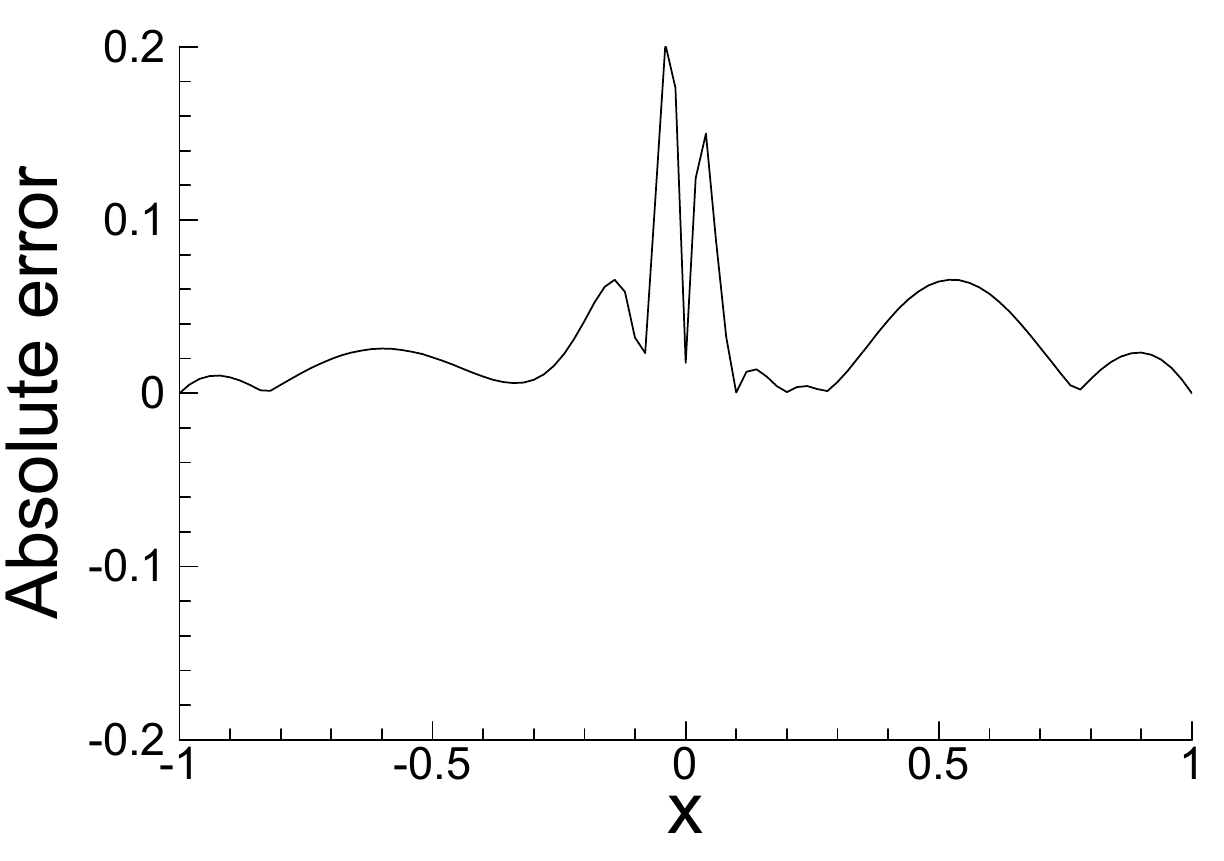}
        \caption{EDNN absolute error}
    \end{subfigure} 
    % \hfill
    \begin{subfigure}[b]{0.24\linewidth}
        \centering
        \includegraphics[width=\linewidth]{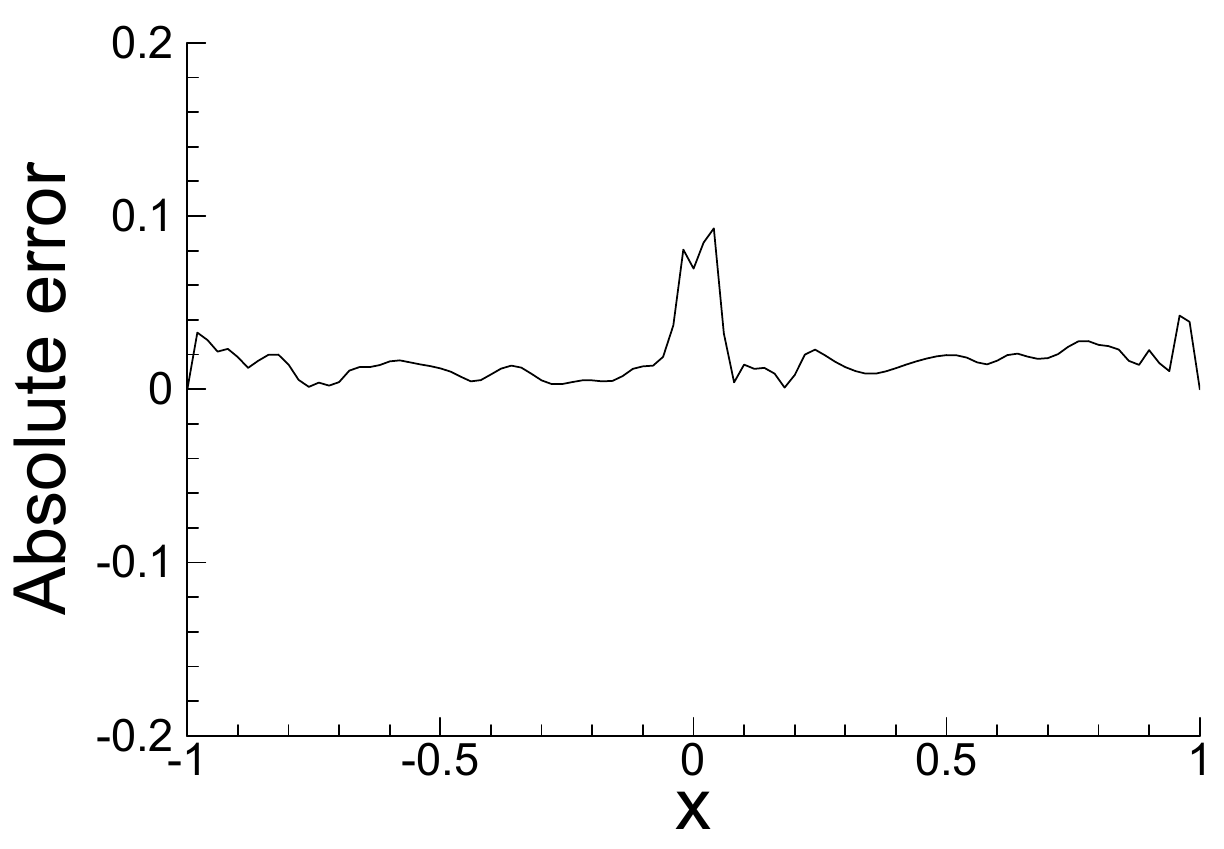}
        \caption{EvoKAN absolute error}
    \end{subfigure}
    \begin{subfigure}[b]{0.24\linewidth}
        \centering
        \includegraphics[width=\linewidth]{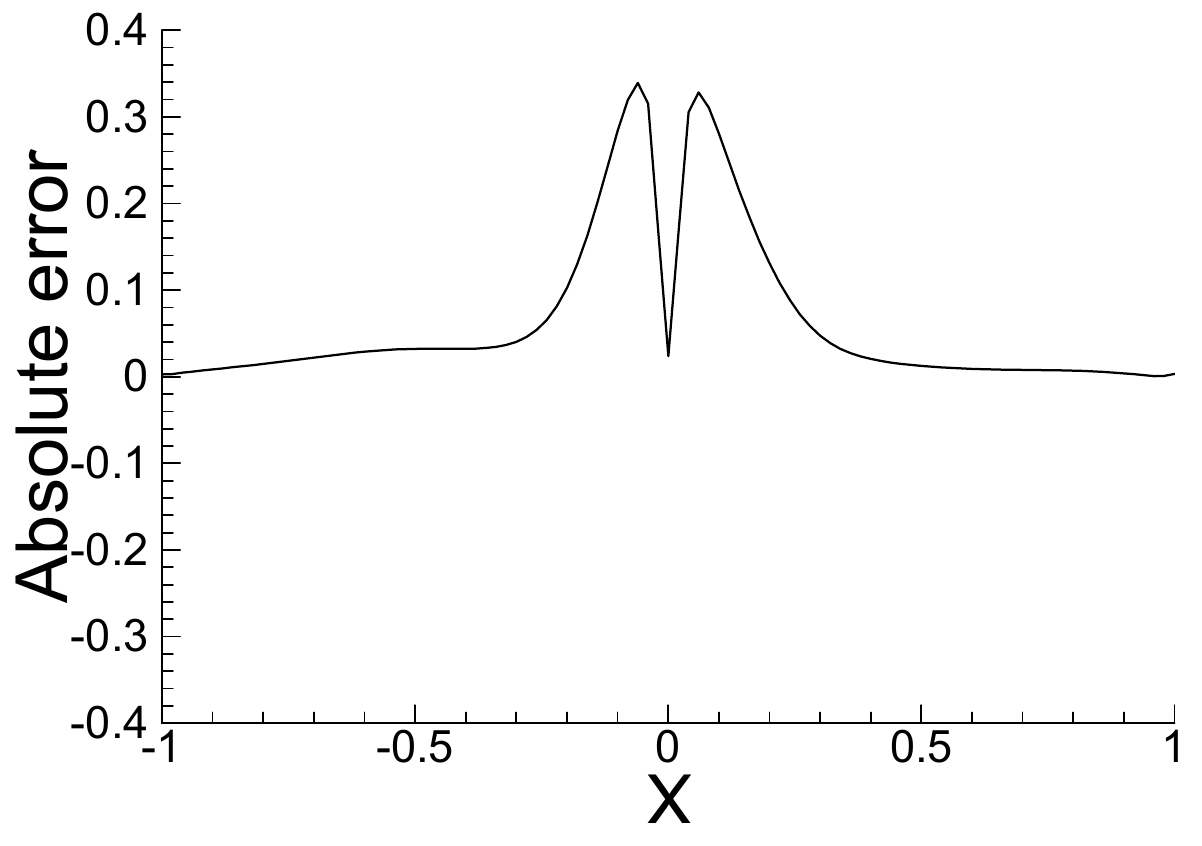}
        \caption{vPINN absolute error}
    \end{subfigure}
    \caption{
Comparison of BEKAN, EvoKAN, and EDNN for solving the 2D Burgers' equation (Eq.~\eqref{eq:Burgers_equation}) with boundary conditions $u(\pm1, y; t) = u(x, \pm1; t) = 0$ at $t = \SI{5e-1}{}$. 
Figures~(a)–(d) show the predicted solutions, and Figs.~(e)–(h) present the corresponding absolute errors. 
As the solution develops sharper features, both EDNN and EvoKAN exhibit increased errors, along with noticeable oscillations near the boundaries, as shown in the zoomed-in plots. 
In contrast, BEKAN maintains stable predictions even in regions with steep gradients and near the domain boundaries, demonstrating the best performance among the models.
}
    \label{fig:Burgers_slice_0.5}
\end{figure}

To facilitate a more intuitive comparison of the errors, we perform a slice cut at $y = -0.25$, where the steepest solution profile develops, and plot the results at $t = \SI{2e-1}{}$ and $t = \SI{5e-1}{}$ in Figs.~\ref{fig:Burgers_slice_0.2} and \ref{fig:Burgers_slice_0.5}, respectively.  
Figures.~\ref{fig:Burgers_slice_0.2}a, b, c, and d show the predicted solutions from BEKAN, \ac{EDNN}, \ac{EvoKAN}, and vanilla \ac{PINN}, respectively, while Figs.~\ref{fig:Burgers_slice_0.2}e, f, g, and h display the corresponding absolute error distributions at $t = \SI{2e-1}{}$.  
At $t = \SI{2e-1}{}$, where the steep central profile is not yet prominent, the three models BEKAN, \ac{EDNN}, and \ac{EvoKAN} accurately capture the solution.  
% However, from the zoomed-in plots, it can be seen that BEKAN nearly overlaps with the ground truth, whereas \ac{EDNN} and \ac{EvoKAN} exhibit noticeble errors.  
However, the zoomed-in plots reveal that BEKAN closely aligns with the ground truth, while \ac{EDNN} and \ac{EvoKAN} exhibit noticeable errors.
The vanilla \ac{PINN} shows a larger error near the center compared to the other three methods.

We plot the predicted solutions at \( t = \SI{5e-1}{} \) from BEKAN, \ac{EDNN}, \ac{EvoKAN}, and vanilla \ac{PINN} in Figs.~\ref{fig:Burgers_slice_0.5}a, b, c, and d. For accuracy evaluation, the corresponding absolute error distributions are shown in Figs.~\ref{fig:Burgers_slice_0.5}e, f, g, and h, respectively.  
At this time, a steep gradient develops near the center of the domain. As shown in Figs.~\ref{fig:Burgers_slice_0.5}b, c, and d, the errors increase for \ac{EDNN}, \ac{EvoKAN}, and vanilla \ac{PINN}. In contrast, BEKAN produces results that closely overlap with the FDM solution and remains accurate without oscillations. As shown in the absolute error distributions in Figs.~\ref{fig:Burgers_slice_0.5}e, f, g, and h, BEKAN exhibits the smallest error among BEKAN, \ac{EDNN}, \ac{EvoKAN}, and vanilla \ac{PINN}.

\begin{figure}[htbp]
    \centering
    \includegraphics[width=0.4\linewidth]{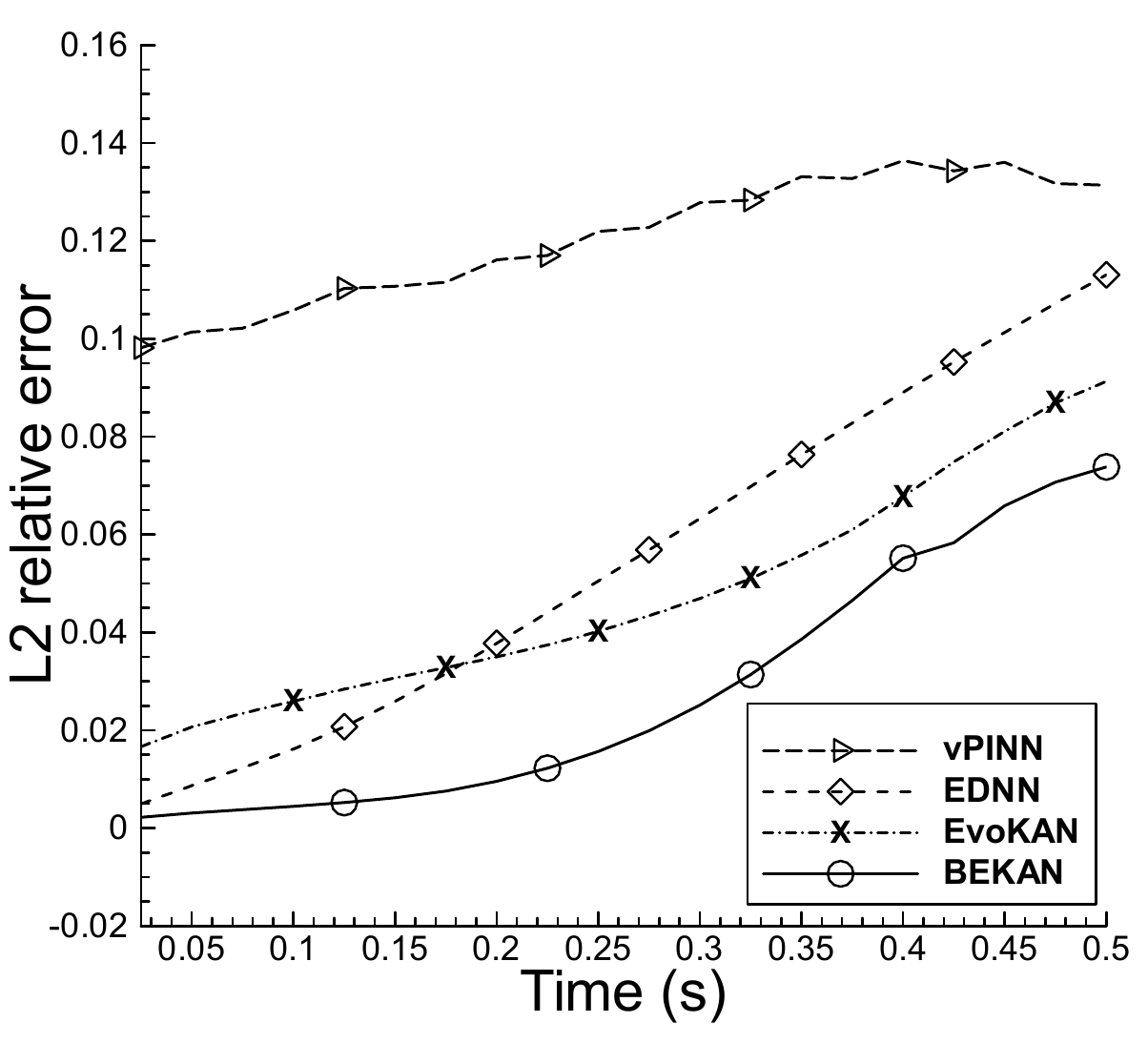}
    \caption{
Time evolution of the $L_2$ relative error for four models: BEKAN, EvoKAN, EDNN, and vanilla PINN, in solving the 2D Burgers' equation (Eq.~\eqref{eq:Burgers_equation}).  
The relative error at each time step is computed using the FDM solution as the reference.  
Among the models, BEKAN consistently shows the lowest $L_2$ relative error, indicating the highest accuracy.
}
    \label{fig:Burgers_L2}
\end{figure}

To evaluate the error over the entire time interval, we plot the $L_2$ relative error in Fig.~\ref{fig:Burgers_L2}. 
Compared to the one-dimensional Allen--Cahn example, all models show a tendency for the error to increase over time. 
Among them, BEKAN exhibits the lowest error, maintaining the smallest $L_2$ relative error throughout the entire time range.
To assess not only the overall error but also the satisfaction of the boundary conditions, Table~\ref{table:Burgers_BC} reports the solution values at the domain boundaries.
The vanilla \ac{PINN}, which adopts a soft constraint approach, does not strictly satisfy the zero boundary condition.
In contrast, BEKAN, \ac{EvoKAN}, and \ac{EDNN}, which enforce hard constraints, produce zero boundary values, thereby exactly satisfying the boundary conditions.

\begin{table}[htbp!]
\footnotesize
\renewcommand{\arraystretch}{1.0}
\centering
\caption{2D Burgers' equation (Eq.~\eqref{eq:Burgers_equation}): Predicted values of the solution $u(x,t)$ at the domain boundaries ($x, y = \pm 1$) obtained from BEKAN, EvoKAN, EDNN, and vanilla PINN at $t = \SI{2e-1}{}$ and $\SI{5e-1}{}$. 
The results are evaluated in terms of how well they satisfy the homogeneous boundary condition given in Eq.~\eqref{Eq:Burgers_equation_BC}. 
BEKAN applies the proposed approach for Dirichlet boundary enforcement described in Sec.~\ref{subsec:Dirichlet}, while EvoKAN and EDNN utilize output transformation techniques to impose hard constraints~\cite{guyiqi}.
}
\begin{tabular}{l c c c c c}
\hline
 & BEKAN & EvoKAN & EDNN & Vanilla PINN & Exact solution \\
\hline
$u(x=-1,\, y=-1, t=0.2)$ & $0.00000\mathrm{e}{+00}$ & $0.00000\mathrm{e}{+00}$ & $0.00000\mathrm{e}{+00}$ & $1.49512\mathrm{e}{-02}$ & $0.00000\mathrm{e}{+00}$ \\
$u(x=1,\,y=-1,  t=0.2)$  & $0.00000\mathrm{e}{+00}$ & $0.00000\mathrm{e}{+00}$ & $0.00000\mathrm{e}{+00}$ & $2.59334\mathrm{e}{-02}$ & $0.00000\mathrm{e}{+00}$ \\
$u(x=-1,\, y=1, t=0.2)$ & $0.00000\mathrm{e}{+00}$ & $0.00000\mathrm{e}{+00}$ & $0.00000\mathrm{e}{+00}$ & $6.57827\mathrm{e}{-03}$ & $0.00000\mathrm{e}{+00}$ \\
$u(x=1,\,y=1,  t=0.2)$  & $0.00000\mathrm{e}{+00}$ & $0.00000\mathrm{e}{+00}$ & $0.00000\mathrm{e}{+00}$ & $1.30068\mathrm{e}{-02}$ & $0.00000\mathrm{e}{+00}$ \\
$u(x=-1,\,y=-1, t=0.5)$ & $0.00000\mathrm{e}{+00}$ & $0.00000\mathrm{e}{+00}$ & $0.00000\mathrm{e}{+00}$ & $1.69440\mathrm{e}{-02}$ & $0.00000\mathrm{e}{+00}$ \\
$u(x=1,\,y=-1, t=0.5)$  & $0.00000\mathrm{e}{+00}$ & $0.00000\mathrm{e}{+00}$ & $0.00000\mathrm{e}{+00}$ & $3.56581\mathrm{e}{-02}$ & $0.00000\mathrm{e}{+00}$ \\
$u(x=-1,\, y=1, t=0.5)$ & $0.00000\mathrm{e}{+00}$ & $0.00000\mathrm{e}{+00}$ & $0.00000\mathrm{e}{+00}$ & $6.03220\mathrm{e}{-02}$ & $0.00000\mathrm{e}{+00}$ \\
$u(x=1,\,y=1,  t=0.5)$  & $0.00000\mathrm{e}{+00}$ & $0.00000\mathrm{e}{+00}$ & $0.00000\mathrm{e}{+00}$ & $5.41353\mathrm{e}{-02}$ & $0.00000\mathrm{e}{+00}$ \\
\hline
\end{tabular}
\label{table:Burgers_BC}
\end{table}

% ============================
% subsection: Kuramoto–Sivashinsky Equation
% ============================
\subsection{Kuramoto–Sivashinsky equation with Periodic Boundary Condition}
\label{sec. 4.3: KS-equation}

The one-dimensional Kuramoto–Sivashinsky (KS) equation is a nonlinear PDE that serves as a canonical model for capturing spatiotemporal instabilities observed in systems such as laminar flame fronts and chaotic flows. It is formulated as
\begin{equation}
\label{eq:KS_equation}
\frac{\partial u}{\partial t} + u\frac{\partial u}{\partial x} + \frac{\partial^2 u}{\partial x^2} + \frac{\partial^4 u}{\partial x^4} = 0,
\end{equation}
subject to the corresponding boundary and initial specifications:
\begin{align}
\label{eq:KS_BC_1}
u(0, t) &= u(100, t), \\
\label{eq:KS_BC_2}
u_x(0, t) &= u_x(100, t), \\
u(x, 0) &= \sin\left(\frac{16\pi x}{100}\right).
\end{align}

The KS equation includes nonlinear advection, diffusion, and hyper-diffusion terms, leading to complex dynamical behavior such as periodic and quasi-periodic patterns. In this study, we impose periodic boundary conditions to ensure continuity of the field variable \( u(x, t) \) and its corresponding spatial gradients at the domain boundaries. Specifically, we employ a periodic layer composed of two sinusoidal functions, \( \sin\left( \frac{2\pi x}{100} \right) \) and \( \cos\left( \frac{2\pi x}{100} \right) \).

\begin{figure}[htbp!]
    \centering
    \includegraphics[width=0.5\linewidth]{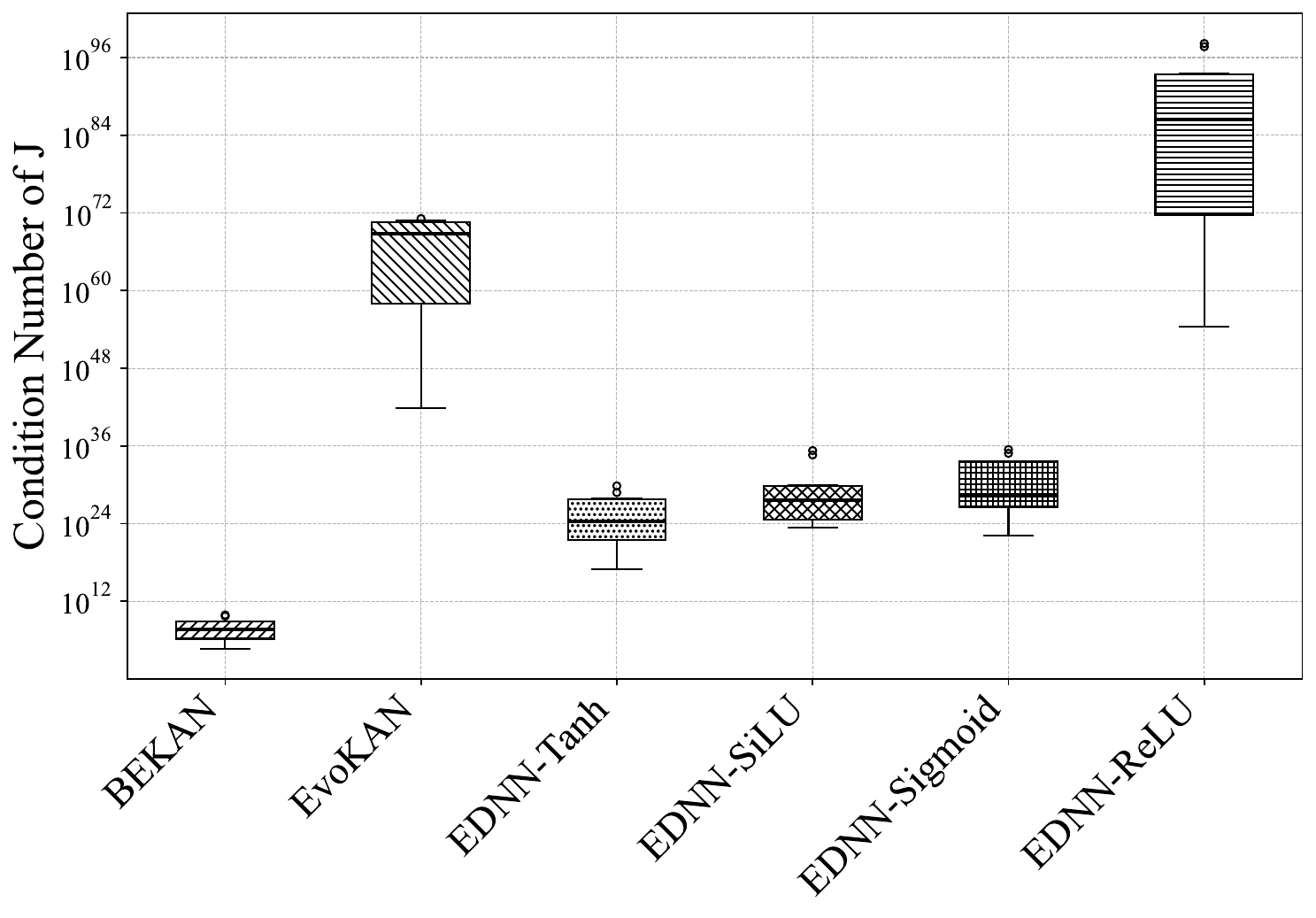}
    \caption{
1D KS equation (Eq.~\eqref{eq:KS_equation}): Box plot of the Jacobian condition numbers of matrix $J$ during parameter evolution across 10 simulations. We calculate the condition number of $J$ in the first iteration in the evolutionary process. The Jacobian$\left( \mathbf{J} \right)_{ij} = \frac{\partial \hat{u}^i}{\partial {W}_j}$ quantifies the influence of parameter changes on the predicted solution.
The strong sensitivity induced by the chaotic behavior of the KS equation can cause $J$ to become ill-conditioned, hindering stable parameter updates. While B-spline-based EvoKAN and EDNNs suffer from ill-conditioning, the Gaussian RBF-based BEKAN handles the KS equation robustly for multiple training without failure of the parameter evolution.
}
    \label{fig:condition}
\end{figure}

We describe the training configuration for the KS equation in Table~\ref{table:KS_training}.  
For the evolutionary methods, training is carried out with a time increment of \( t = \SI{1e-2}{} \), while the vanilla \ac{PINN} is trained across the entire time domain without iteration.
For the KS equation, BEKAN converged during the parameter evolution process, whereas \ac{EvoKAN} and \ac{EDNN} failed to converge due to suffering from ill-conditioning while solving the least-squares problem in Eq.~\eqref{OptCondApprox}. To examine this quantitatively, we computed the Jacobian matrix entries defined by $\left( \mathbf{J} \right)_{ij} = \frac{\partial \hat{u}^i}{\partial {W}_j}$ for BEKAN, \ac{EvoKAN}, and \ac{EDNN}, and visualized the condition numbers using box plots in Fig.~\ref{fig:condition}. 
In the case of \ac{EDNN}, we test different activation functions including tanh, SiLU, Sigmoid, and ReLU, and compute the condition number of the Jacobian at the third iteration step of parameter evolution.
As shown in Fig.~\ref{fig:condition}, BEKAN yielded the smallest condition numbers, while the other models exhibited significantly larger values.

\sisetup{group-separator={,}, group-minimum-digits=4}
\begin{table} [hbt!]
\footnotesize 
	\renewcommand{\arraystretch}{1.0}
	\begin{center} 
		\caption{Training configuration for the 1D KS equation (Eq.~\eqref{eq:KS_equation}).
}
		\begin{tabular}{l c c c c}
			\hline
			{\, \, \, } & \makecell[c]{BEKAN} & \makecell[c]{EvoKAN} & \makecell[c]{EDNN} & {Vanilla PINN} \\
			\hline
			{Hidden layers} & {[3, 3, 3, 3]} & \makecell[c]{[3, 3, 3, 3]} & {[10, 10, 10, 10]} & {[10, 10, 10, 10]}\\
            % {} & {} & {} & {} & {[90, 90, 90, 90]}\\
            % {} & {[7, 7, 7, 7]} & \makecell[c]{[7, 7, 7, 7]} & {[20, 20, 20, 20]} & {[20, 20, 20, 20]}\\
            {Activation functions} & \makecell[c]{Gaussian RBFs/SiLU} & \makecell[c]{B-splines/SiLU} & {tanh} & {tanh} \\
            % \hline
            \makecell[l]{Grid points number \\ of activation functions} & {8} & \makecell[c]{8} & {-} & {-}\\
			\makecell[l]{Number of \\ trainable parameters} & {\SI{210}{}} & {\SI{360}{}} & {\SI{361}{}} & \makecell[c]{\SI{361}{}}\\
            {Optimizer} & \makecell[c]{Adam} & \makecell[c]{Adam} & \makecell[c]{Adam} & \makecell[c]{Adam/L-BFGS-B}\\
            {Timestep} & \makecell[c]{1e-02} & \makecell[c]{1e-02} & \makecell[c]{1e-02} & \makecell[c]{-}\\
			\hline
		\end{tabular}
		\label{table:KS_training}
	\end{center}
\end{table}

% The original KAN encounters significant challenges in this test, primarily due to its inability to accurately compute high-order derivatives through automatic differentiation. The use of spline basis functions, while effective in simpler scenarios, falls short in capturing the complex dynamics and high-order derivative effects present in the KS equation. This results in solutions that fail to reflect the true behavior of the system. 

% Contrary to the original KAN, RadialKAN excels in this test, providing accurate solutions that vividly capture the unique spatiotemporal patterns characteristic of the KS equation. These patterns, as illustrated in the accompanying figure, show intricate wave-like structures that are crucial for understanding the dynamics of the system. The effectiveness of RadialKAN in this context can be attributed to its superior handling of automatic differentiation, especially in terms of high-order derivatives. This advancement opens the door to tackling more complex and challenging problems within the field of dynamical systems and turbulence research.

\begin{figure}[htbp!]
    \centering
    \includegraphics[width=0.45\linewidth]{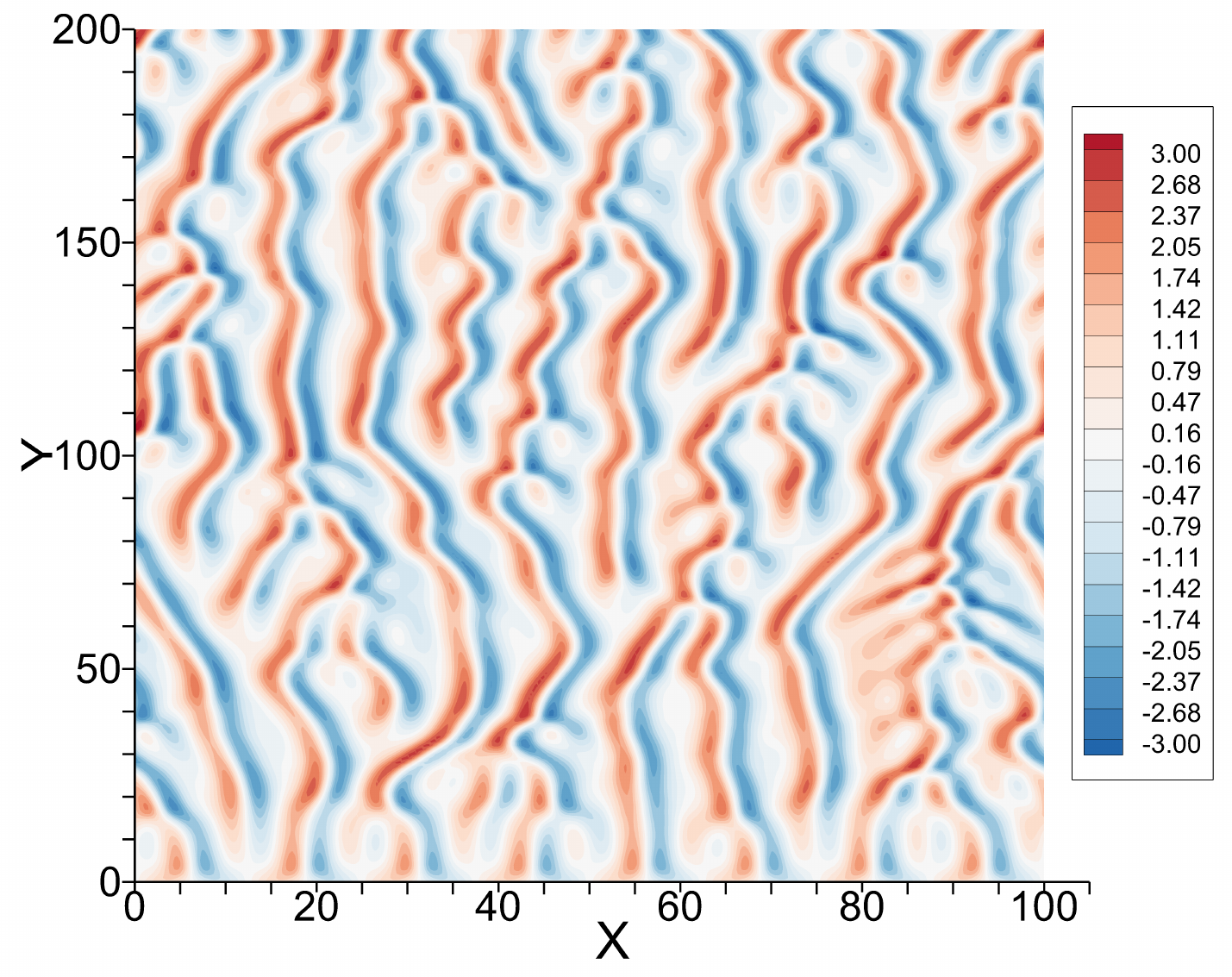}
    \caption{
1D KS equation (Eq.~\eqref{eq:KS_equation}): Predicted distribution by BEKAN. The x-axis represents the spatial domain, while the y-axis corresponds to time.
}
\label{fig:KS_distribution}
\end{figure}

Figure~\ref{fig:KS_distribution} shows the converged solution obtained by BEKAN. The figure displays spatial location \( x \) along the horizontal axis and time \( t \) along the vertical axis.
As time evolves from \( t = 0 \), the initial sinusoidal profile \( \sin\left(\frac{16 \pi x}{100} \right) \) develops into a disordered state, demonstrating chaotic dynamics.  

\begin{figure}[htbp!]
    \centering
    \begin{tabular}{cc}
        \begin{subfigure}[b]{0.43\linewidth}
            \centering
            \includegraphics[width=\linewidth]{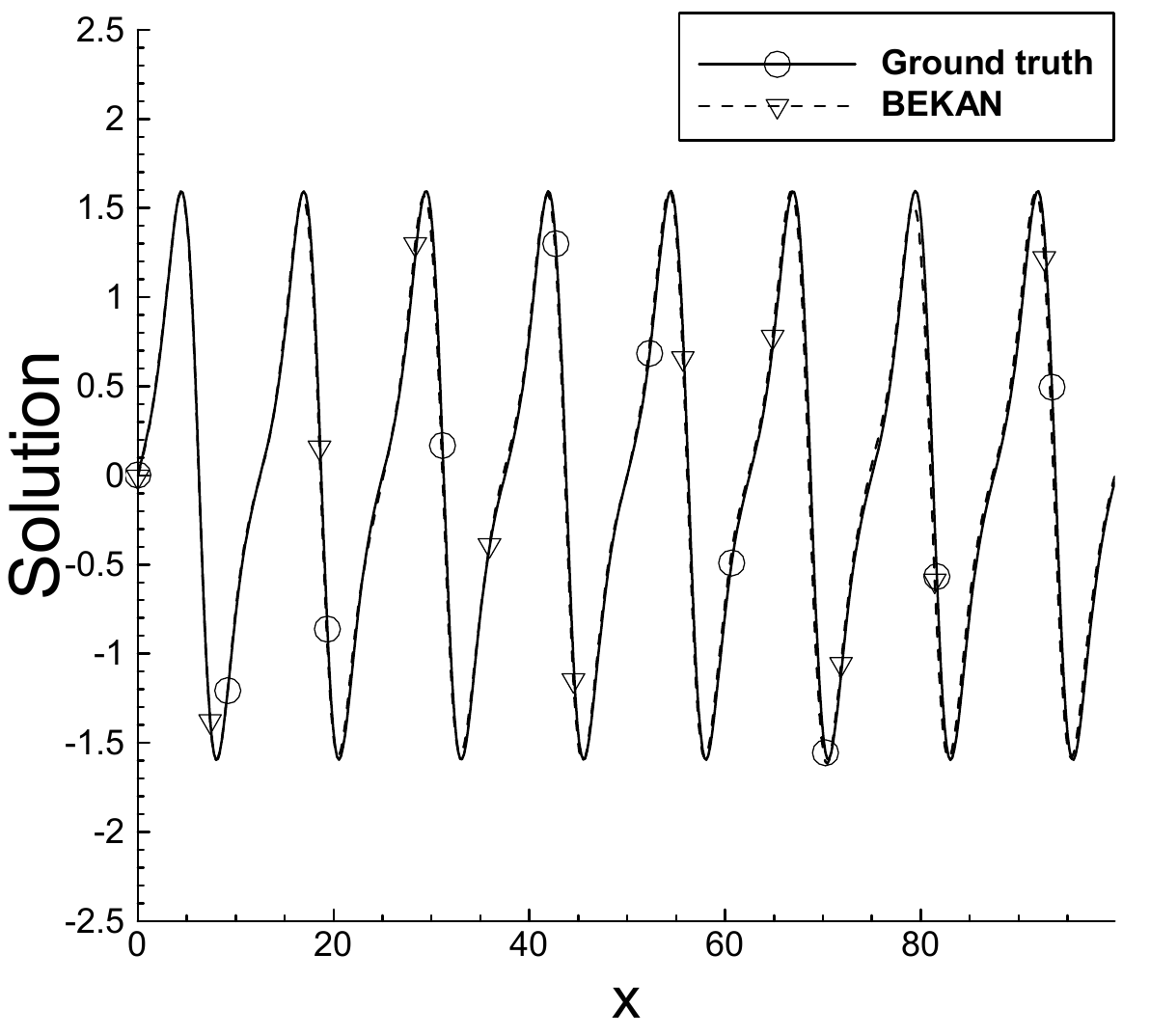}
            \caption{BEKAN solution}
        \end{subfigure} &
        \begin{subfigure}[b]{0.43\linewidth}
            \centering
            \includegraphics[width=\linewidth]{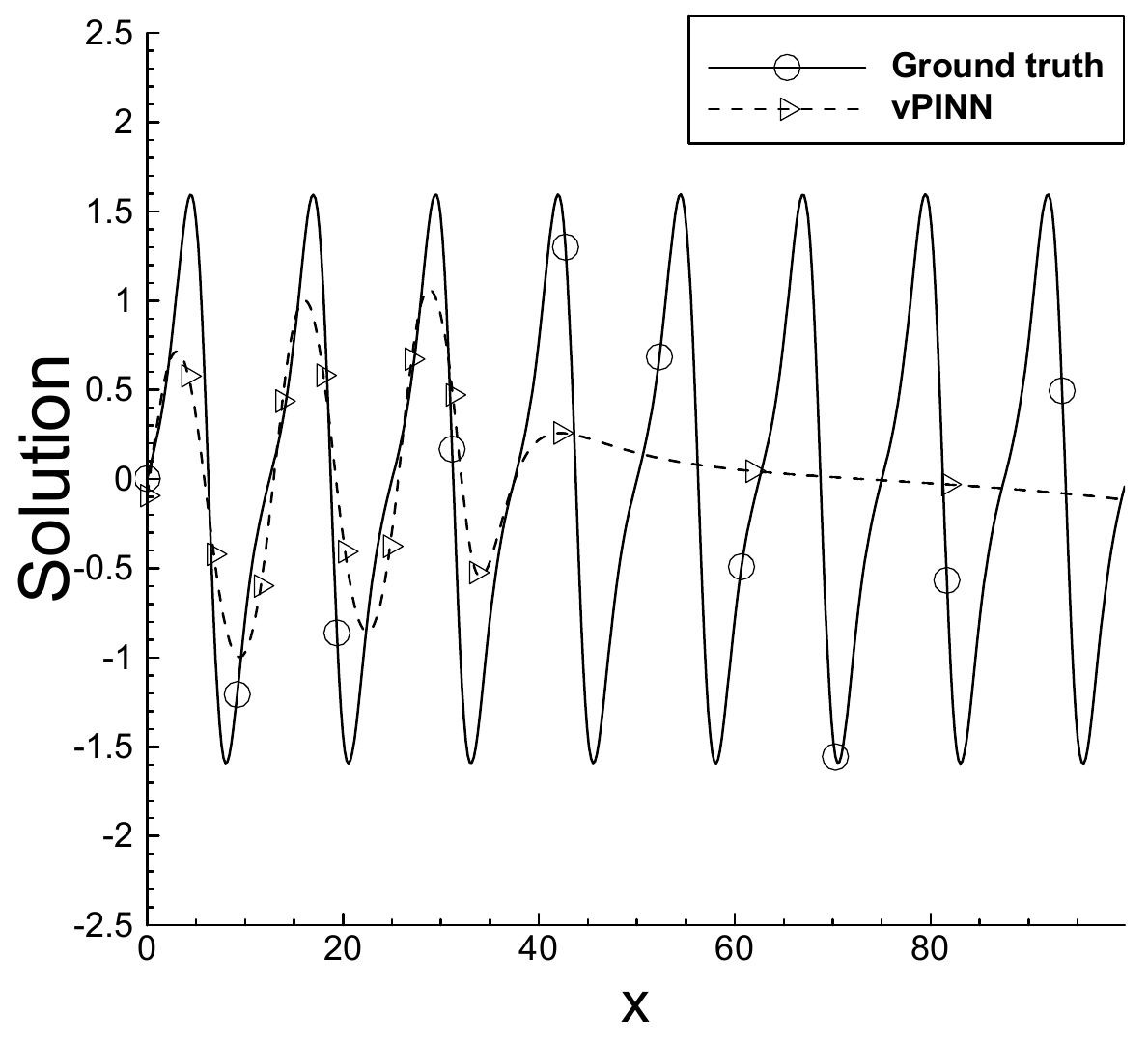}
            \caption{vPINN solution}
        \end{subfigure} \\
        \begin{subfigure}[b]{0.43\linewidth}
            \centering
            \includegraphics[width=\linewidth]{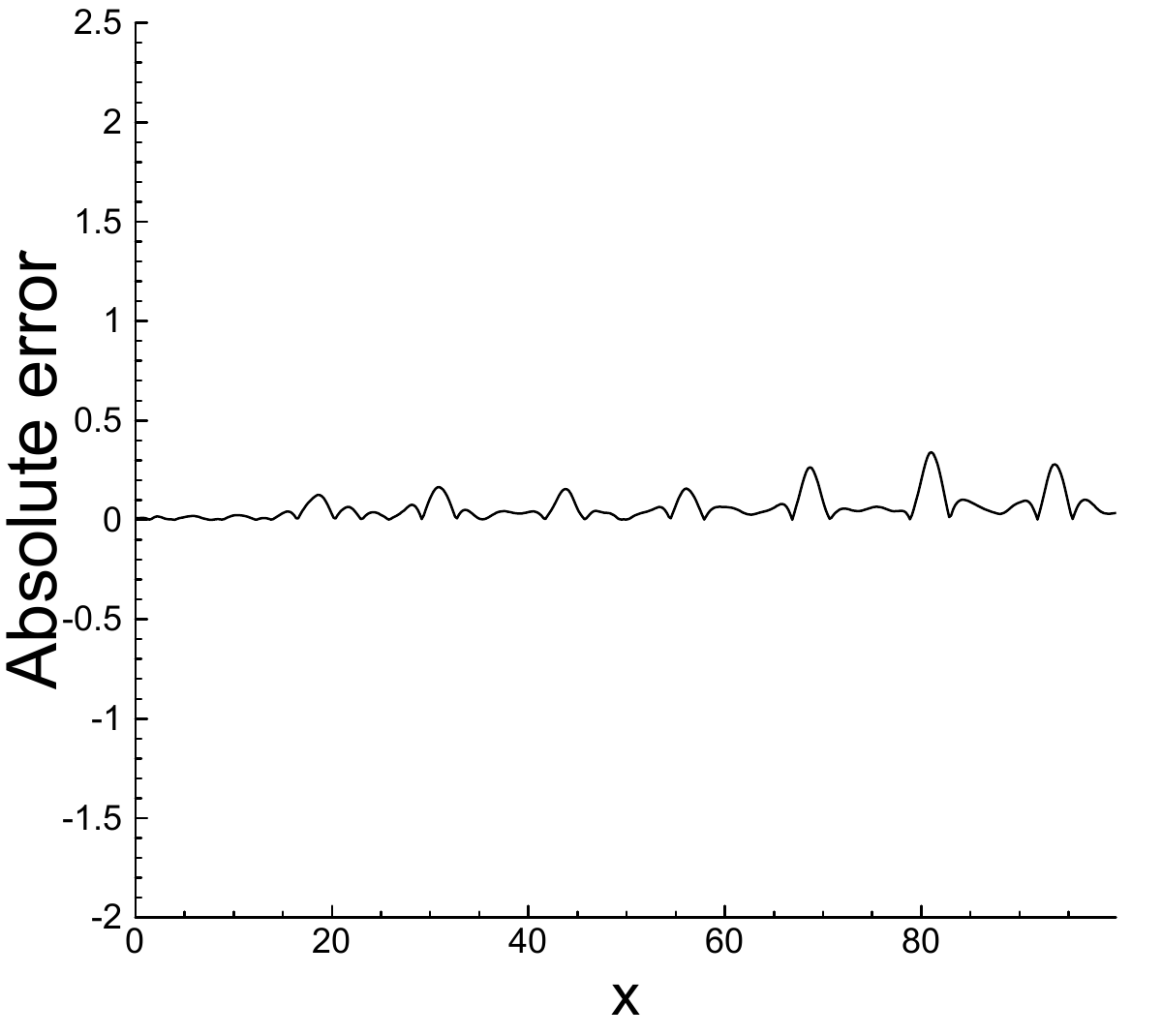}
            \caption{BEKAN absolute error}
        \end{subfigure} &
        \begin{subfigure}[b]{0.43\linewidth}
            \centering
            \includegraphics[width=\linewidth]{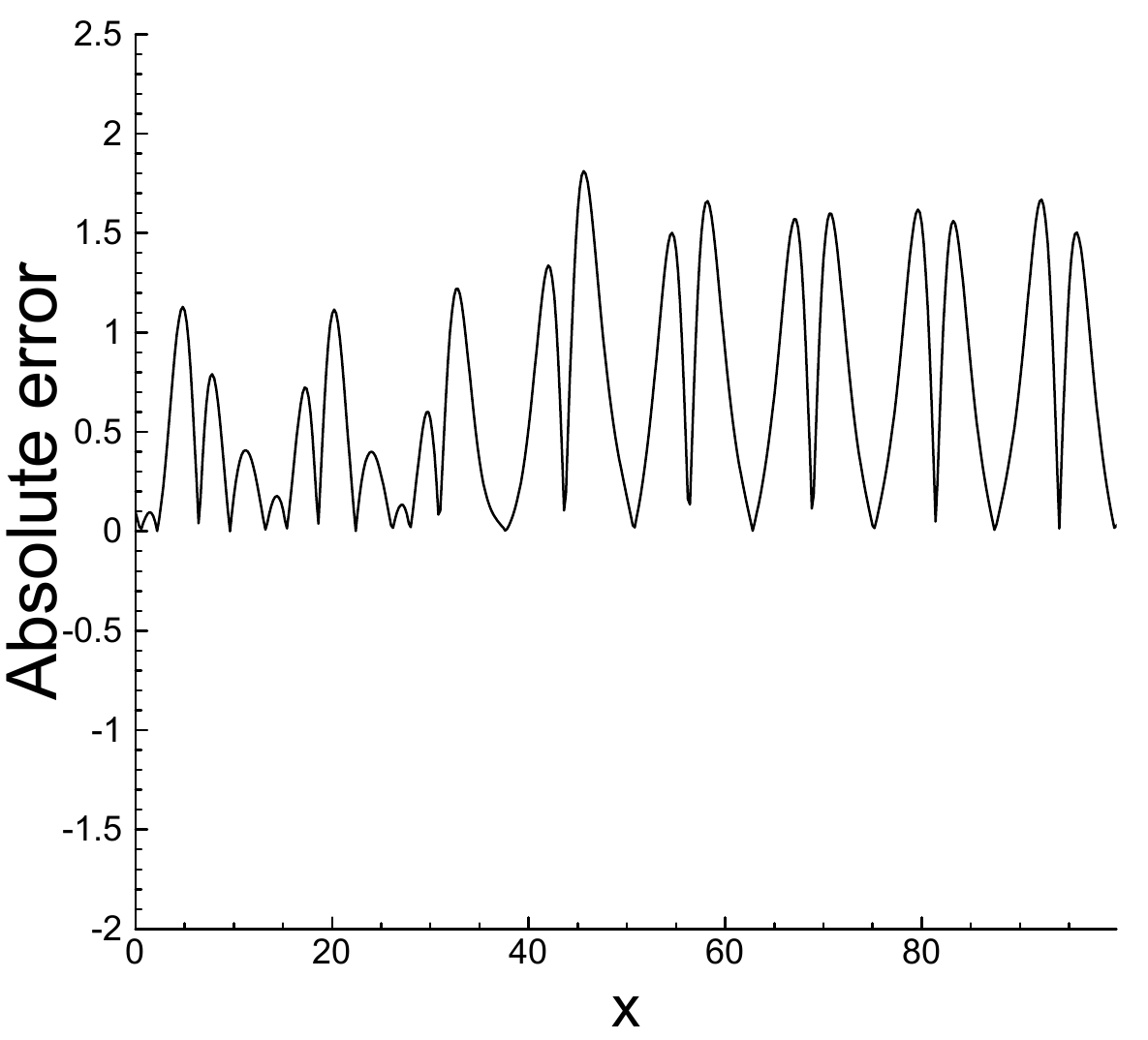}
            \caption{vPINN absolute error}
        \end{subfigure}
    \end{tabular}
    \caption{1D KS equation (Eq.~\eqref{eq:KS_equation}): Comparison of the predicted solution \( u \) at \( t = 2 \) between BEKAN and vPINN.  
Subfigures (a) and (b) display the predicted solutions, while (c) and (d) show the corresponding absolute errors with respect to the spectral reference solution.  
Due to ill-conditioning of the Jacobian matrix during the parameter evolution process, EDNN and EvoKAN failed, and thus only BEKAN and vPINN are included in the comparison plots.
The prediction by BEKAN closely overlaps with the reference solution, outperforming the vPINN.
}
    \label{fig:KS_t_2}
\end{figure}

To evaluate accuracy, we generated the ground truth using the spectral method and plotted the BEKAN and vanilla PINN solution at \( t = 2 \) along with its corresponding absolute error distribution in Fig.~\ref{fig:KS_t_2}.
As shown in Fig.~\ref{fig:KS_t_2}a, the BEKAN solution at \( t = 2 \) closely overlaps with the ground truth. The corresponding absolute error distribution in Fig.~\ref{fig:KS_t_2}c also confirms that the error remains small. In contrast, the vanilla \ac{PINN} fails to capture the ground truth accurately, as illustrated in Fig.~\ref{fig:KS_t_2}b, resulting in significantly larger errors as seen in Fig.~\ref{fig:KS_t_2}d.

\begin{figure}[tbp!]
    \centering
    \begin{tabular}{cc}
        \begin{subfigure}[b]{0.43\linewidth}
            \centering
            \includegraphics[width=\linewidth]{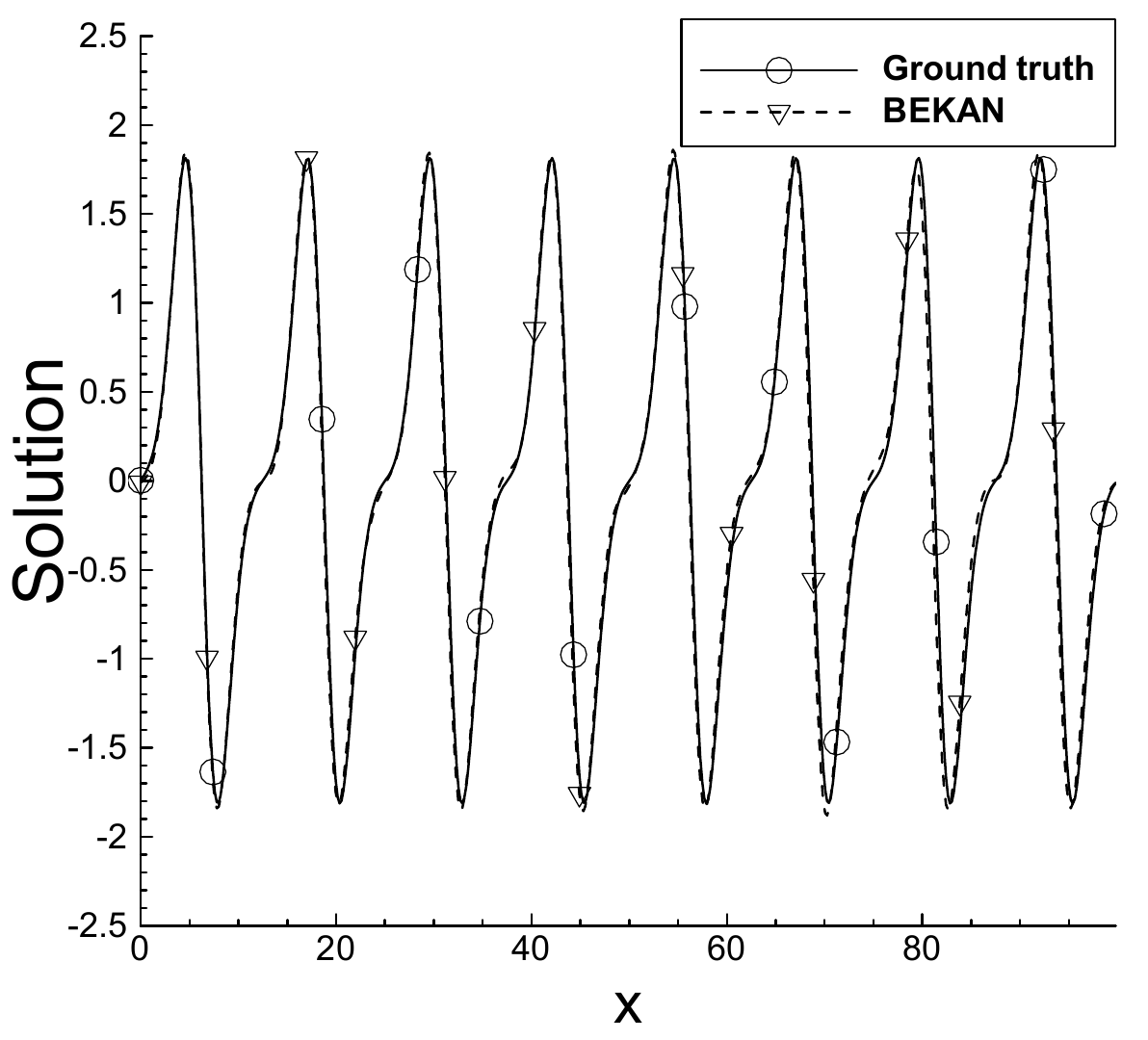}
            \caption{BEKAN solution}
        \end{subfigure} &
        \begin{subfigure}[b]{0.43\linewidth}
            \centering
            \includegraphics[width=\linewidth]{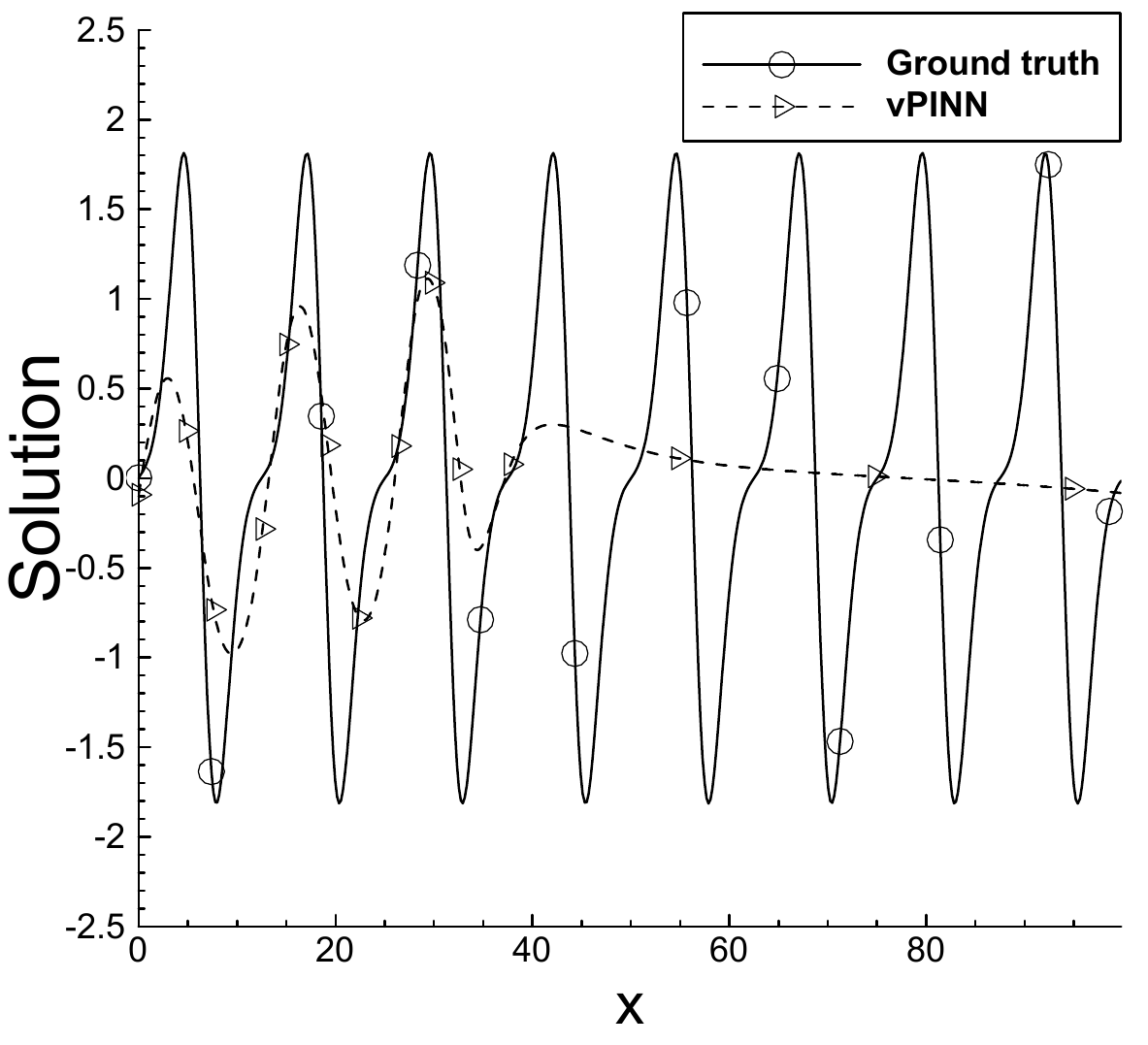}
            \caption{PINN solution}
        \end{subfigure} \\
        \begin{subfigure}[b]{0.43\linewidth}
            \centering
            \includegraphics[width=\linewidth]{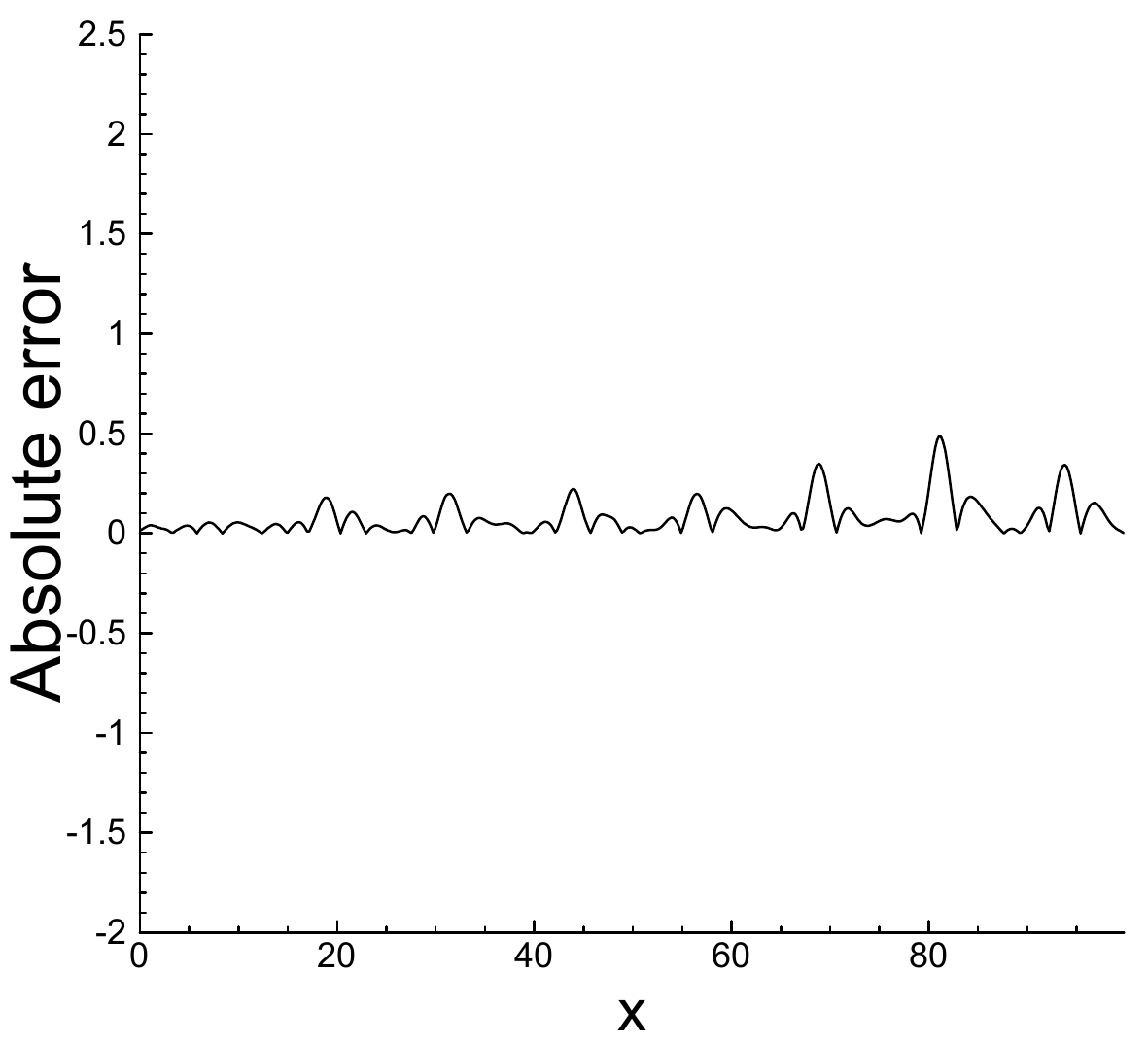}
            \caption{BEKAN absolute error}
        \end{subfigure} &
        \begin{subfigure}[b]{0.43\linewidth}
            \centering
            \includegraphics[width=\linewidth]{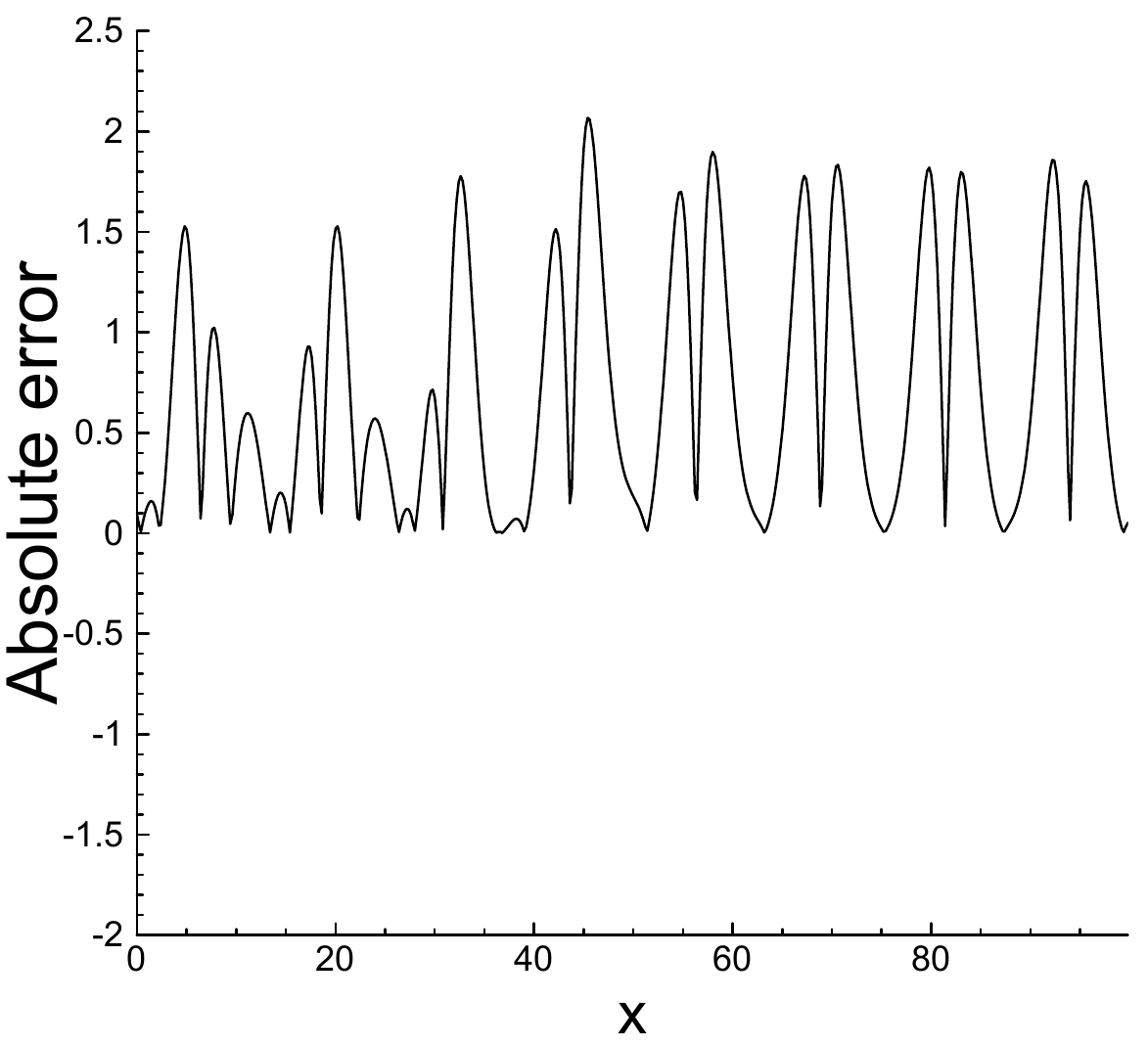}
            \caption{vPINN absolute error}
        \end{subfigure}
    \end{tabular}
    \caption{1D KS equation (Eq.~\eqref{eq:KS_equation}): A comparative analysis is conducted between BEKAN and vPINN for the predicted solution \( u \) at \( t = 3 \).  
Subfigures (a) and (b) present the predicted solutions, whereas (c) and (d) show the corresponding absolute errors with respect to the spectral reference solution.  
EDNN and EvoKAN are excluded from the comparison due to numerical instability arising from Jacobian ill-conditioning during the parameter evolution procedure.  
In this comparison, the solution obtained by BEKAN exhibits agreement with the spectral reference, demonstrating higher accuracy than vPINN.
}
\label{fig:KS_t_3}
\end{figure}

We also computed the ground truth at \( t = 3 \) using the spectral method and compared it with the BEKAN and vanilla \ac{PINN} predictions in Fig.~\ref{fig:KS_t_3}.  
In Fig.~\ref{fig:KS_t_3}a, The BEKAN solution exhibits strong concordance with the reference data, and the corresponding absolute error depicted in Fig.~\ref{fig:KS_t_3}c remains consistently low throughout the domain. 
On the other hand, the vanilla \ac{PINN} deviates from the ground truth, as shown in Fig.~\ref{fig:KS_t_3}b, resulting in larger errors depicted in Fig.~\ref{fig:KS_t_3}d.

\begin{table}[htbp!]

\footnotesize
\renewcommand{\arraystretch}{1.0}
\centering
\caption{1D KS equation (Eq.~\eqref{eq:KS_equation}): The predicted values of $|u(0,t) - u(100,t)|$ and $|u_x(0,t) - u_x(100,t)|$ are evaluated to assess compliance with the periodic boundary conditions specified in Eqs.~\eqref{eq:KS_BC_1} and \eqref{eq:KS_BC_2}.
EDNN and EvoKAN failed to converge due to ill-conditioning during the parameter evolution process, and their results are marked as NaN in the table.
} 
\begin{tabular}{l c c c c c}
\hline
 & BEKAN & Vanilla PINN & EDNN & EvoKAN &Exact values \\
\hline
$|u(0, 0.1) - u(100,0.1)|$ & $0.00000\mathrm{e}{+00}$ & $3.96012\mathrm{e}{-03}$ & NaN & NaN & $0.00000\mathrm{e}{+00}$ \\
$|u_x(0, 0.1) - u_x(100,0.1)|$  & $0.00000\mathrm{e}{+00}$ & $5.83714\mathrm{e}{-01}$ & NaN & NaN & $0.00000\mathrm{e}{+00}$ \\
$|u(0, 2) - u(100,2)|$ & $0.00000\mathrm{e}{+00}$ & $2.20110\mathrm{e}{-01}$ & NaN & NaN & $0.00000\mathrm{e}{+00}$ \\
$|u_x(0, 2) - u_x(100,2)|$  & $0.00000\mathrm{e}{+00}$ & $4.96792\mathrm{e}{-01}$ & NaN & NaN & $0.00000\mathrm{e}{+00}$ \\
$|u(0, 3) - u(100,3)|$ & $0.00000\mathrm{e}{+00}$ & $1.49798\mathrm{e}{-02}$ & NaN & NaN & $0.00000\mathrm{e}{+00}$ \\
$|u_x(0, 3) - u_x(100,3)|$  & $0.00000\mathrm{e}{+00}$ & $4.40734\mathrm{e}{-01}$ & NaN & NaN & $0.00000\mathrm{e}{+00}$ \\
$|u(0, 100) - u(100,100)|$ & $0.00000\mathrm{e}{+00}$ & $7.77670\mathrm{e}{-04}$ & NaN & NaN & $0.00000\mathrm{e}{+00}$ \\
$|u_x(0, 100) - u_x(100,100)|$  & $0.00000\mathrm{e}{+00}$ & $2.34550\mathrm{e}{-04}$ & NaN & NaN & $0.00000\mathrm{e}{+00}$ \\
$|u(0, 200) - u(100,200)|$ & $0.00000\mathrm{e}{+00}$ & $3.95951\mathrm{e}{-03}$ & NaN & NaN & $0.00000\mathrm{e}{+00}$ \\
$|u_x(0, 200) - u_x(100,200)|$  & $0.00000\mathrm{e}{+00}$ & $5.83738\mathrm{e}{-01}$ & NaN & NaN & $0.00000\mathrm{e}{+00}$ \\
\hline
\label{table:periodic}
\end{tabular}
\end{table}

For quantitative assessment of periodic boundary condition satisfaction, we summarize the boundary values at selected time steps in Table~\ref{table:periodic}. Since \ac{EDNN} and \ac{EvoKAN} failed to converge, their values are denoted as NaN. The table includes results from BEKAN, vanilla \ac{PINN}, and the exact solution. The numerical results indicate that BEKAN satisfies the periodic boundary condition with exactness, while the vanilla \ac{PINN} exhibits a noticeable deviation from the boundary values.

% ============================
% subsection: Heat Equation
% ============================
\subsection{Heat Equation with Neumann Boundary Condition}
\label{subsec:heat}
The classical two-dimensional heat equation describes the diffusion of thermal energy within a medium, assuming purely conductive transport. To account for more complex physical phenomena, such as external forces or internal reactive dynamics, the classical heat equation is extended by incorporating a nonlinear forcing term.
The resulting equation is defined on the spatial domain \(\Omega = [-1, 1] \times [-1, 1]\) as:
\begin{equation}
\label{eq:heat_equation}
  \frac{\partial u}{\partial t}
  = \alpha \left( \frac{\partial^2 u}{\partial x^2} + \frac{\partial^2 u}{\partial y^2} \right)
    + u(1 - u),
  \quad (x,y)\in \Omega,\; t>0.
\end{equation}
In this test, the diffusion coefficient is set to \(\alpha = 1\). We impose the following initial condition:
\begin{equation}
  u(x, y, 0) = \cos(\pi x)\cos(\pi y),
  \quad (x,y)\in \Omega.
\end{equation}
We impose homogeneous Neumann boundary conditions to ensure no flux through the boundaries:
\begin{equation}
\label{eq:heat_equation_BC}
  u_x(-1, y, t) = u_x(1, y, t) = 0,
  \quad
  u_y(x, -1, t) = u_y(x, 1, t) = 0,
  \quad (x,y)\in \partial\Omega,\; t>0.
\end{equation}
Finally we define the energy functional as:
\begin{equation}
  E[u]
  = \iint_{\Omega} \frac{1}{2} \left| \nabla u(x,y,t) \right|^2 \,\mathrm{d}x\,\mathrm{d}y
  = \int_{-1}^{1}\!\!\int_{-1}^{1}
      \frac{1}{2} \left( u_x^2 + u_y^2 \right)
      \,\mathrm{d}x\,\mathrm{d}y.
\end{equation}

Details of the training setup adopted in this study are provided in Table~\ref{table:heat_training}.
BEKAN and \ac{EvoKAN} adopt the same hidden layer structure but differ in the type of basis functions.  
\ac{EvoKAN} uses B spline functions, which may use additional scaling parameters.  
This results in 600 trainable parameters in \ac{EvoKAN}, whereas BEKAN contains 352.  
For evolutionary models, training proceeds step by step with a time interval of \( t = \SI{5e-5}{} \).  
In contrast, the vanilla \ac{PINN} is trained over the entire time domain in a single stage.

\sisetup{group-separator={,}, group-minimum-digits=4}
\begin{table} [hbtp!]
\footnotesize 
	\renewcommand{\arraystretch}{1.0}
	\begin{center} 
		\caption{Training configuration for the 2D heat equation with nonlinear forcing term (Eq.~\eqref{eq:heat_equation}).
}
		\begin{tabular}{l c c c c}
			\hline
			{\, \, \, } & \makecell[c]{BEKAN} & \makecell[c]{EvoKAN} & \makecell[c]{EDNN} & {Vanilla PINN} \\
			\hline
			{Hidden layers} & {[4, 4, 4, 4]} & \makecell[c]{[4, 4, 4, 4]} & {[15, 15, 15]} & {[15, 15, 15]}\\
            {Activation functions} & \makecell[c]{Gaussian RBFs/SiLU} & \makecell[c]{B-splines/SiLU} & {tanh} & {tanh} \\
            % \hline
            \makecell[l]{Grid points number\\ of activation functions} & {5} & \makecell[c]{5} & {-} & {-}\\
			\makecell[l]{Number of \\ trainable parameters} & {\SI{352}{}} & {\SI{600}{}} & {\SI{541}{}} & \makecell[c]{\SI{541}{}}\\
            {Optimizer} & \makecell[c]{Adam} & \makecell[c]{Adam} & \makecell[c]{Adam} & \makecell[c]{Adam/L-BFGS-B}\\
            {Timestep} & \makecell[c]{5e-05} & \makecell[c]{5e-05} & \makecell[c]{5e-05} & \makecell[c]{-}\\
			\hline
		\end{tabular}
		\label{table:heat_training}
	\end{center}
\end{table}

\begin{figure}[htbp!]
    \centering \,
    \begin{subfigure}[b]{0.24\linewidth}
        \centering
        \includegraphics[width=\linewidth]{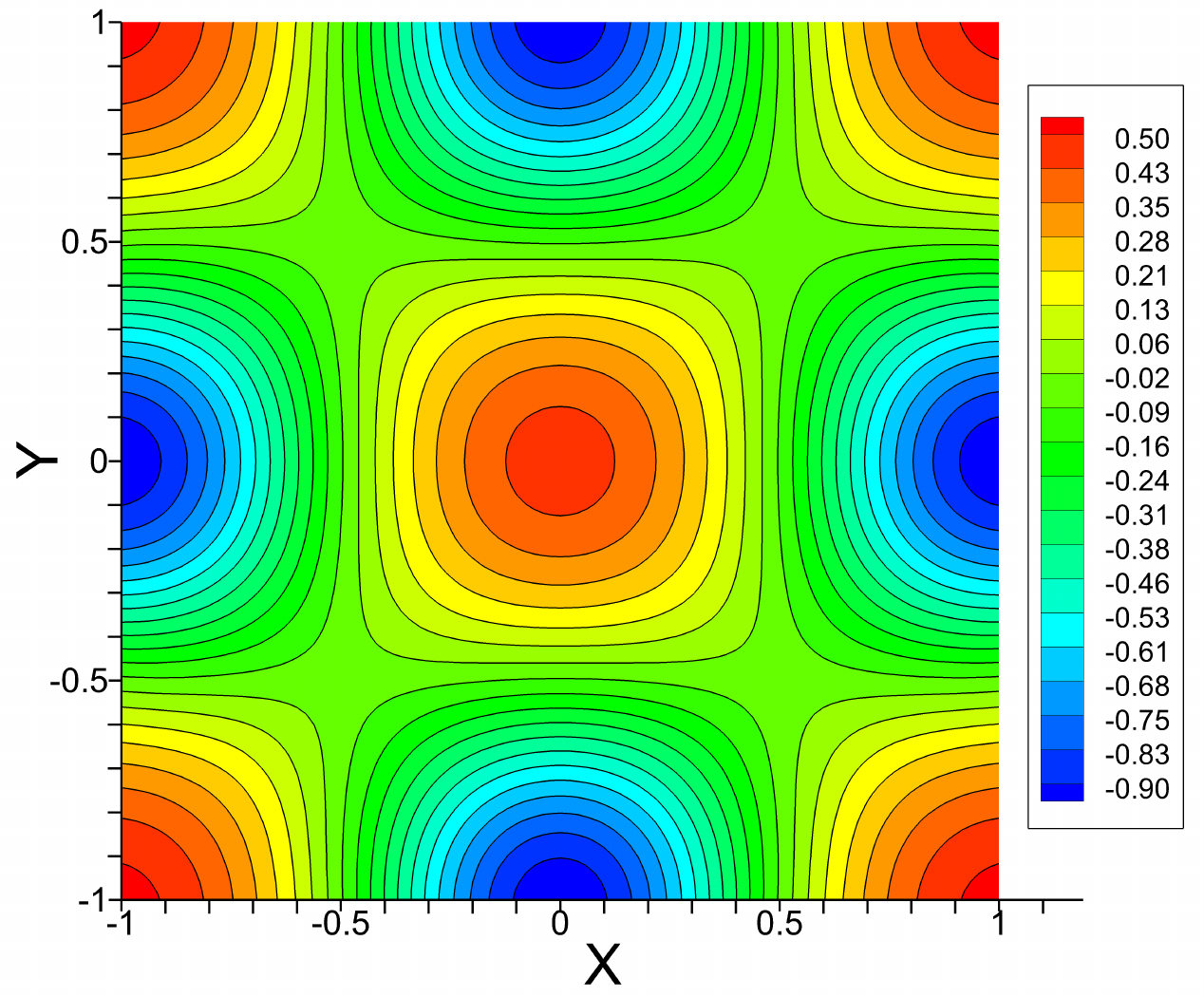}
        \caption{BEKAN solution}
    \end{subfigure}
    \hfill
    \begin{subfigure}[b]{0.24\linewidth}
        \centering
        \includegraphics[width=\linewidth]{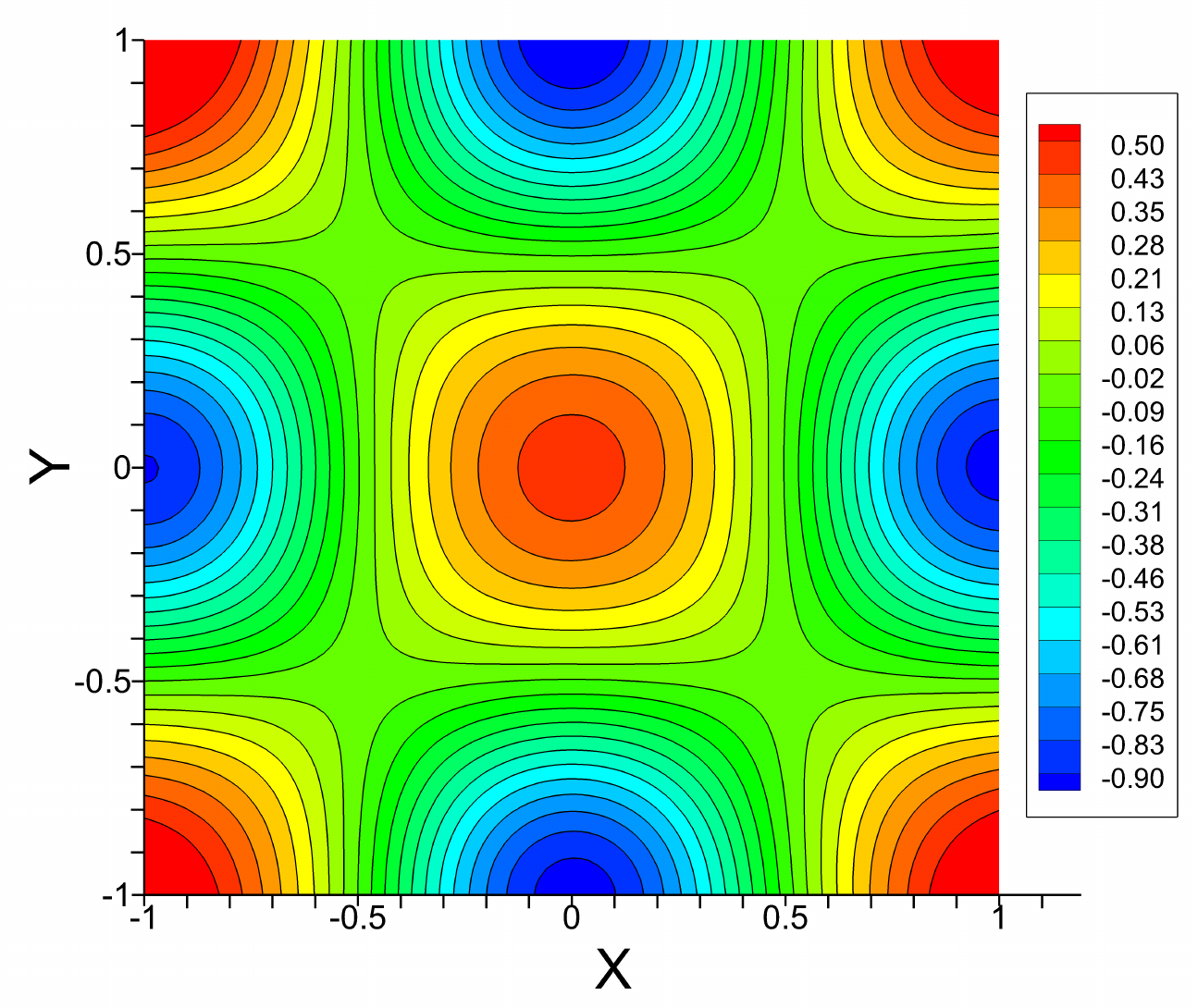}
        \caption{EDNN solution}
    \end{subfigure}
    \hfill
    \begin{subfigure}[b]{0.24\linewidth}
        \centering
        \includegraphics[width=\linewidth]{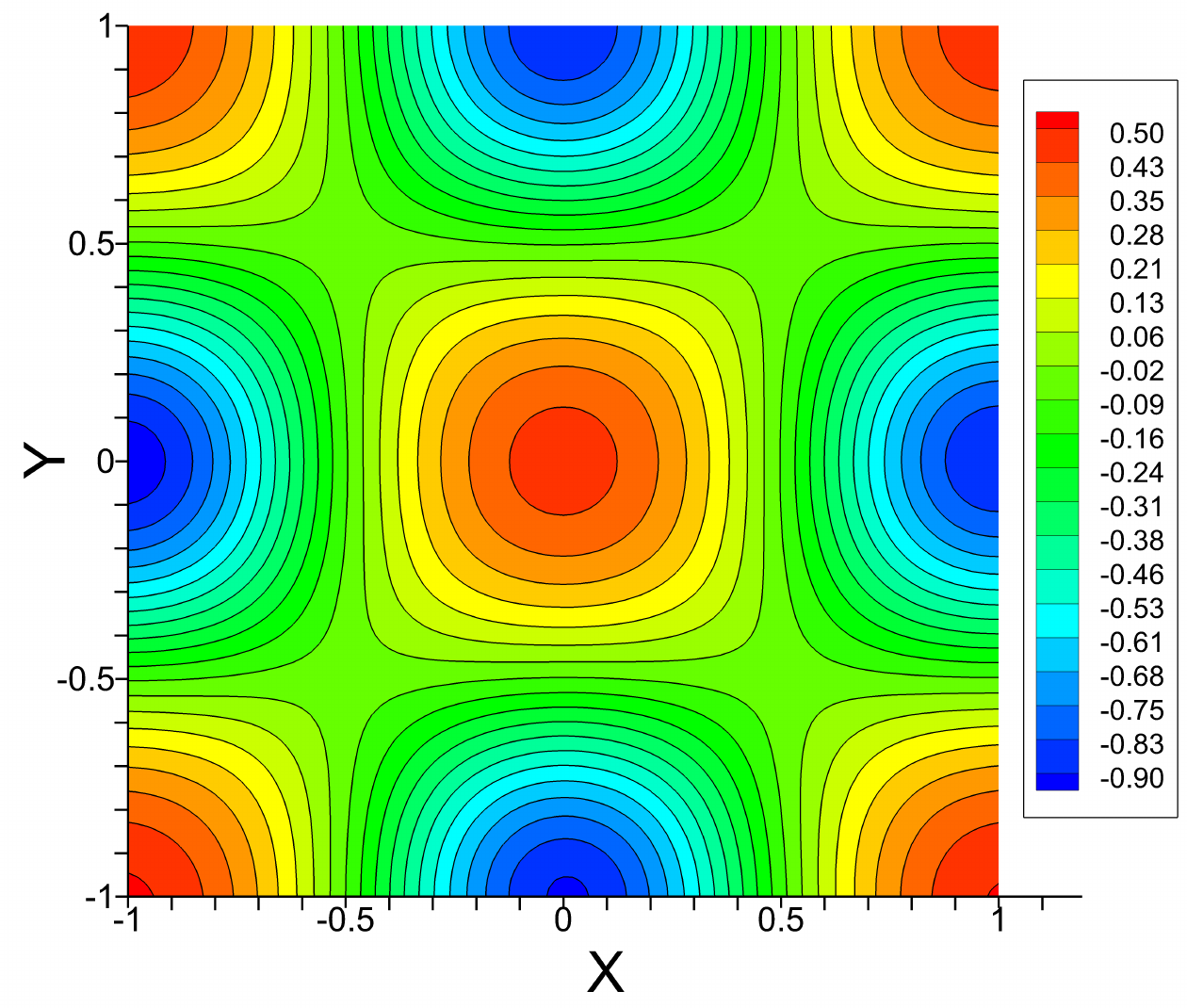}
        \caption{EvoKAN solution}
    \end{subfigure}
    \begin{subfigure}[b]{0.24\linewidth}
        \centering
        \includegraphics[width=\linewidth]{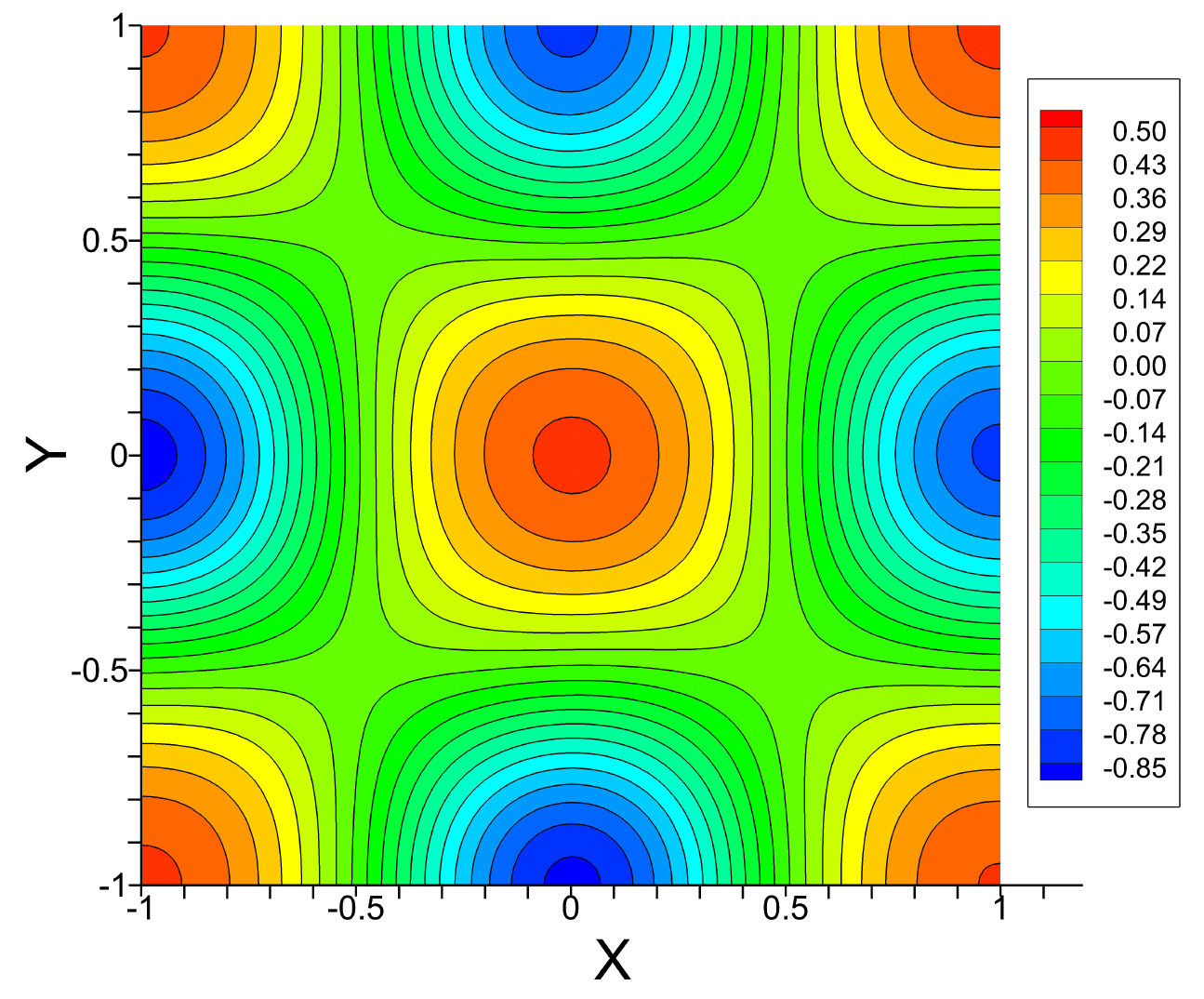}
        \caption{Vanilla PINN solution}
    \end{subfigure} \\
    \begin{subfigure}[b]{0.24\linewidth}
        \centering
        \includegraphics[width=\linewidth]{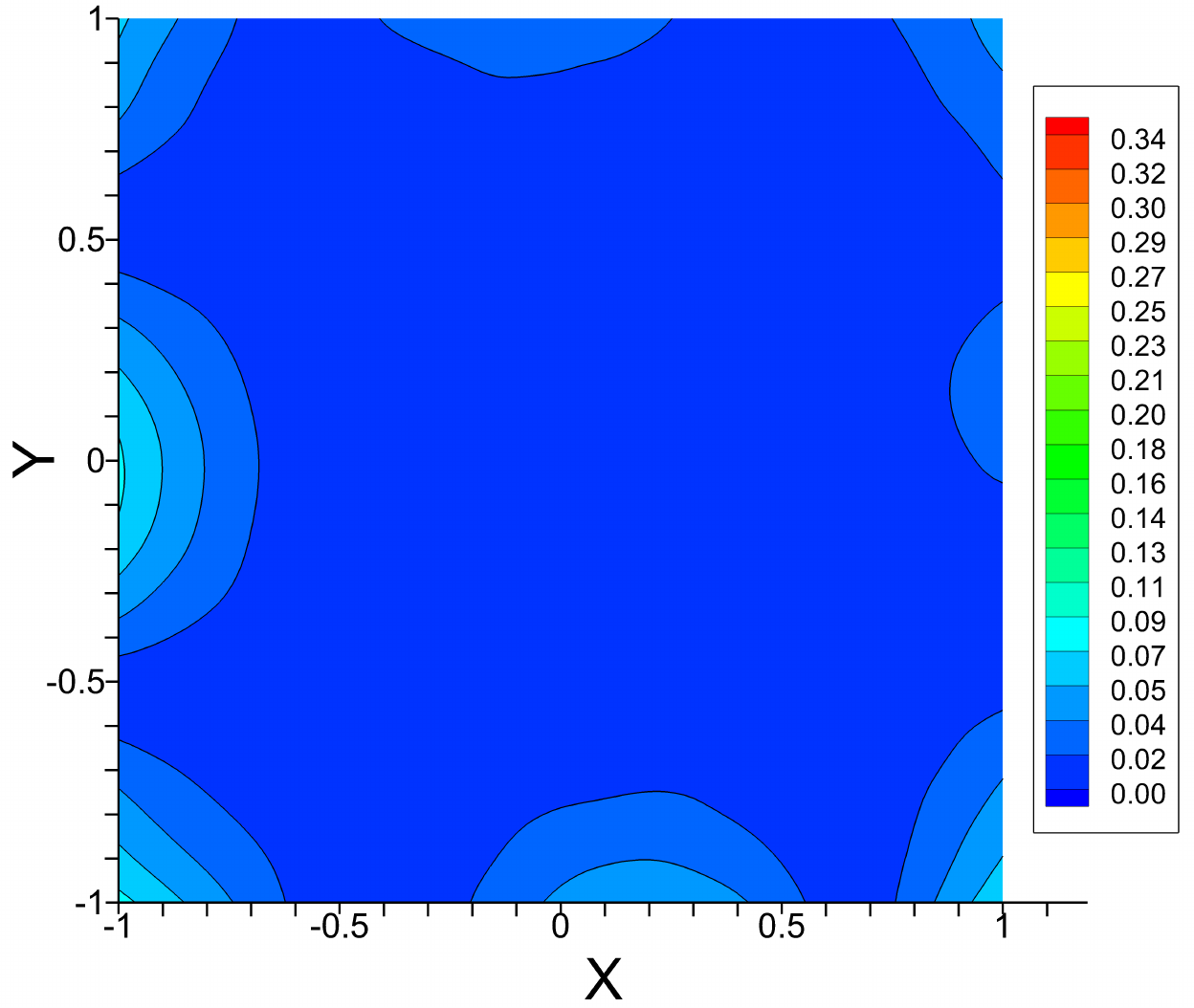}
        \caption{BEKAN absolute error} 
    \end{subfigure} 
    \hfill
    \begin{subfigure}[b]{0.24\linewidth}
        \centering
        \includegraphics[width=\linewidth]{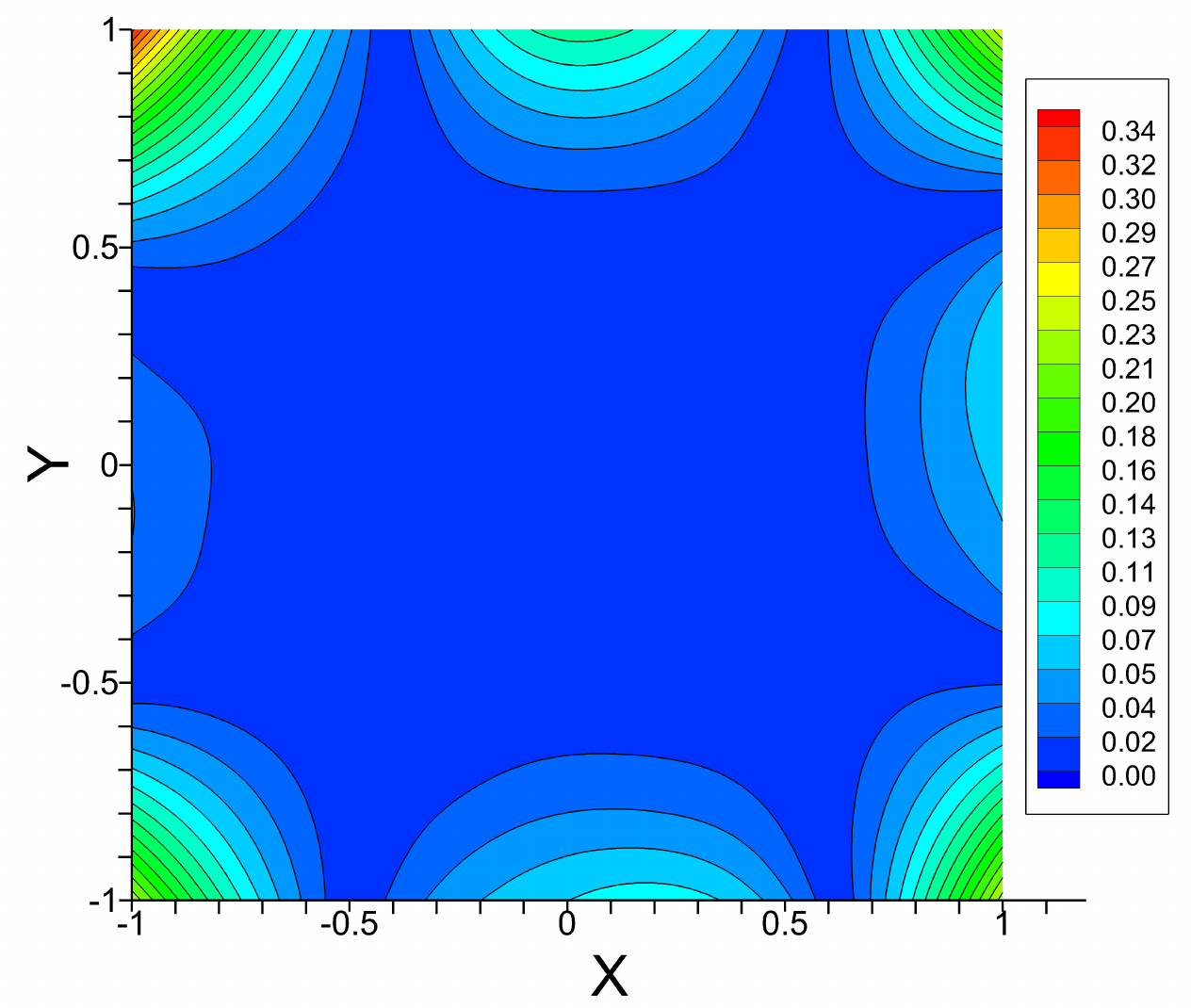}
        \caption{EDNN absolute error} 
    \end{subfigure} 
    \hfill
    \begin{subfigure}[b]{0.24\linewidth}
        \centering
        \includegraphics[width=\linewidth]{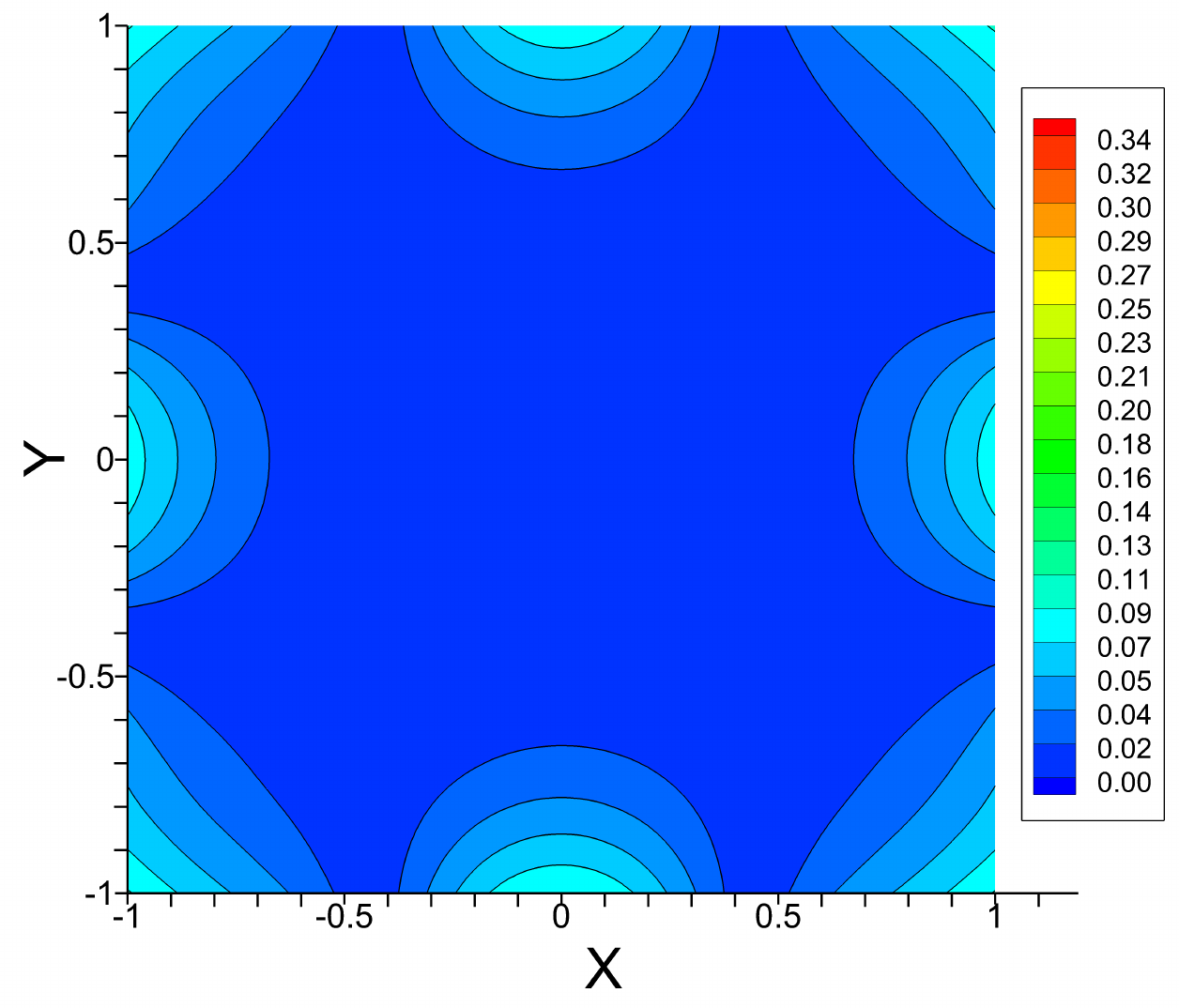}
        \caption{EvoKAN absolute error} 
    \end{subfigure} 
    \hfill
    \begin{subfigure}[b]{0.24\linewidth}
        \centering
        \includegraphics[width=\linewidth]{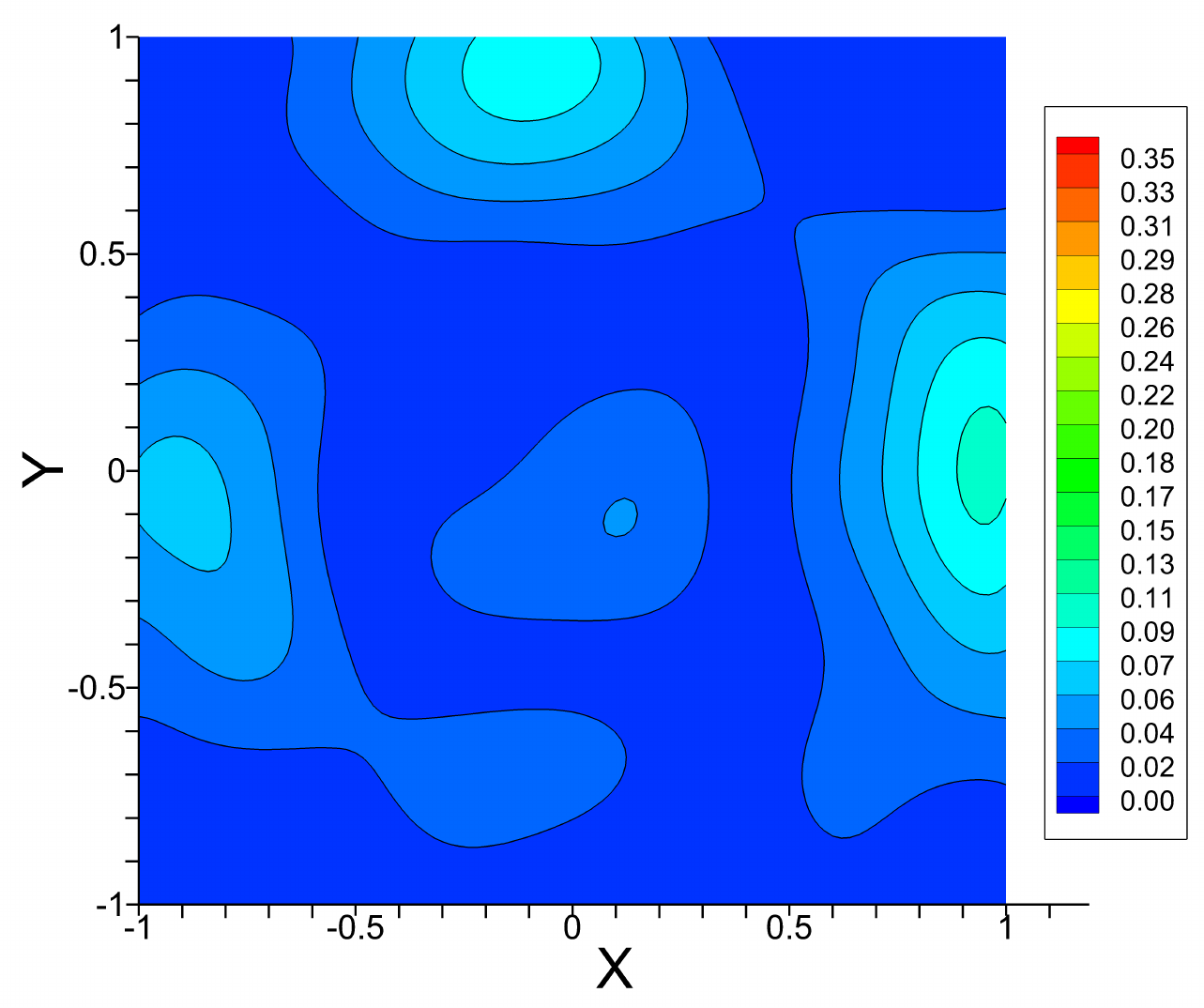}
        \caption{Vanilla PINN absolute error} 
    \end{subfigure}
    \caption{
% 2D heat equation with a nonlinear forcing term (Eq.~\eqref{eq:heat_equation}): Distributions of the solution and absolute error for BEKAN, EDNN, EvoKAN, and vanilla PINN at \( t = \SI{5e-1}{} \).  
% We evaluate the absolute error by comparing the predicted distributions with the FDM reference solution.  
% Compared to more challenging PDEs such as the Allen–Cahn (Eq.~\eqref{eq:Allen-Cahn}), Burgers (Eq.~\eqref{eq:Burgers_equation}), and KS (Eq.~\eqref{eq:KS_equation}) equations, all four models provide reasonably accurate predictions for the heat equation.  
% However, EDNN and EvoKAN produce the largest errors near the four corners of the domain, while vPINN shows noticeable errors along the right and upper boundaries.  
% BEKAN also shows error near the corners, but its magnitude remains relatively small among the models, indicating stable and accurate performance.
BEKAN achieves the highest accuracy for the 2D heat equation with a nonlinear forcing term (Eq.~\eqref{eq:heat_equation}): Distributions of the solution and absolute error for BEKAN, EDNN, EvoKAN, and vanilla PINN at \( t = \SI{5e-1}{} \).  
The absolute error is computed by comparing each prediction with the reference solution from the finite difference method (FDM).  
All models demonstrate reasonably accurate predictions for the heat equation, which is comparatively simpler than the Allen–Cahn (Eq.~\eqref{eq:Allen-Cahn}), Burgers (Eq.~\eqref{eq:Burgers_equation}), and KS (Eq.~\eqref{eq:KS_equation}) equations.  
Among the models, EDNN and EvoKAN show large errors near the domain corners, while vPINN exhibits noticeable errors along the right and upper boundaries.  
BEKAN also shows error near the corners, but its magnitude remains relatively small among the models, indicating stable and accurate performance.
}
    \label{fig:heat_distribution}
\end{figure}

\begin{figure}[htbp!]
    \centering
    \includegraphics[width=0.5\linewidth]{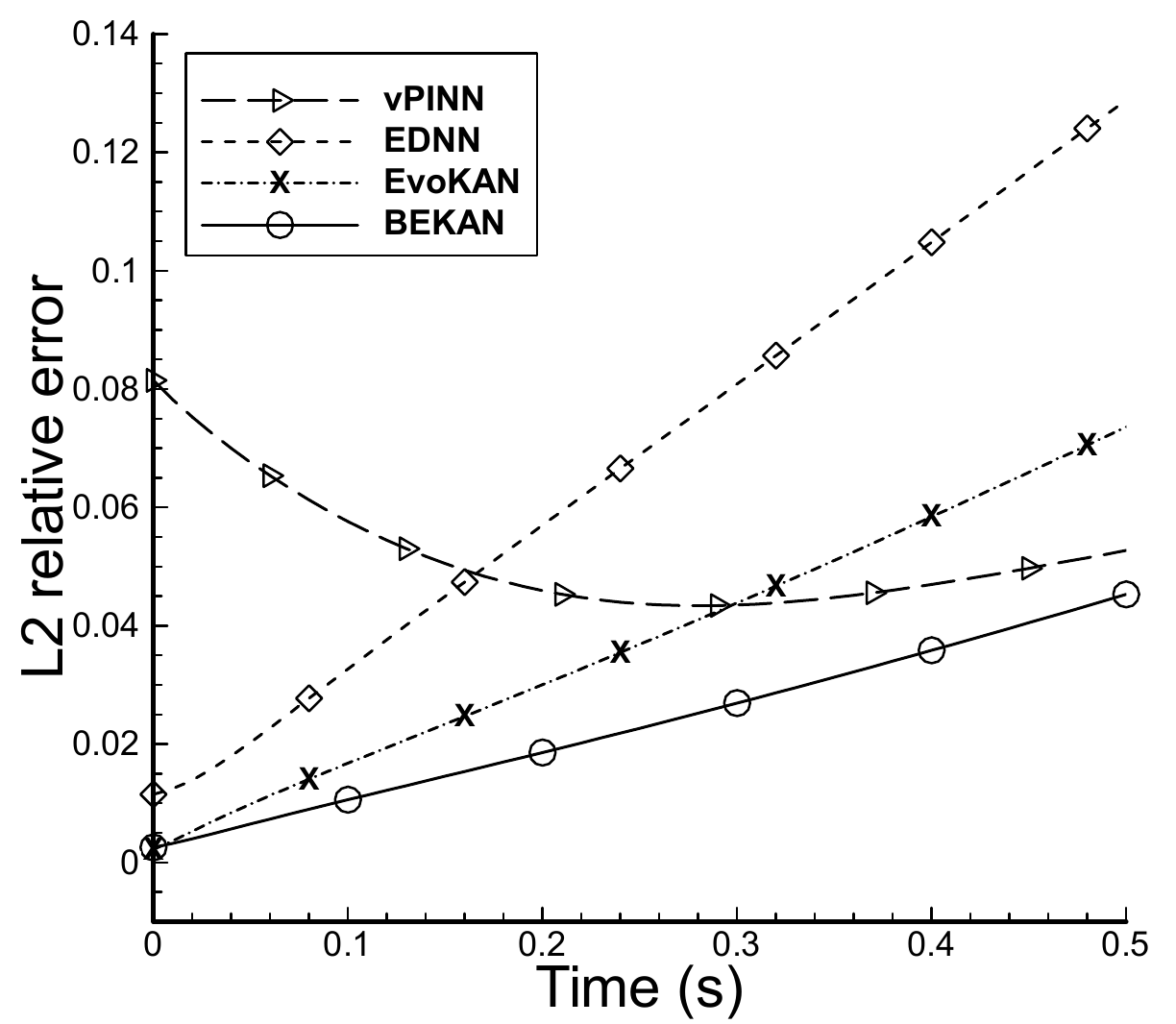}
    \caption{
Time evolution of the \( L_2 \) relative error for three models: BEKAN, EvoKAN, and EDNN, applied to the 2D heat equation with a nonlinear forcing term (Eq.~\eqref{eq:heat_equation}).  
The error is evaluated at each time step with respect to the FDM solution used as the reference.  
All models begin with small \( L_2 \) relative errors at \( t = 0 \), but BEKAN shows a lower rate of error accumulation, maintaining the smallest error throughout the simulation.
}
    \label{fig:heat_L2}
\end{figure}

We now evaluate the accuracy for the 2D Heat equation with a nonlinear source term by visualizing the predicted solutions at the final time step \( t = \SI{5e-1}{} \) and comparing them with the reference FDM solution in Fig.~\ref{fig:heat_distribution}.  
All models produce similar solution patterns, but the absolute error distributions reveal that BEKAN exhibits the lowest absolute error across the domain as shown in Fig.~\ref{fig:heat_distribution}e--h.
Additionally, we track the evolution of the $L_2$ relative error throughout the entire simulation in Fig.~\ref{fig:heat_L2}.
BEKAN, EvoKAN, and EDNN show a gradual increase in error as time progresses, while the vanilla \ac{PINN} initially decreases before increasing again.  
Among all models, BEKAN consistently maintains the lowest \( L_2 \) relative error throughout the simulation.

\begin{figure}[htbp]
    \centering \,
    \begin{subfigure}[b]{0.43\linewidth}
        \centering
        \includegraphics[width=\linewidth]{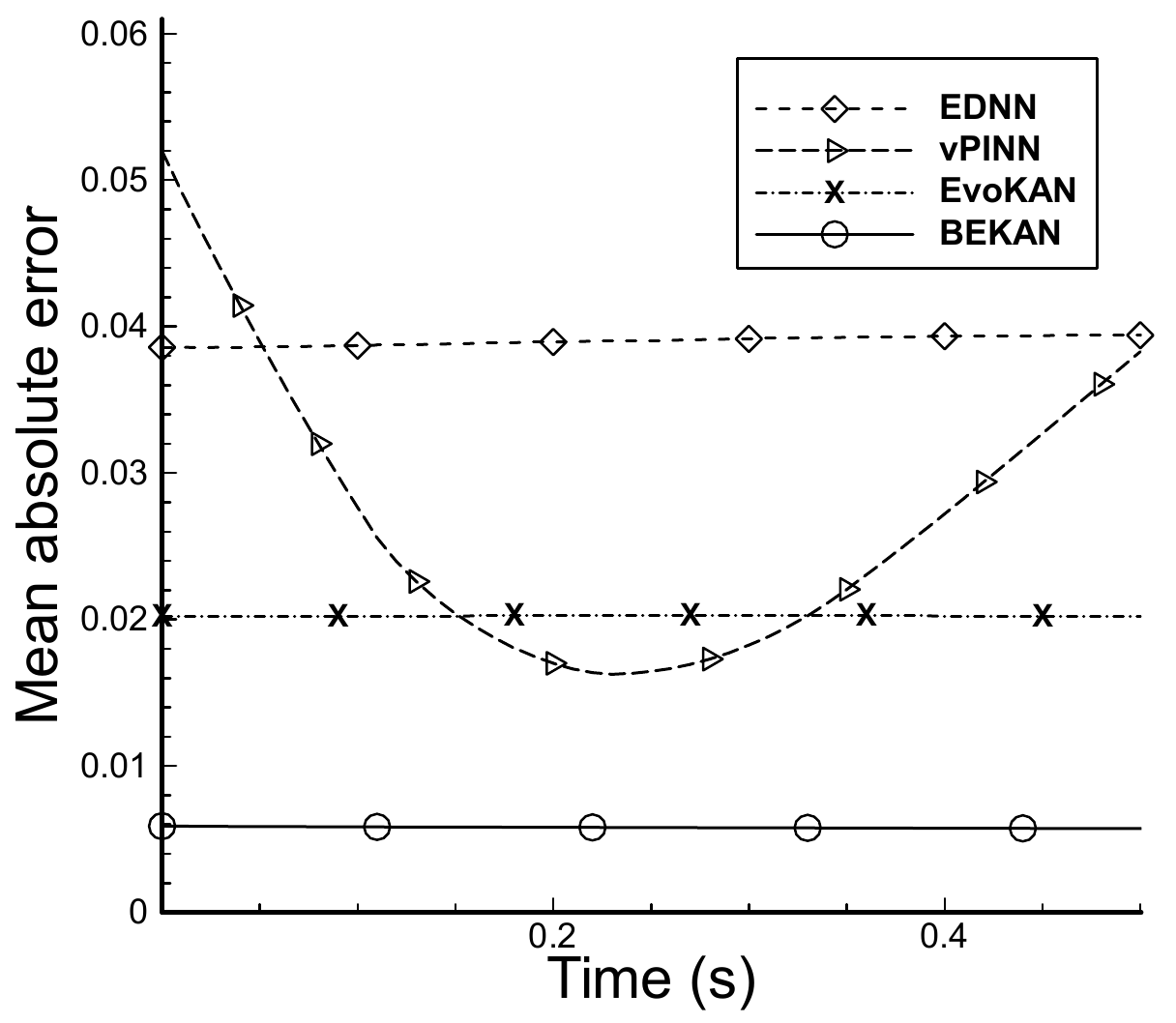}
        \caption{Error of gradient on the left boundary}
    \end{subfigure}
    % \hfill
    \begin{subfigure}[b]{0.43\linewidth}
        \centering
        \includegraphics[width=\linewidth]{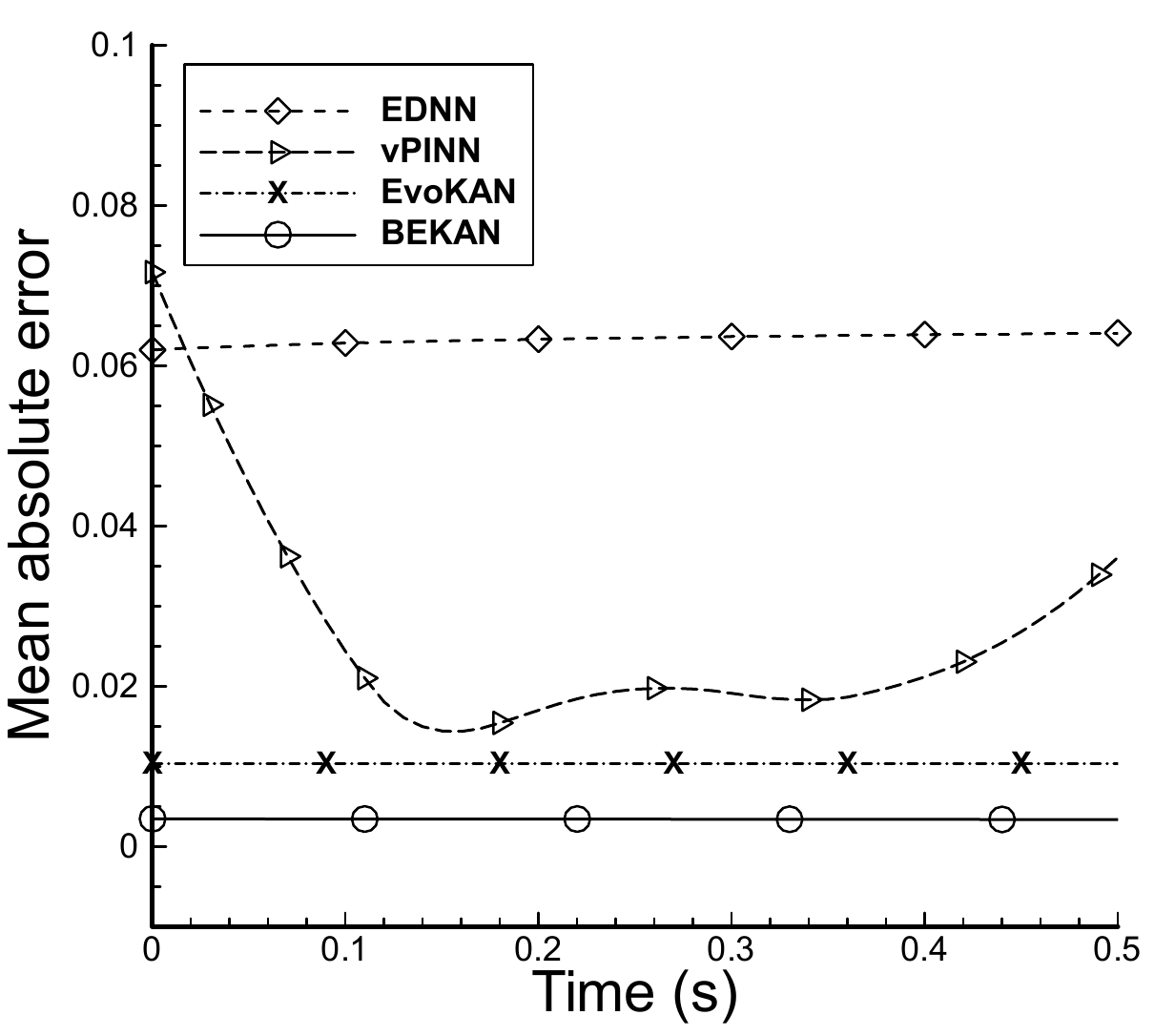}
        \caption{Error of gradient on the right boundary} 
    \end{subfigure} \\
    \begin{subfigure}[b]{0.43\linewidth}
        \centering
        \includegraphics[width=\linewidth]{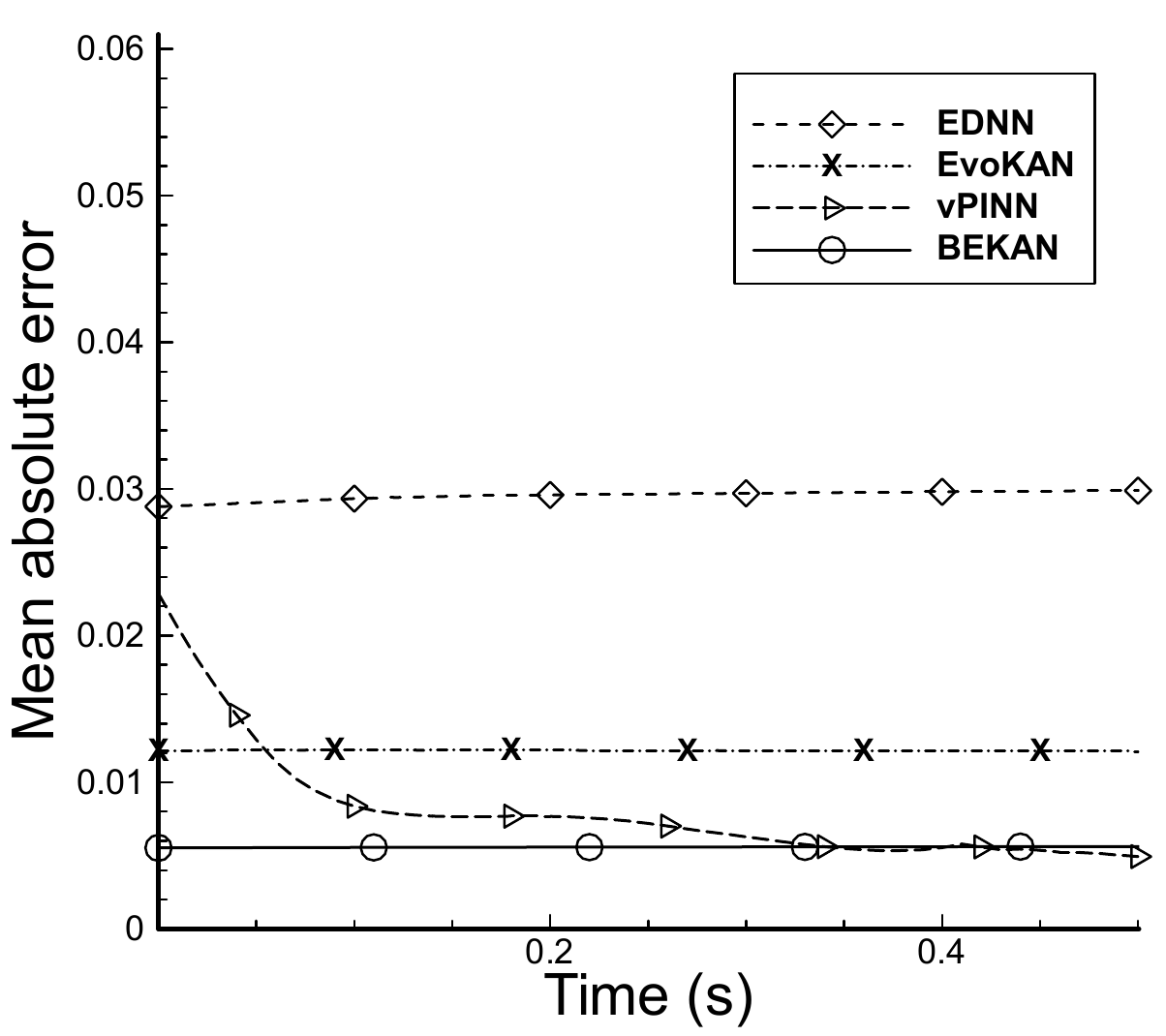}
        \caption{Error of gradient on the lower boundary}
    \end{subfigure}
    % \hfill
    \begin{subfigure}[b]{0.43\linewidth}
        \centering
        \includegraphics[width=\linewidth]{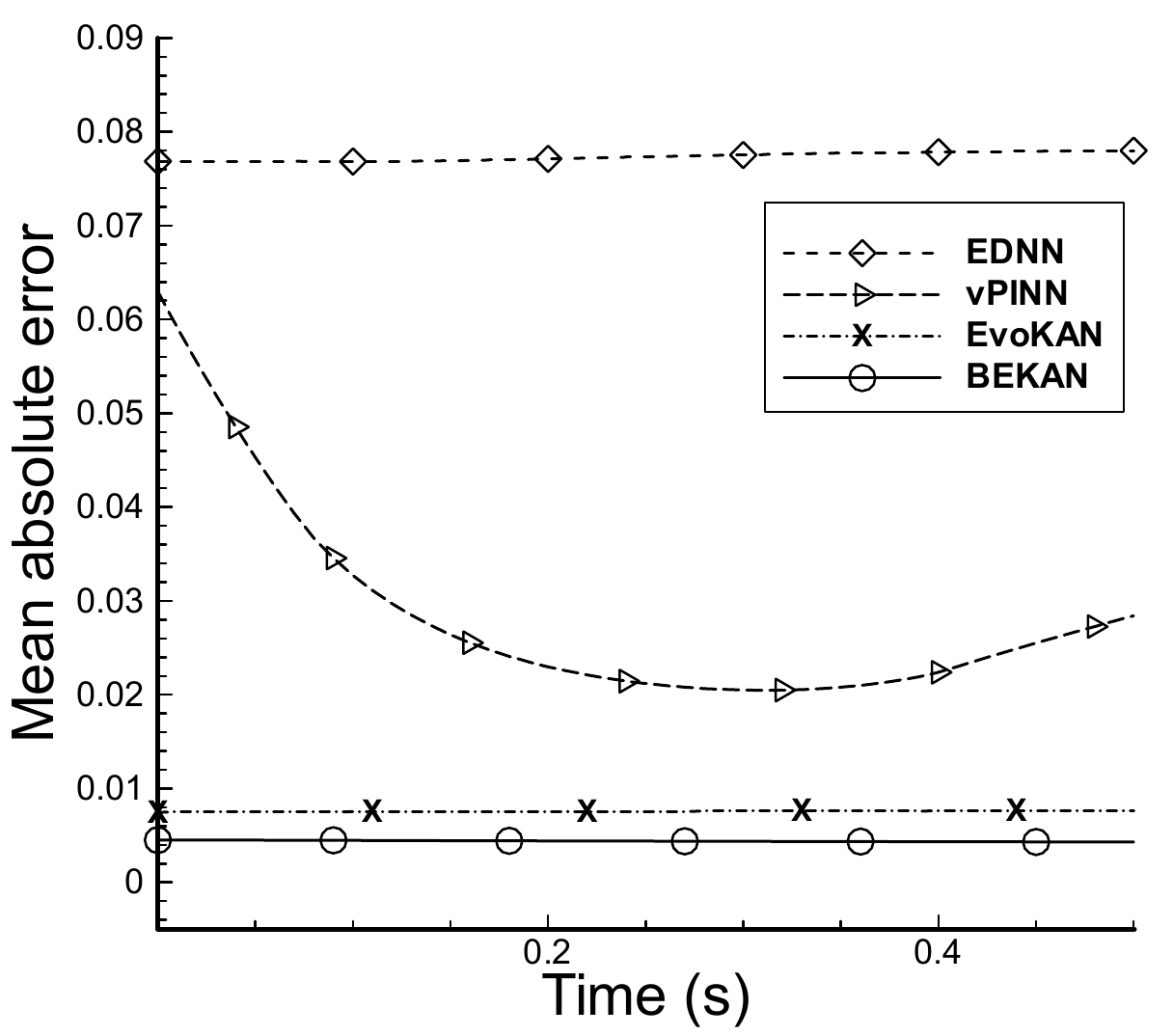}
        \caption{Error of gradient on the upper boundary} 
    \end{subfigure}
    \caption{
2D heat equation with a nonlinear forcing term (Eq.~\eqref{eq:heat_equation}): Mean absolute error of the gradient on each boundary for BEKAN, EvoKAN, EDNN, and vanilla PINN, compared with the FDM solution, to assess compliance with the Neumann boundary conditions specified in Eq.~\eqref{eq:heat_equation_BC}.  
BEKAN, EDNN, and EvoKAN, which incorporate the Neumann boundary conditions through evolutionary approaches, maintain relatively stable boundary errors over time, whereas vanilla PINN exhibits noticeable fluctuations during the simulation.
}
    \label{fig:Error of gradient on the boundary}
\end{figure}

We assess the Neumann boundary condition accuracy across all time steps by tracking the mean absolute gradient error along the four boundaries.
Figure~\ref{fig:Error of gradient on the boundary} shows that BEKAN, EDNN, and EvoKAN, which use the proposed approach described in Sec~\ref{subsec:Neumann}, maintain steady error levels over time.
In Fig.~\ref{fig:Error of gradient on the boundary}, the vanilla PINN shows variations in boundary error as time progresses.
Overall, BEKAN achieves the smallest error throughout the entire simulation.

\subsection{Heat Equation with Mixed Boundary Condition}
\label{sec:experiment_mixed}

In this experiment, we explore the capability of BEKAN to handle mixed boundary conditions that include both Dirichlet and Neumann types.  
% The classical two-dimensional heat equation describes the diffusion of thermal energy in a medium under purely conductive effects.  
% However, in more realistic situations where external influences or internal reactions are present, it is necessary to incorporate a nonlinear source term that accounts for these effects.  
We examine a generalized version of the heat equation augmented with a nonlinear source term:
\begin{equation}
\label{eq:heat_equation_mixed}
    \frac{\partial u}{\partial t} = \alpha \left( \frac{\partial^2 u}{\partial x^2} + \frac{\partial^2 u}{\partial y^2} \right) + u(1 - u),
\end{equation}
where \(\alpha = 1\). 
The problem is initialized with:
\begin{equation}
    u(x, y, 0) = \sin \left( \frac{\pi}{2}x \right) \sin \left( \frac{\pi}{2}y \right),
\end{equation}
together with the following mixed boundary conditions:
\begin{equation}
\label{eq:heat_mixed_BC}
    u(0, y, t) = 0, \quad \frac{\partial u}{\partial x}(1, y, t) = 0, \quad u(x, 0, t) = 0, \quad \frac{\partial u}{\partial y}(x, 1, t) = 0.
\end{equation}
We define the total energy of the system by:
\begin{equation}
    E[u] = \iint_{\Omega} \frac{1}{2} \left( u_x^2 + u_y^2 \right) \, dx\,dy,
\end{equation}
which provides a quantitative measure of the spatial gradient magnitude over the domain \(\Omega\).

\sisetup{group-separator={,}, group-minimum-digits=4}
\begin{table} [hbtp!]
\footnotesize
	\renewcommand{\arraystretch}{1.0}
	\begin{center} 
		\caption{Training configuration for the 2D heat equation with mixed boundary condition (Eq.~\eqref{eq:heat_equation}).
}
		\begin{tabular}{l c c c c}
			\hline
			{\, \, \, } & \makecell[c]{BEKAN} & \makecell[c]{EvoKAN} & \makecell[c]{EDNN} & {Vanilla PINN} \\
			\hline
			{Hidden layers} & {[4, 4, 4, 4]} & \makecell[c]{[4, 4, 4, 4]} & {[15, 15, 15]} & {[15, 15, 15]}\\
            {Activation functions} & \makecell[c]{Gaussian RBFs/SiLU} & \makecell[c]{B-splines/SiLU} & {tanh} & {tanh} \\
            % \hline
            \makecell[l]{Grid points number\\ of activation functions} & {5} & \makecell[c]{5} & {-} & {-}\\
			\makecell[l]{Number of \\ trainable parameters} & {\SI{352}{}} & {\SI{600}{}} & {\SI{541}{}} & \makecell[c]{\SI{541}{}}\\
            {Optimizer} & \makecell[c]{Adam} & \makecell[c]{Adam} & \makecell[c]{Adam} & \makecell[c]{Adam/L-BFGS-B}\\
            {Timestep} & \makecell[c]{5e-05} & \makecell[c]{5e-05} & \makecell[c]{5e-05} & \makecell[c]{-}\\
			\hline
		\end{tabular}
		\label{table:heat_mixed_training}
	\end{center}
\end{table}

Table~\ref{table:heat_mixed_training} outlines the configuration employed for model training.  
While BEKAN and \ac{EvoKAN} share an identical hidden layer design, they differ in their choice of basis functions. 
\ac{EvoKAN} incorporates B-spline basis functions, which require additional scaling parameters, leading to a total of 600 trainable weights.  
In comparison, BEKAN involves 352 trainable parameters.  
The evolutionary models are trained progressively in discrete time steps of \( t = \SI{5e-5}{} \),  
whereas the vanilla \ac{PINN} is optimized across the full temporal domain in a single training cycle.

\begin{figure}[htbp]
    \centering \,
    \begin{subfigure}[b]{0.24\linewidth}
        \centering
        \includegraphics[width=\linewidth]{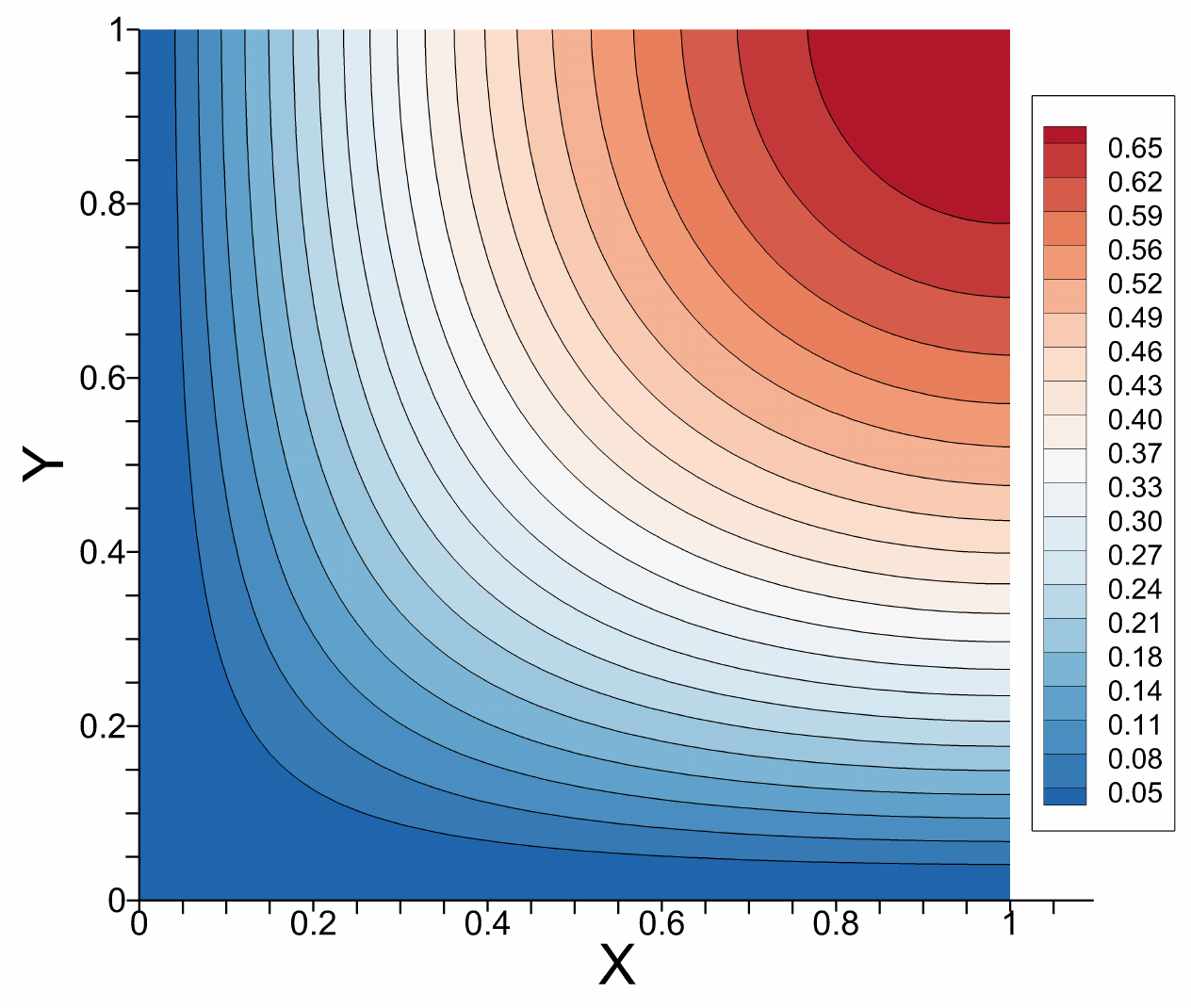}
        \caption{BEKAN solution}
    \end{subfigure}
    \hfill
    \begin{subfigure}[b]{0.24\linewidth}
        \centering
        \includegraphics[width=\linewidth]{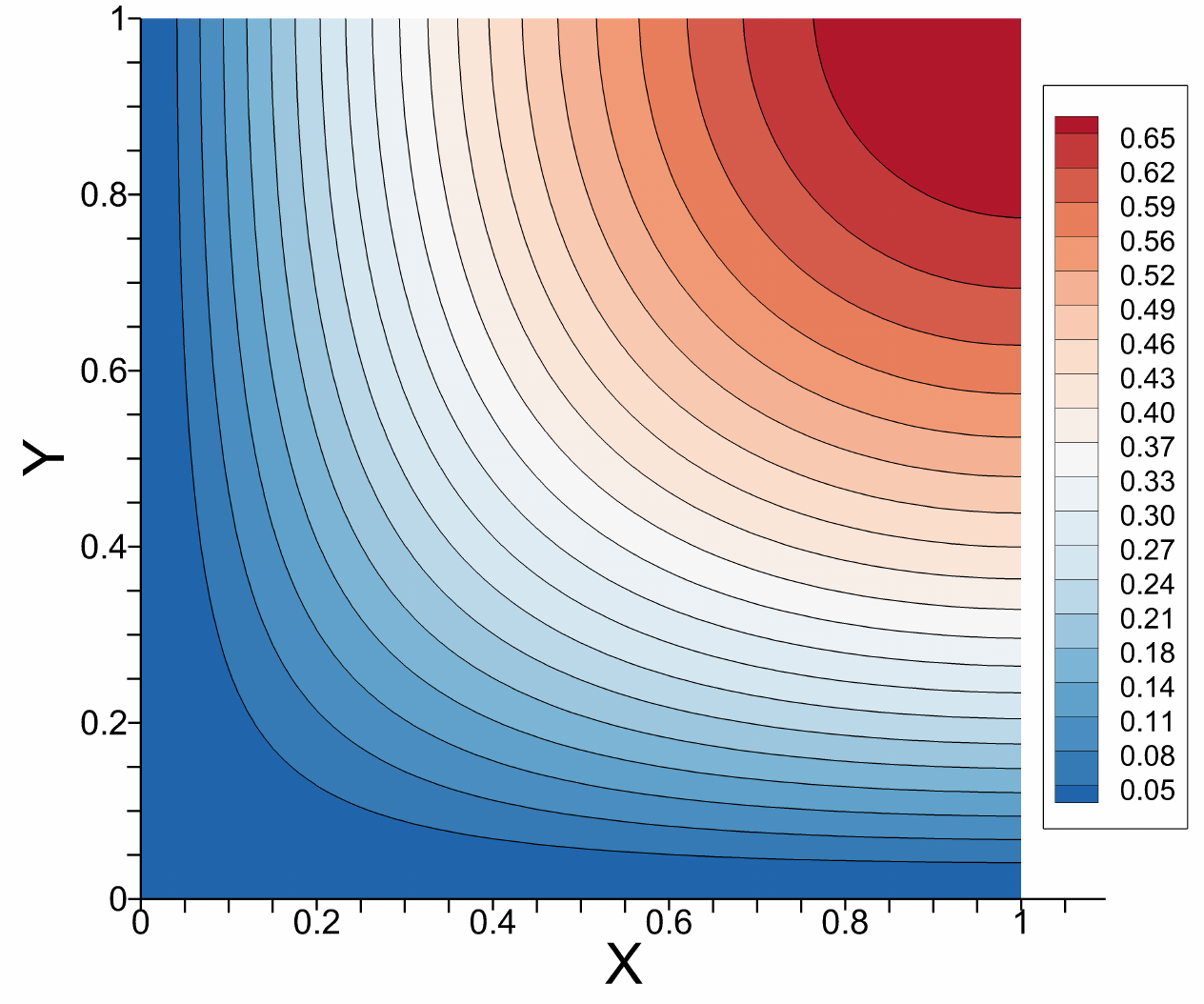}
        \caption{EDNN solution}
    \end{subfigure}
    \begin{subfigure}[b]{0.24\linewidth}
        \centering
        \includegraphics[width=\linewidth]{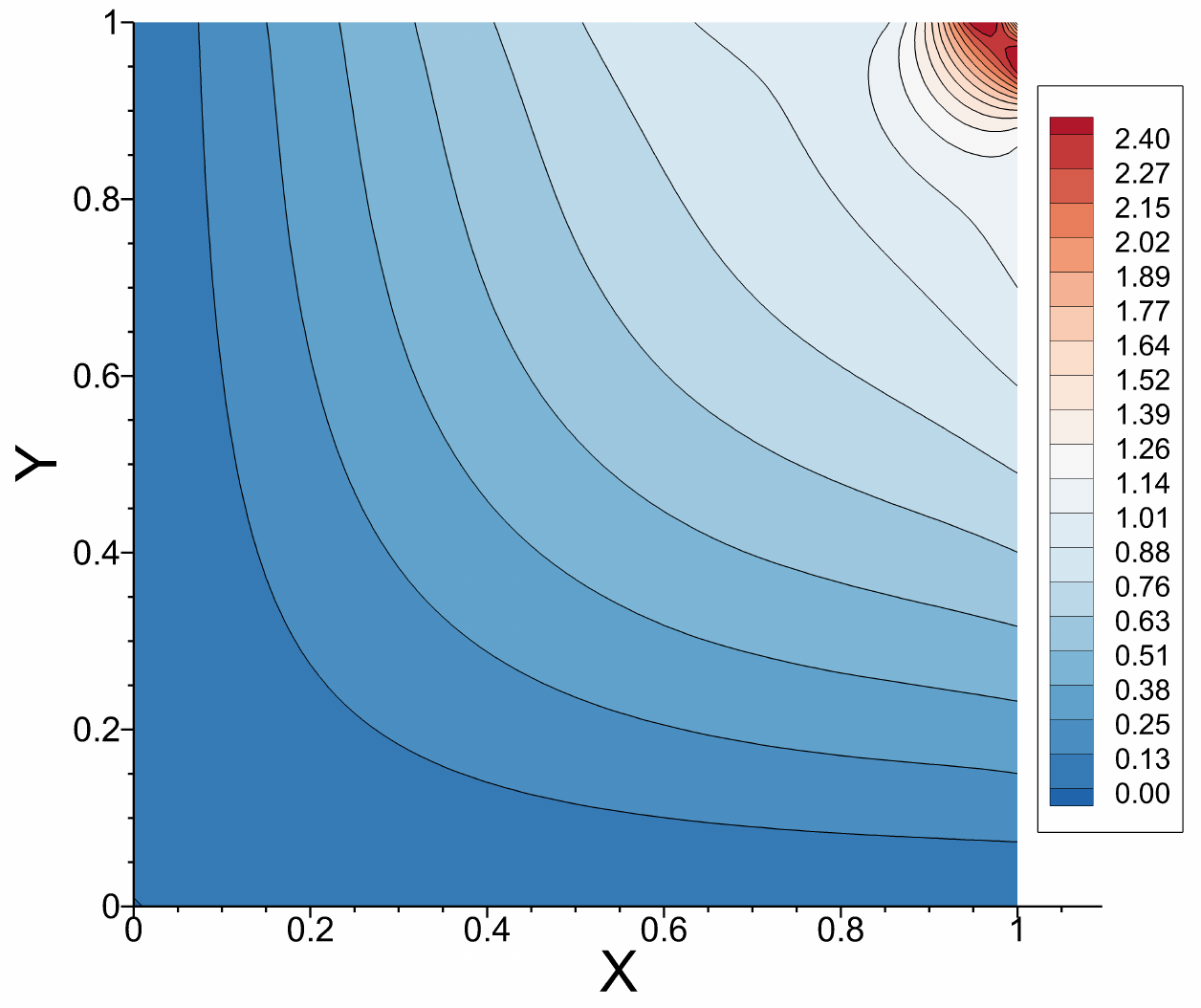}
        \caption{EvoKAN solution}
    \end{subfigure}
    \begin{subfigure}[b]{0.24\linewidth}
        \centering
        \includegraphics[width=\linewidth]{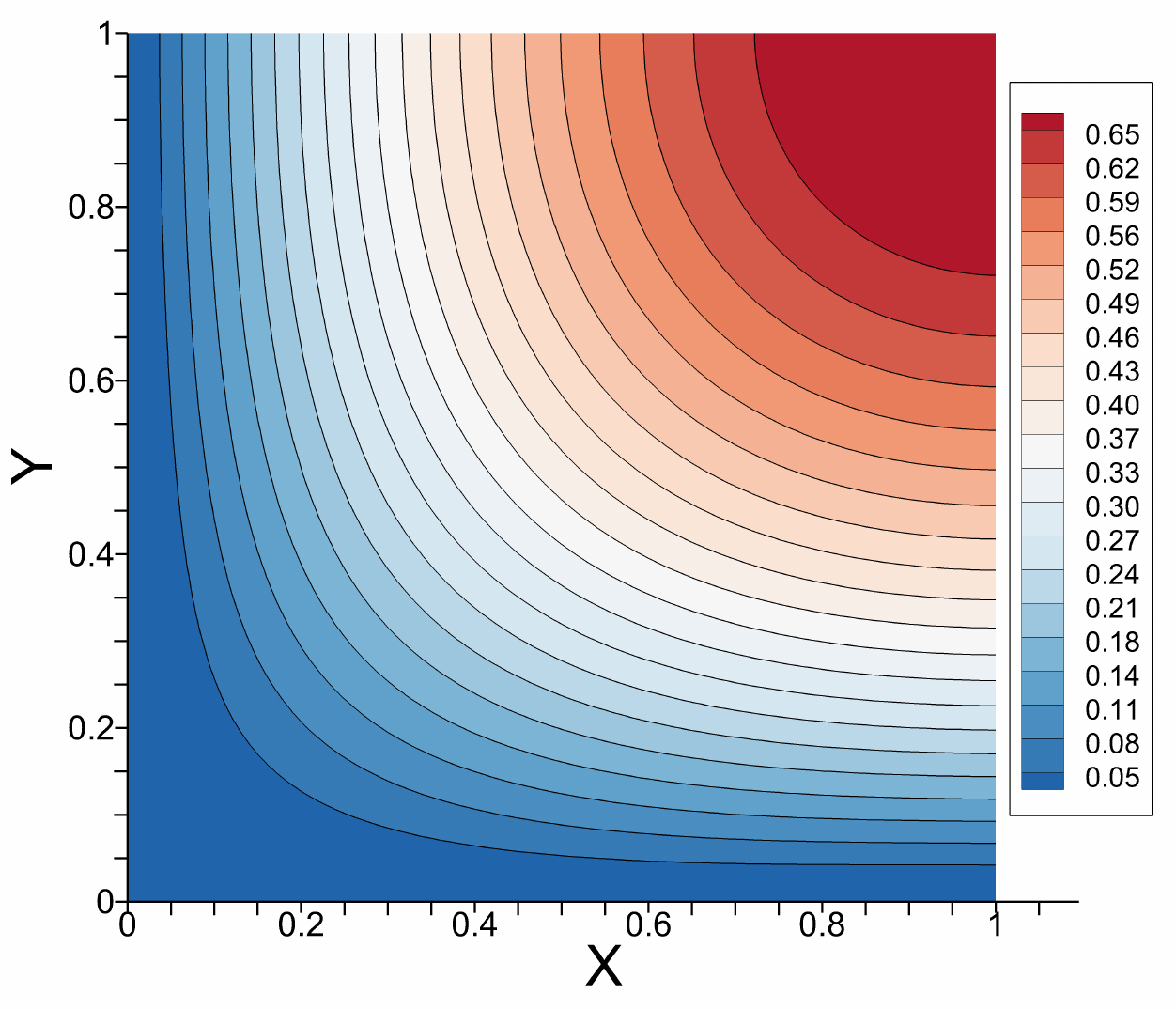}
        \caption{vPINN solution}
    \end{subfigure} \\
    \begin{subfigure}[b]{0.24\linewidth}
        \centering
        \includegraphics[width=\linewidth]{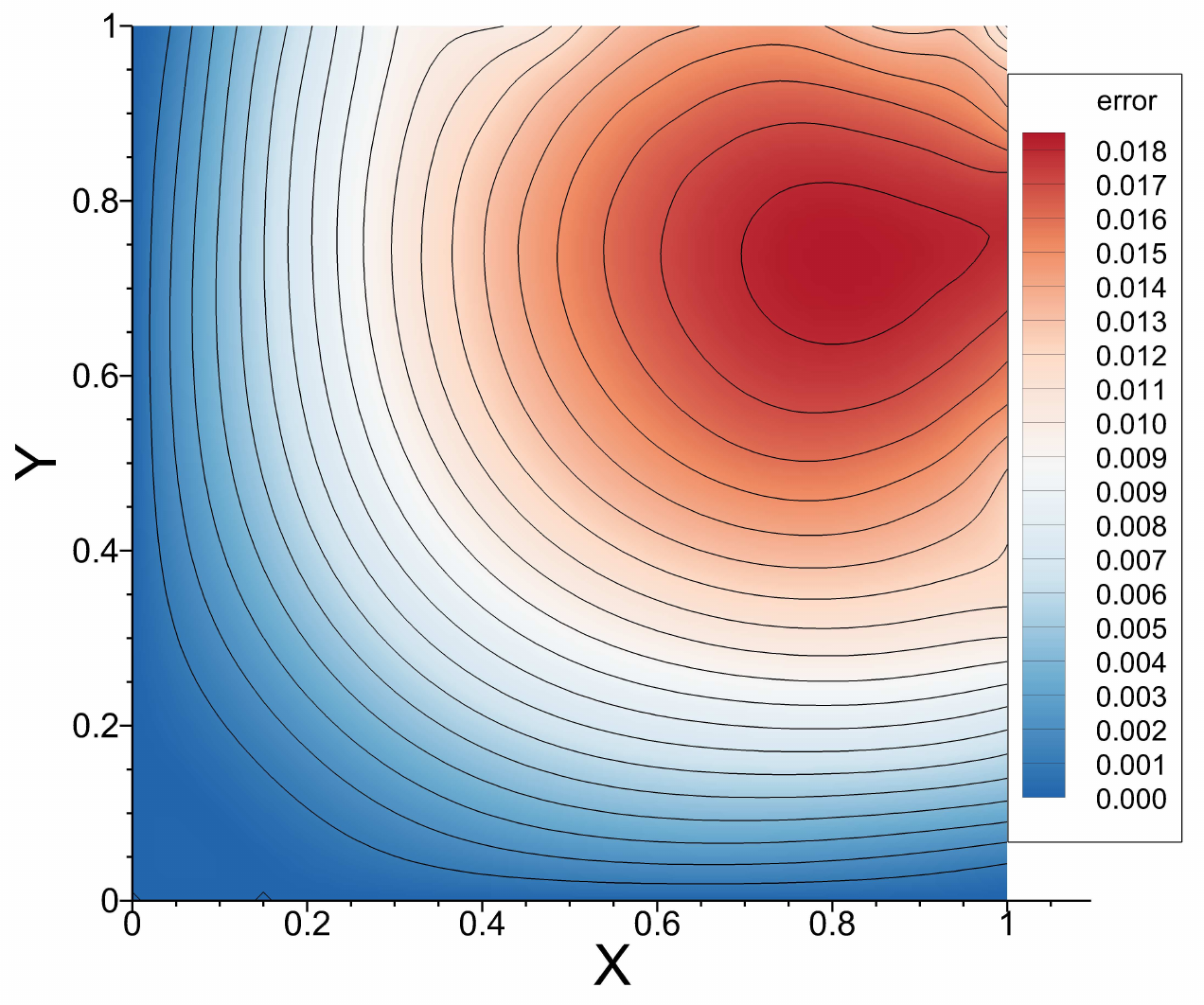}
        \caption{BEKAN absolute error} 
    \end{subfigure} 
    \begin{subfigure}[b]{0.24\linewidth}
        \centering
        \includegraphics[width=\linewidth]{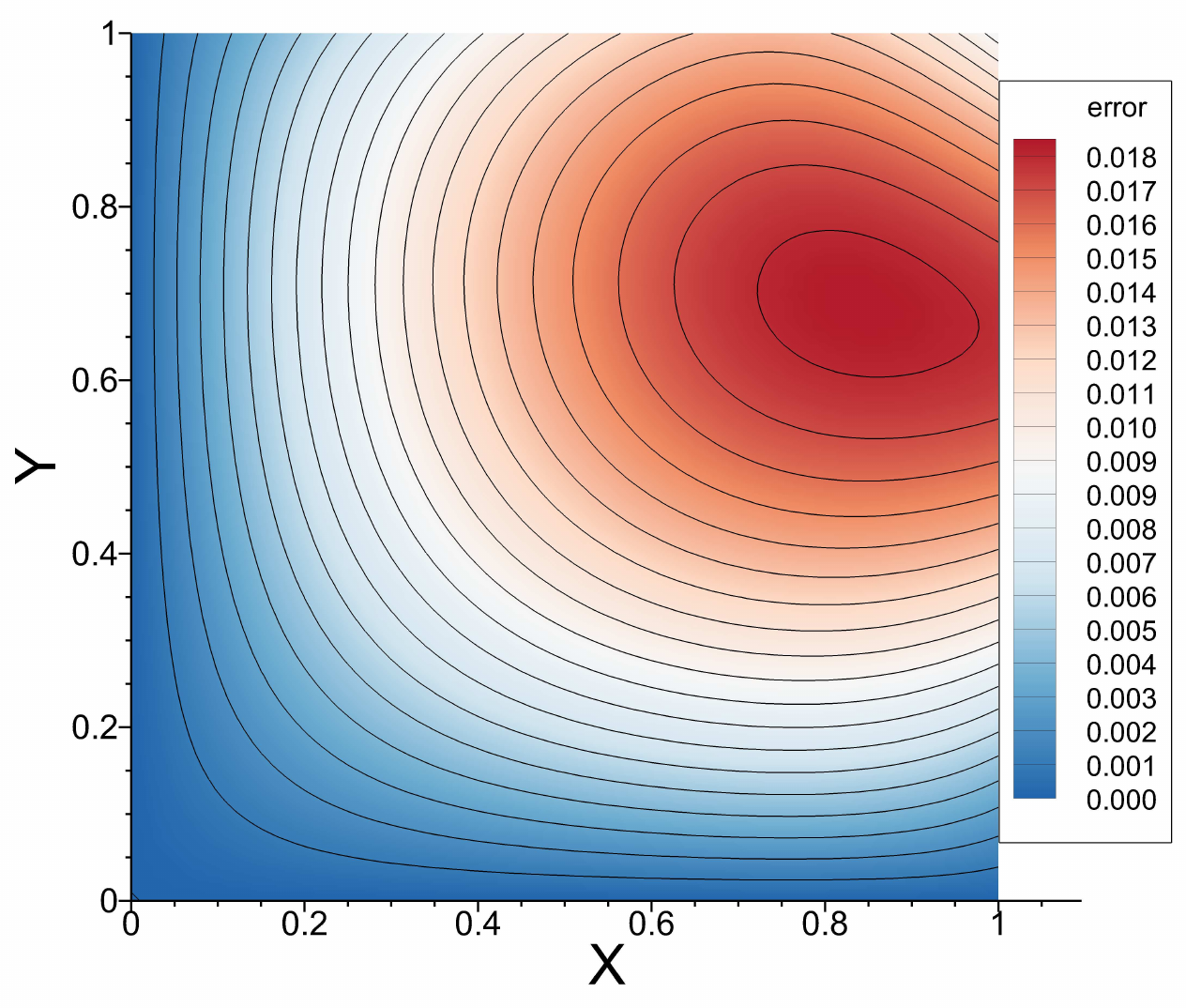}
        \caption{EDNN absolute error} 
    \end{subfigure} 
    % \hfill
    \hfill
    \begin{subfigure}[b]{0.24\linewidth}
        \centering
        \includegraphics[width=\linewidth]{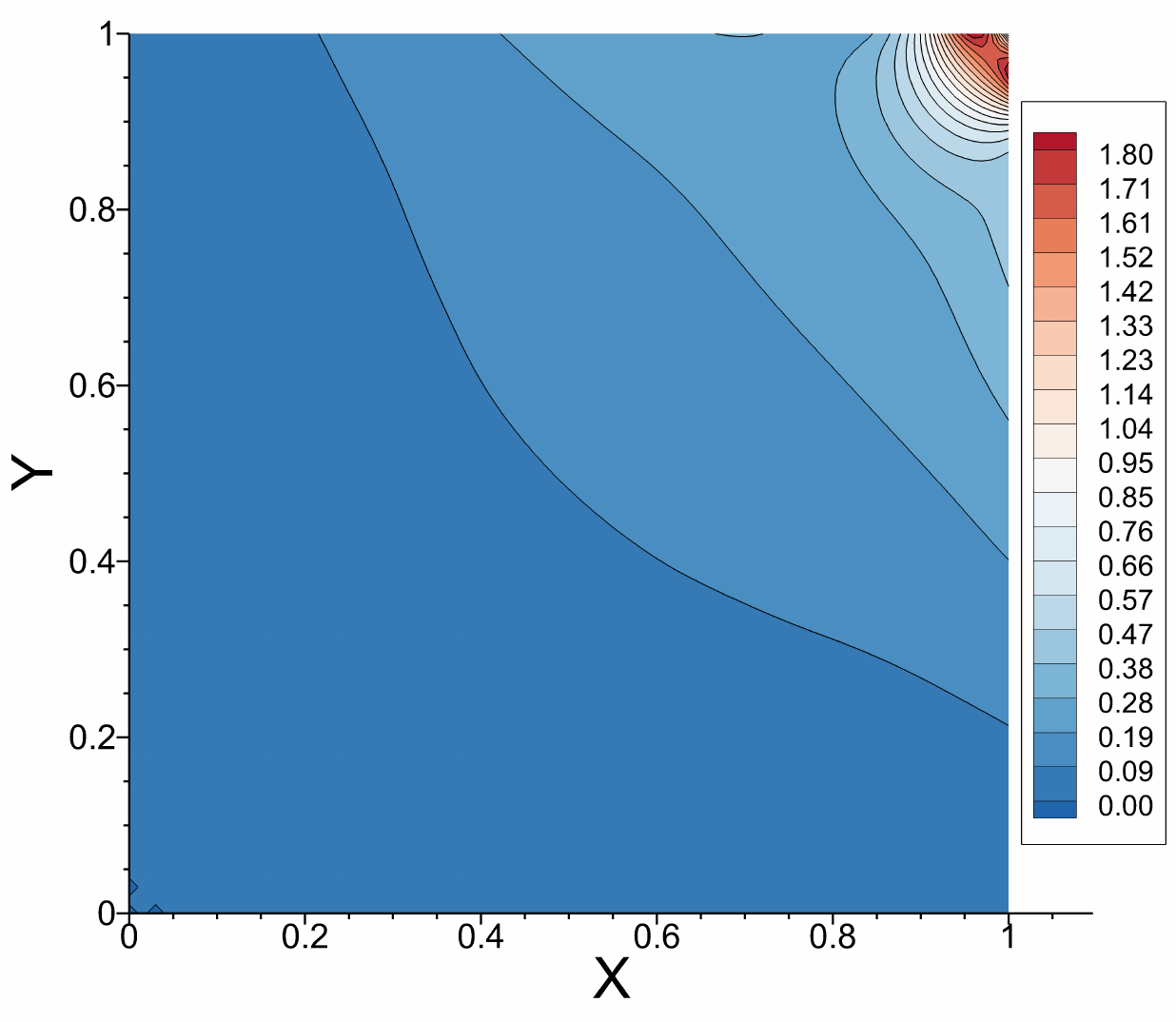}
        \caption{EvoKAN absolute error} 
    \end{subfigure} 
    \begin{subfigure}[b]{0.24\linewidth}
        \centering
        \includegraphics[width=\linewidth]{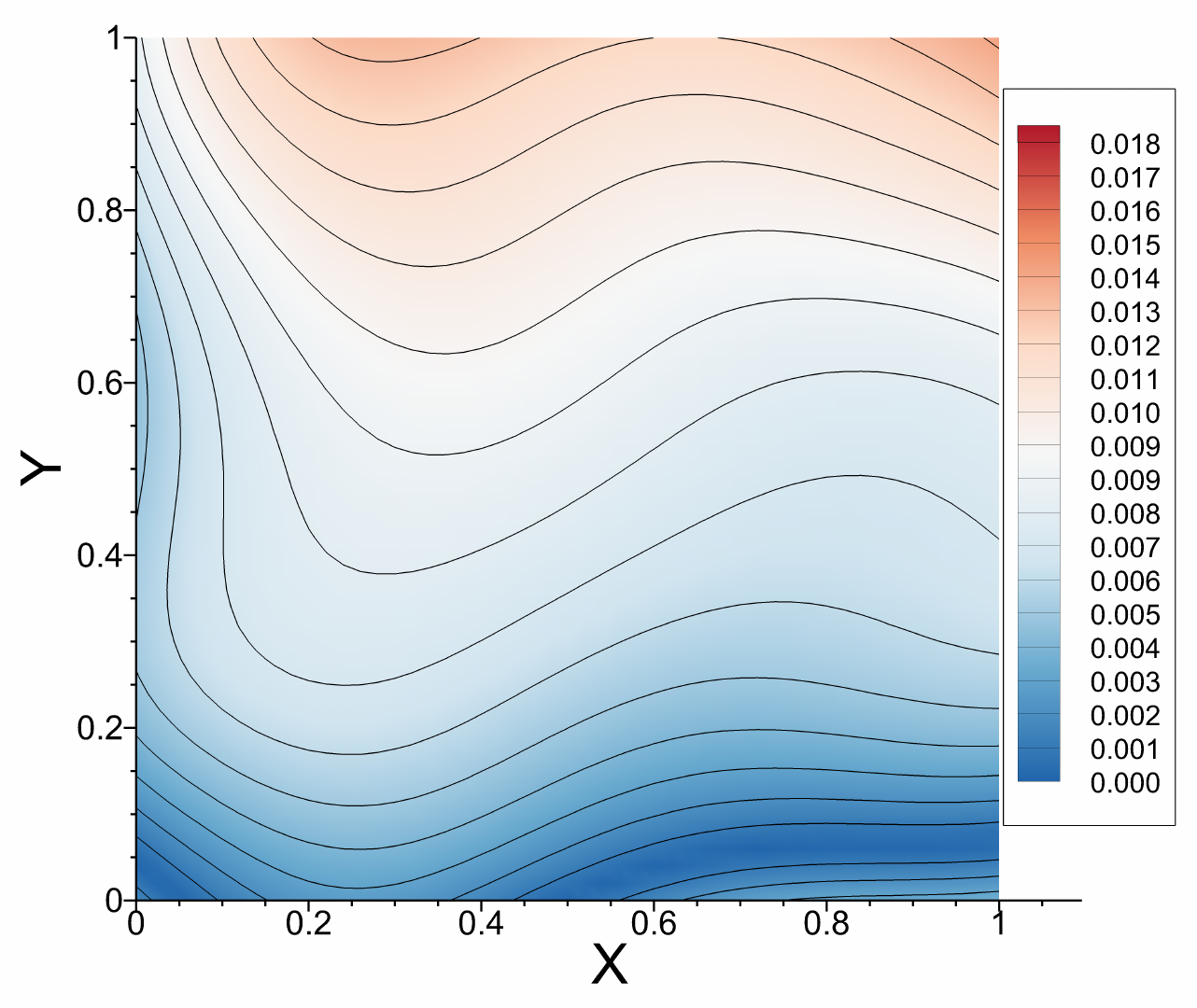}
        \caption{vPINN absolute error} 
    \end{subfigure} 
    \caption{
    % 2D heat equation with mixed boundary conditions (Eq.~\eqref{eq:heat_equation_mixed}): We present the distributions of the solution and absolute error for BEKAN, EDNN, EvoKAN, and vanilla PINN at \( t = 5 \times 10^{-1} \).  
% We compute the absolute error by comparing the predicted solutions with the FDM reference.  
% Compared to more challenging PDEs such as the Burgers (Eq.~\eqref{eq:Burgers_equation}), Allen–Cahn (Eq.~\eqref{eq:Allen-Cahn}), or KS equations (Eq.~\eqref{eq:KS_equation}), the heat equation allows EDNN and BEKAN to achieve similar performance.  
% EvoKAN shows difficulty in generating accurate predictions when both Neumann and Dirichlet conditions are imposed, and although vPINN maintains reasonable overall accuracy, it produces errors along the left and lower boundaries where Dirichlet conditions apply.
BEKAN and EDNN show reasonable performance on the 2D heat equation with mixed boundary conditions (Eq.~\eqref{eq:heat_equation_mixed}), while EvoKAN has difficulty under the combined constraints.  
Distributions of the solution and absolute error for BEKAN, EDNN, EvoKAN, and vanilla PINN are presented at \( t = 5 \times 10^{-1} \).  
The absolute error is computed with respect to the reference solution obtained from the FDM.  
EDNN and BEKAN yield similar levels of accuracy, likely due to the relatively simple structure of the heat equation compared to the Burgers (Eq.~\eqref{eq:Burgers_equation}), Allen–Cahn (Eq.~\eqref{eq:Allen-Cahn}), and KS (Eq.~\eqref{eq:KS_equation}) equations.  
EvoKAN shows limited accuracy when both Neumann and Dirichlet conditions are imposed, and vPINN produces errors near the left and lower boundaries where Dirichlet conditions apply.
    }
    \label{fig:heat_mixed_contour}
\end{figure}

We assess the accuracy of the models for the 2D Heat equation with mixed boundary conditions by visualizing the predicted solutions at the final time step \( t = \SI{5e-1}{} \), as compared to the reference FDM solution in Fig.~\ref{fig:heat_mixed_contour}.  
While BEKAN, EDNN, and vanilla PINN produce broadly similar solution profiles, EvoKAN fails to capture the solution behavior.  
Unlike the previous case in Sec.~\ref{subsec:heat} where only Neumann boundary conditions were applied, this example includes hard constraints on the output to enforce Dirichlet boundaries, which EvoKAN is unable to handle effectively.  
From the absolute error distributions, we observe that BEKAN and EDNN yield the most accurate results.  
In contrast, vanilla PINN shows visible errors near the domain boundaries, indicating that it does not strictly satisfy the imposed boundary conditions.

\begin{figure}[htbp]
    \centering
    \includegraphics[width=0.5\linewidth]{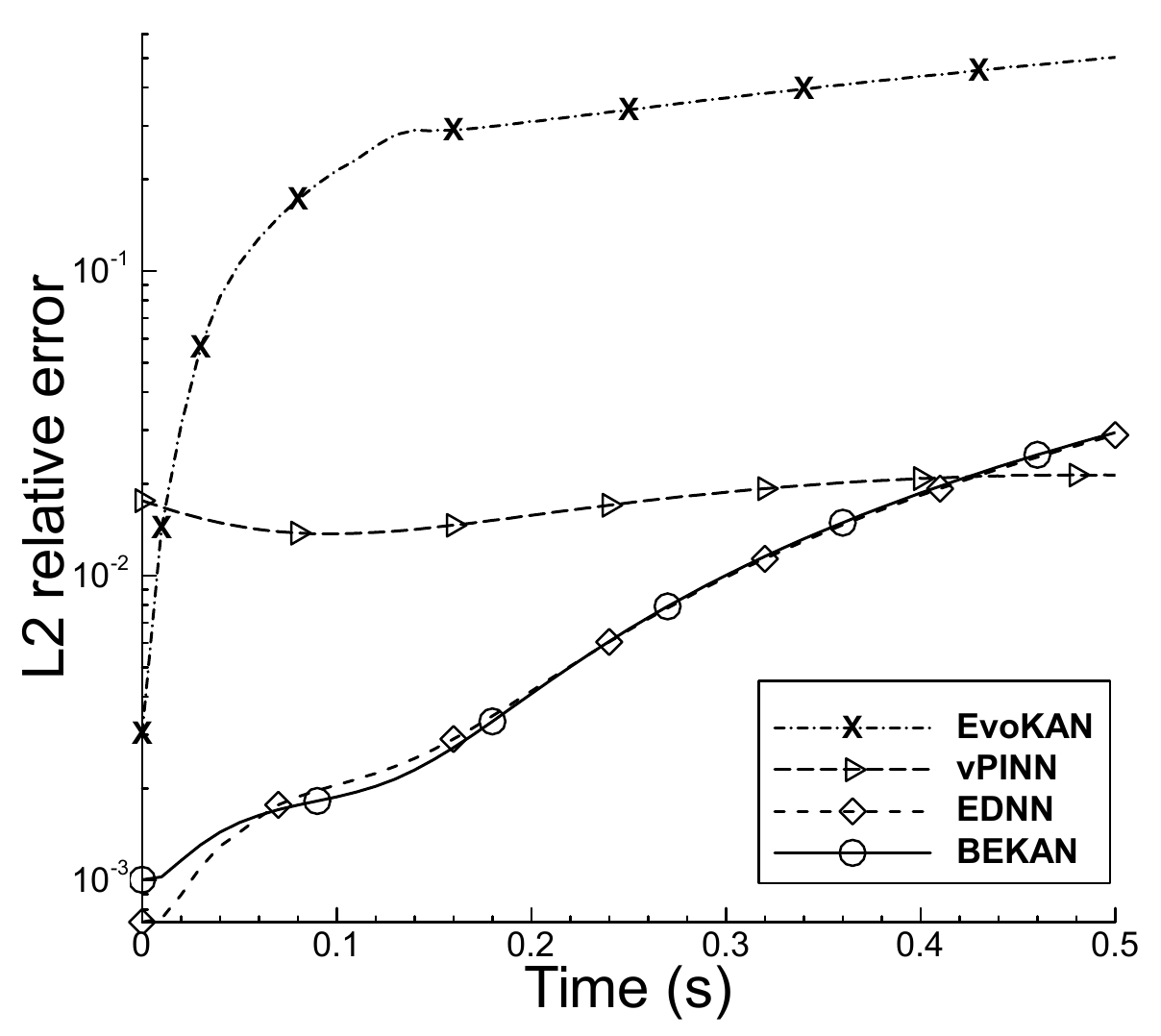}
    \caption{
Time evolution of the $L_2$ relative error for three models: BEKAN, EvoKAN, and EDNN, applied to the 2D heat equation with a mixed boundary condition (Eq.~\eqref{eq:heat_equation_mixed}).  
The error is computed at each time step using the FDM solution as the reference.  
Among the models, both BEKAN and EDNN consistently show the lowest $L_2$ relative error throughout the simulation.
}
    \label{fig:L2_mixed}
\end{figure}

\begin{figure}[htbp!]
\hspace{45 pt}
    \begin{subfigure}[b]{0.4\linewidth}
        \includegraphics[width=\linewidth]{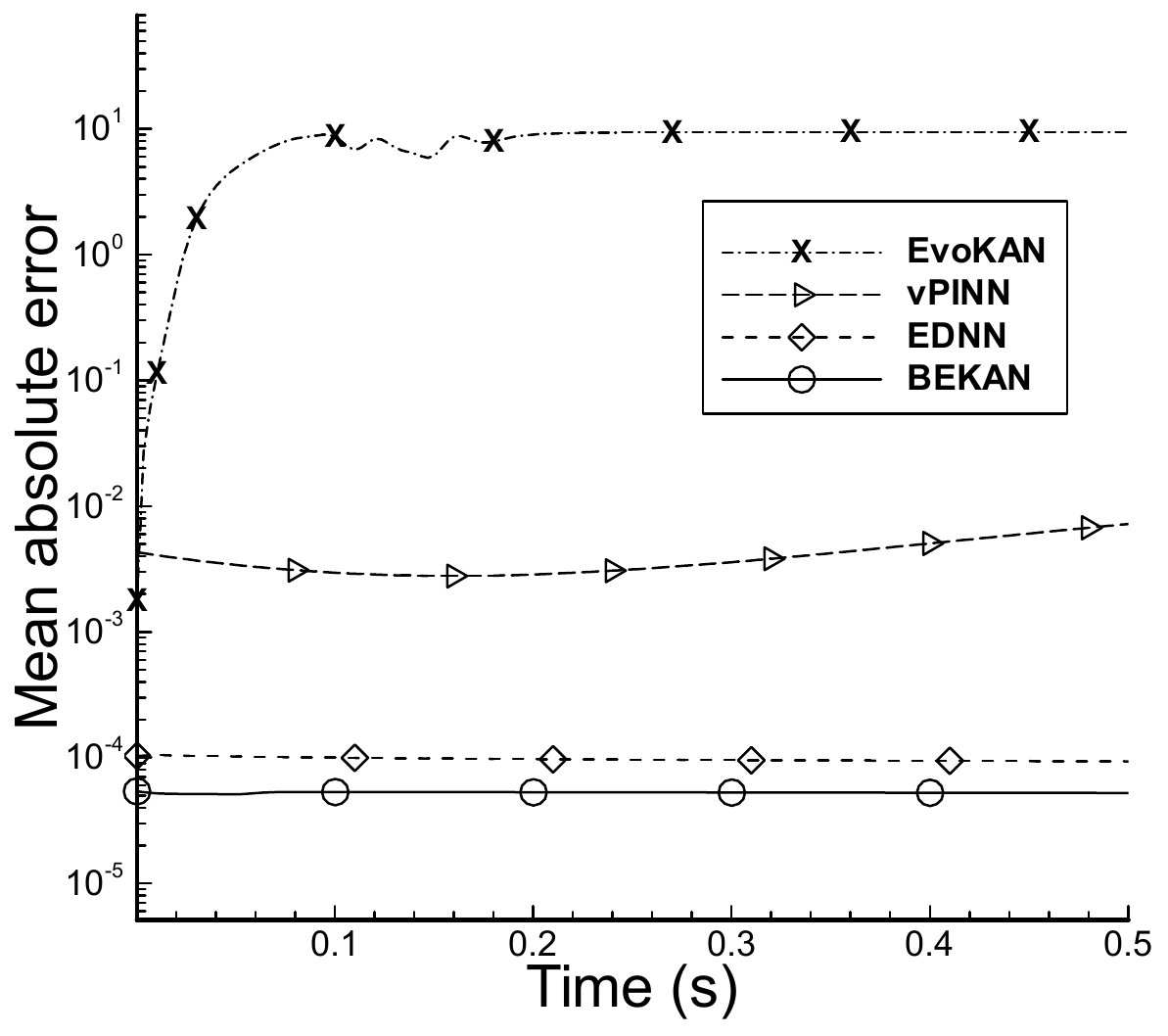}
        \caption{Error of gradient on the right boundary}
    \end{subfigure}
    % \hfill
    \begin{subfigure}[b]{0.4\linewidth}
        \includegraphics[width=\linewidth]{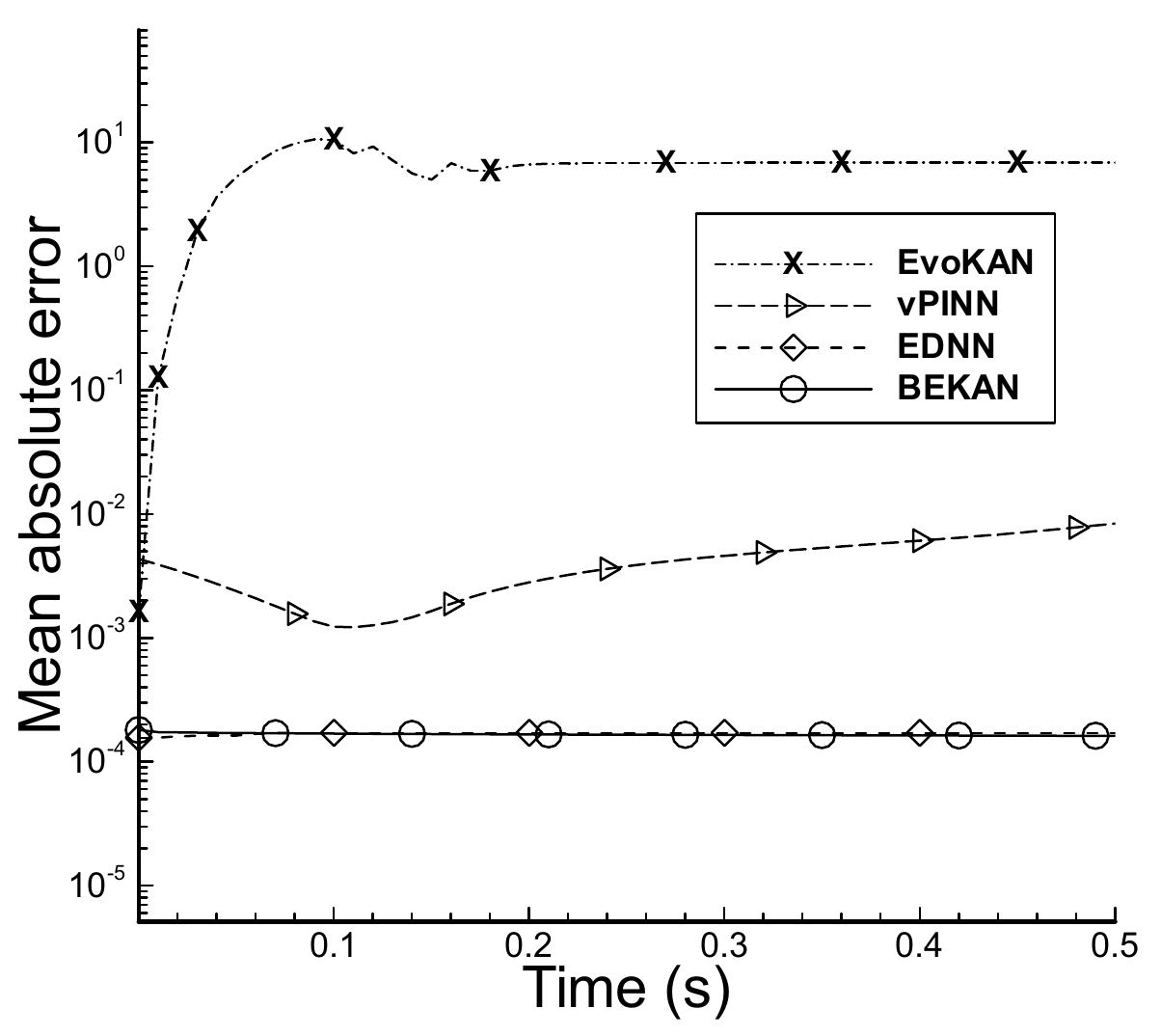}
        \caption{Error of gradient on the upper boundary} 
    \end{subfigure} 
    \caption{
2D heat equation with mixed boundary conditions (Eq.~\eqref{eq:heat_equation_mixed}): Mean absolute error of the gradient on each boundary for BEKAN, EvoKAN, EDNN, and PINN, evaluated against the FDM solution.  
While BEKAN, EDNN, and PINN all show reasonably small errors, BEKAN and EDNN, which adopt the proposed method described in Sec.~\ref{subsec:Neumann}, maintain consistently low and stable errors over the entire time range.
}
    \label{fig:mixed_gradient}
\end{figure}
% We evaluate the Neumann boundary condition accuracy across all time steps by tracking the mean absolute gradient error along the four boundaries.
% Figure~\ref{fig:mixed_gradient} shows that BEKAN, EDNN, and EvoKAN, which use the proposed approach described in Sec~\ref{subsec:Neumann}, maintain steady error levels over time.
% In Fig.~\ref{fig:mixed_gradient}, the vanilla PINN shows variations in boundary error as time progresses.
% Overall, BEKAN achieves the smallest error throughout the entire simulation.

\begin{table}[htbp!]
\footnotesize
\renewcommand{\arraystretch}{1.0}
\centering
\caption{2D heat equation with mixed boundary conditions (Eq.~\eqref{eq:heat_equation_mixed}): Predicted values of the solution \( u(x, t) \) at the domain boundaries for BEKAN, EvoKAN, EDNN, and vanilla PINN, evaluated at \( t = \SI{2e-1}{} \) and \( \SI{5e-1}{} \), to assess compliance with the Dirichlet boundary conditions specified in Eq.~\eqref{eq:heat_equation_mixed}.  
BEKAN employs the proposed method for enforcing Dirichlet conditions as described in Sec.~\ref{subsec:Dirichlet}, while EvoKAN and EDNN implement output transformation techniques to impose hard constraints~\cite{guyiqi}.
}
\begin{tabular}{l c c c c c}
\hline
 & BEKAN & EvoKAN & EDNN & Vanilla PINN & Exact solution \\
\hline
$u(x=0,\, y=0, t=0.2)$ & $0.00000\mathrm{e}{+00}$ & $0.00000\mathrm{e}{+00}$ & $0.00000\mathrm{e}{+00}$ & $-2.81790\mathrm{e}{-03}$ & $0.00000\mathrm{e}{+00}$ \\
$u(x=0,\,y=1,  t=0.2)$  & $0.00000\mathrm{e}{+00}$ & $0.00000\mathrm{e}{+00}$ & $0.00000\mathrm{e}{+00}$ & $2.39291\mathrm{e}{-03}$ & $0.00000\mathrm{e}{+00}$ \\
$u(x=1,\, y=0, t=0.2)$ & $0.00000\mathrm{e}{+00}$ & $0.00000\mathrm{e}{+00}$ & $0.00000\mathrm{e}{+00}$ & $2.96441\mathrm{e}{-02}$ & $0.00000\mathrm{e}{+00}$ \\
$u(x=0,\,y=0, t=0.5)$  & $0.00000\mathrm{e}{+00}$ & $0.00000\mathrm{e}{+00}$ & $0.00000\mathrm{e}{+00}$ & $2.46784\mathrm{e}{-03}$ & $0.00000\mathrm{e}{+00}$ \\
$u(x=0,\, y=1, t=0.5)$ & $0.00000\mathrm{e}{+00}$ & $0.00000\mathrm{e}{+00}$ & $0.00000\mathrm{e}{+00}$ & $-2.47578\mathrm{e}{-03}$ & $0.00000\mathrm{e}{+00}$ \\
$u(x=1,\,y=0,  t=0.5)$  & $0.00000\mathrm{e}{+00}$ & $0.00000\mathrm{e}{+00}$ & $0.00000\mathrm{e}{+00}$ & $2.37548\mathrm{e}{-03}$ & $0.00000\mathrm{e}{+00}$ \\
\hline
\end{tabular}
\label{table:Mixed_BC}
\end{table}

Next, we track the evolution of the \( L_2 \) relative error over the entire simulation in Fig.~\ref{fig:L2_mixed}.  
BEKAN, EvoKAN, and EDNN show a gradual increase in error as time advances.  
In contrast, vanilla PINN maintains a relatively steady error level, and at the final time step \( t = \SI{5e-1}{} \), its error is slightly lower than those of BEKAN and EDNN.  
However, as shown in Table~\ref{table:Mixed_BC}, The vanilla PINN exhibits difficulty in satisfying the homogeneous Dirichlet boundary constraints specified at \( x = 0 \) and \( y = 0 \).
To assess how well the homogeneous Neumann boundary conditions at \( x = 1 \) and \( y = 1 \) are maintained, Fig.~\ref{fig:mixed_gradient} displays the temporal evolution of the mean absolute gradient error.  
Among the models, EvoKAN shows the largest gradient error throughout the simulation.  
Although vanilla PINN maintains a moderate level of $L_2$ error, it exhibits temporal fluctuations and performs worse than BEKAN and EDNN as depicted in Fig.~\ref{fig:mixed_gradient}.  
Overall, for the mixed boundary condition example, both BEKAN and EDNN enforce the Dirichlet conditions exactly and achieve the lowest gradient errors on the Neumann boundaries.

% ==============================
% Section: Conclusion
% ==============================
\section{Conclusion} \label{sec:Conclusion}

We proposed a novel approach, BEKAN, for solving \ac{PDEs} with rigorous enforcement of Dirichlet, periodic, Neumann boundary conditions, and their combinations.
To address Dirichlet boundary value problems, we leveraged \ac{KAN} and Gaussian \ac{RBFs} to encode boundary information directly into the network.
Inspired by the interpretability of \ac{KAN}, we designed boundary-guaranteed activation functions composed of basis functions formed by smooth and globalized Gaussian \ac{RBFs}.
For periodic boundary condition problems, we introduced a periodic layer as the first hidden layer to ensure that the solution exactly satisfies periodicity.
For Neumann boundary value problems, we employed an evolutionary network to guide the network parameters at each discretized time step toward satisfying the Neumann boundary condition.

As a result, we demonstrate the effectiveness of our approach by solving PDEs subject to different boundary conditions, achieving high accuracy across five numerical examples.  
For the Dirichlet boundary condition, the capability of BEKAN is tested on two representative PDEs: the 1D Allen–Cahn equation in Eq.~\eqref{eq:Allen-Cahn} and the 2D Burgers’ equation in Eq.~\eqref{eq:Burgers_equation}.  
BEKAN outperforms EvoKAN, EDNN, and vanilla PINN in terms of accuracy, both over the entire domain and on the boundaries, as depicted in Figs.~\ref{fig:Burgers_field_distribution} and \ref{fig:Burgers_L2}.  
For the periodic boundary condition, we solve the 1D KS equation in Eq.~\eqref{eq:KS_equation}, which is a challenging PDE due to its high-order derivatives and chaotic behavior.  
The choice of Gaussian RBFs over B-splines is critical for numerical stability when solving stiff or chaotic PDEs. As demonstrated with the KS equation, the smooth, non-vanishing nature of Gaussian RBFs leads to well-conditioned Jacobians during parameter evolution, as delineated in Fig.~\ref{fig:condition}, a problem that renders B-spline based KANs and traditional NNs unstable. This makes BEKAN uniquely suited for such challenging physical systems.  
Regarding periodic boundary enforcement, BEKAN achieves exact satisfaction of the periodic boundary condition, as shown in Table~\ref{table:periodic}.  
As a test case for the Neumann boundary condition, we analyze the behavior of the 2D heat equation incorporating a nonlinear forcing component in Eq.~\eqref{eq:heat_equation}.  
While BEKAN, EvoKAN, EDNN, and vanilla PINN exhibit similar solution distributions across the domain, BEKAN achieves the lowest \( L_2 \) relative error and gradient error on the boundaries, as depicted in Figs.~\ref{fig:heat_L2} and \ref{fig:Error of gradient on the boundary}, respectively.  
Lastly, we examine a heat equation problem subject to a combination of Dirichlet and Neumann boundary conditions in Eq.~\eqref{eq:heat_equation_mixed}.  
In this mixed boundary value problem, BEKAN and EDNN show high accuracy on both the domain and boundaries, as represented in Figs.~\ref{fig:L2_mixed} and \ref{fig:mixed_gradient}, respectively.  
Vanilla PINN also demonstrates reasonable accuracy, but it fails to exactly satisfy the Dirichlet condition and shows fluctuating gradient error on the Neumann boundary.

In conclusion, we demonstrated that the proposed method can accurately solve various challenging PDE problems while enforcing boundary conditions. 
% The primary contribution of this work is not just a single new component, but the successful synthesis of an evolutionary framework with a novel, boundary-guaranteed KAN architecture. This hybrid approach leverages the expressiveness of KANs, the stability of Gaussian RBFs, and the time-domain accuracy of an evolutionary solver to create a robust and accurate tool for a wide class of PDE problems that have remained challenging for monolithic deep learning models.
This study addresses the difficulty of incorporating boundary constraints into black-box neural network models in a principled manner.  
The BEKAN framework offers a potential pathway toward reliable machine learning-based predictions in computational science and engineering.  
As future work, BEKAN can be extended to uncertainty quantification, enabling efficient simulation of uncertainty propagation in initial conditions or coefficients of PDEs under strictly enforced boundary conditions.

\section*{Acknowledgments}
We would like to thank the support of National Science Foundation (DMS-2053746, DMS-2134209, ECCS-2328241, CBET-2347401 and OAC-2311848h), and U.S.~Department of Energy (DOE) Office of Science Advanced Scientific Computing Research program DE-SC0023161, and DOE–Fusion Energy Science, under grant number: DE-SC0024583.
%%Vancouver style references.
% \clearpage
\bibliographystyle{model1-num-names}
\bibliography{ref,ref_qg,ref_kan}

\end{document}